\documentclass[twoside,11pt]{article}

\usepackage{jmlr2e}
\usepackage{nicefrac}
\usepackage{amsmath}
\usepackage{textcomp}

\usepackage{enumitem}
\setlist[itemize]{leftmargin=0.8cm}
\setlist[enumerate]{leftmargin=0.8cm}

\usepackage{hyperref}
\hypersetup{hypertexnames=false}
\usepackage{bookmark}
\usepackage{caption}
\usepackage{subcaption}

\usepackage{floatrow}
\newfloatcommand{capbtabbox}{table}[][\FBwidth]

\newcommand{\itemEq}[1]{%
         \begingroup%
         \setlength{\abovedisplayskip}{0pt}%
         \setlength{\belowdisplayskip}{0pt}%
         \parbox[c]{\linewidth}{\begin{flalign}#1&&\end{flalign}}%
         \endgroup}

\usepackage{lastpage}
\jmlrheading{19}{2018}{1-\pageref{LastPage}}{}{}{}{Bin Dai and David Wipf}
\ShortHeadings{Diagnosing and Enhancing VAE Models}{Dai and Wipf}

\firstpageno{1}
\input{def.set}

\begin{document}

\title{Diagnosing and Enhancing VAE Models}

\author{\name Bin Dai \email daib13@mails.tsinghua.edu.cn \\
       \addr Institute for Advanced Study\\
       Tsinghua University\\
       Beijing, China
       \AND
       \name David Wipf \email davidwipf@gmail.com\\
       \addr Microsoft Research \\
       Beijing, China
}

\editor{}

\maketitle

\begin{abstract}
Although variational autoencoders (VAEs) represent a widely influential deep generative model, many aspects of the underlying energy function remain poorly understood.  In particular, it is commonly believed that Gaussian encoder/decoder assumptions reduce the effectiveness of VAEs in generating realistic samples.  In this regard, we rigorously analyze the VAE objective, differentiating situations where this belief is and is not actually true.  We then leverage the corresponding insights to develop a simple VAE enhancement that requires no additional hyperparameters or sensitive tuning.  Quantitatively, this proposal produces crisp samples and stable FID scores that significantly reduce the gap with GAN models when a neutral architecture is applied, all while retaining desirable attributes of the original VAE architecture.  A shorter version of this work has been accepted to the ICLR 2019 conference proceedings \citep{Dai2019iclr}. The code for our model is available at \url{https://github.com/daib13/TwoStageVAE}.
\end{abstract}

\begin{keywords}
Variational Autoencoder, Deep Generative Model
\end{keywords}


\section{Introduction}\label{sec:intro}

Our starting point is the desire to learn a probabilistic generative model of observable variables $\bx\in\bchi$, where $\bchi$ is a $r$-dimensional manifold embedded in $\mathbb{R}^d$. Note that if $r=d$, then this assumption places no restriction on the distribution of $\bx \in \mathbb{R}^d$ whatsoever; however, the added formalism is introduced to handle the frequently encountered case where $\bx$ possesses low-dimensional structure relative to a high-dimensional ambient space, i.e., $r\ll d$.  In fact, the very utility of generative models of continuous data, and their attendant low-dimensional representations, often hinges on this assumption \citep{bengio2013representation}.  It therefore behooves us to explicitly account for this situation.

Beyond this, we assume that $\bchi$ is a simple Riemannian manifold, which means there exists a diffeomorphism $\varphi$ between $\bchi$ and $\mathbb{R}^r$, or more explicitly, the mapping ~$\varphi: \bchi \mapsto \mathbb{R}^r$~is invertible and differentiable.\footnote{Although this assumption is sufficient for modeling a wide variety of data structures arbitrarily well, there are of course real-world examples that involve alternative topologies.  For one interesting representative example, see \citep{davidson2018hyperspherical}.} Denote a ground-truth probability measure on $\bchi$ as $\mu_{gt}$ such that the probability mass of an infinitesimal $d\bx$ on the manifold is $\mu_{gt}(d\bx)$ and $\int_{\bchi} \mu_{gt}(d \bx)=1$.

The variational autoencoder (VAE) \citep{Kingma2014,Rezende2014} attempts to approximate this ground-truth measure using a parameterized density $p_\theta(\bx)$ defined across all of $\mathbb{R}^d$  since any underlying generative manifold is unknown in advance.  This density is further assumed to admit the latent decomposition $p_\theta(\bx) = \int p_\theta(\bx|\bz) p(\bz) d\bz$, where $\bz \in \mathbb{R}^{\kappa}$ serves as a low-dimensional representation, with $\kappa \approx r$ and prior $p(\bz) = \calN(\bz| {\bf 0}, \bI)$.

Ideally we might like to minimize the negative log-likelihood $-\log p_\theta(\bx)$ averaged across the ground-truth measure $\mu_{gt}$, i.e., solve $\min_{\theta} \int_{\bchi} -\log p_\theta(\bx) \mu_{gt}(d\bx)$.  Unfortunately though, the required marginalization over $\bz$ is generally infeasible.  Instead the VAE model relies on tractable \emph{encoder} $q_\phi(\bz|\bx)$ and \emph{decoder} $p_\theta(\bx|\bz)$ distributions, where $\phi$ represents additional trainable parameters. The canonical VAE cost is a bound on the average negative log-likelihood given by
\begin{equation} \label{eqn:objective_general}
    \textstyle{ \mathcal{L}(\theta,\phi)  \triangleq   \int_{\bchi} \left\{ -\log p_\theta(\bx) ~ + ~ \mathbb{KL}\left[ q_\phi(\bz|\bx) || p_\theta(\bz|\bx) \right] \right\} \mu_{gt}(d\bx) \geq \int_{\bchi} -\log p_\theta(\bx) \mu_{gt}(d\bx)},
\end{equation}
where the inequality follows directly from the non-negativity of the KL-divergence.  Here $\phi$ can be viewed as tuning the tightness of bound, while $\theta$ dictates the actual estimation of $\mu_{gt}$.  Using a few standard manipulations, this bound can also be expressed as
\begin{equation} \label{eq:objective_general2}
\textstyle{\mathcal{L}(\theta,\phi) = \int_{\bchi} \left\{ - \mathbb{E}_{q_{\tiny \phi}\left(\bz|\bx \right)} \left[\log p_{\tiny \theta} \left(\bx | \bz  \right)  \right] + \mathbb{KL}\left[ q_\phi(\bz|\bx) || p(\bz) \right]\right\} \mu_{gt}(d\bx)},
\end{equation}
which explicitly involves the encoder/decoder distributions and is conveniently amenable to SGD optimization of $\{\theta,\phi\}$ via a reparameterization trick \citep{Kingma2014,Rezende2014}.  The first term in (\ref{eq:objective_general2}) can be viewed as a reconstruction cost (or a stochastic analog of a traditional autoencoder), while the second penalizes posterior deviations from the prior $p(\bz)$.  Additionally, for any realizable implementation via SGD, the integration over $\bchi$ must be approximated via a finite sum across training samples $\{ \bx^{(i)} \}_{i=1}^n$ drawn from $\mu_{gt}$.  Nonetheless, examining the true objective $\mathcal{L}(\theta,\phi)$ can lead to important, practically-relevant insights.

At least in principle, $q_\phi(\bz|\bx)$ and $p_\theta(\bx|\bz)$ can be arbitrary distributions, in which case we could simply enforce $q_\phi(\bz|\bx) = p_\theta(\bz|\bx) \propto p_\theta(\bx|\bz)p(\bz)$ such that the bound from (\ref{eqn:objective_general}) is tight.  Unfortunately though, this is essentially always an intractable undertaking.  Consequently, largely to facilitate practical implementation, a commonly adopted distributional assumption for continuous data is that both $q_\phi(\bz|\bx)$ and $p_\theta(\bx|\bz)$ are Gaussian.  This design choice has previously been cited as a key limitation of VAEs \citep{burda2015importance,kingma2016improved}, and existing quantitative tests of generative modeling quality thus far dramatically favor contemporary alternatives such as generative adversarial networks (GAN) \citep{goodfellow2014generative}.  Regardless, because the VAE possesses certain desirable properties relative to GAN models (e.g., stable training \citep{tolstikhin2018wasserstein}, interpretable encoder/inference network \citep{brock2016neural}, outlier-robustness \citep{dai2018jmlr}, etc.), it remains a highly influential paradigm worthy of examination and enhancement.


In Section \ref{sec:impact_gaussian_assumption} we closely investigate the implications of VAE Gaussian assumptions leading to a number of interesting diagnostic conclusions.  In particular, we differentiate the situation where $r=d$, in which case we prove that recovering the ground-truth distribution is actually possible iff the VAE global optimum is reached, and $r<d$, in which case the VAE global optimum can be reached by solutions that reflect the ground-truth distribution almost everywhere, but not necessarily uniquely so.  In other words, there could exist alternative solutions that both reach the global optimum and yet do not assign the same probability measure as $\mu_{gt}$.

Section \ref{sec:optima_property} then further probes this non-uniqueness issue by inspecting necessary conditions of global optima when $r<d$.  This analysis reveals that an optimal VAE parameterization will provide an encoder/decoder pair capable of perfectly reconstructing all $\bx \in \bchi$ using \emph{any} $\bz$ drawn from $q_\phi(\bz|\bx)$.  Moreover, we demonstrate that the VAE accomplishes this using a degenerate latent code whereby only $r$ dimensions are effectively active.  Collectively, these results indicate that the VAE global optimum can in fact uniquely learn a mapping to the correct ground-truth manifold when $r<d$, but not necessarily the correct probability measure \emph{within} this manifold, a critical distinction.

Next we leverage these analytical results in Section \ref{sec:model} to motivate an almost trivially-simple, two-stage VAE enhancement for addressing typical regimes when $r<d$.  In brief, the first stage just learns the manifold per the allowances from Section \ref{sec:optima_property}, and in doing so, provides a mapping to a lower dimensional intermediate representation with no degenerate dimensions that mirrors the $r=d$ regime. The second (much smaller) stage then only needs to learn the correct probability measure on this intermediate representation, which is possible per the analysis from Section \ref{sec:impact_gaussian_assumption}.  Experiments from Section \ref{sec:experiments} reveal that this procedure can generate high-quality crisp samples, avoiding the blurriness often attributed to VAE models in the past \citep{dosovitskiy2016generating, larsen2015autoencoding}.  And to the best of our knowledge, this is the first demonstration of a VAE pipeline that can produce stable FID scores, an influential recent metric for evaluating generated sample quality \citep{heusel2017gans}, that are comparable to at least some popular GAN models under neutral testing conditions.  Moreover, this is accomplished without additional penalty functions, cost function modifications, or sensitive tuning parameters.  Finally, Section \ref{sec:discussion} provides concluding thoughts and a discussion of broader VAE modeling paradigms, such as those involving normalizing flows, parameterized families for $p(\bz)$, or modifications to encourage disentangled representations.



\section{High-Level Impact of VAE Gaussian Assumptions}\label{sec:impact_gaussian_assumption}


Conventional wisdom suggests that VAE Gaussian assumptions will introduce a gap between $\mathcal{L}(\theta,\phi)$ and the ideal negative log-likelihood $\int_{\bchi} -\log p_\theta(\bx) \mu_{gt}(d\bx)$, compromising efforts to learn the ground-truth measure.  However, we will now argue that this pessimism is in some sense premature.  In fact, we will demonstrate that, even with the stated Gaussian distributions, there exist parameters $\phi$ and $\theta$ that can simultaneously: (\emph{i}) Globally optimize the VAE objective and, (\emph{ii}) Recover the ground-truth probability measure in a certain sense described below.  This is possible because, at least for some coordinated values of $\phi$ and $\theta$, $q_\phi(\bz|\bx)$ and $p_\theta(\bz|\bx)$  can indeed become arbitrarily close.  Before presenting the details, we first formalize a $\kappa$-simple VAE, which is merely a VAE model with explicit Gaussian assumptions and parameterizations:
\begin{definition}
    A $\kappa$-simple VAE is defined as a VAE model with $\mbox{dim}[\bz] = \kappa$ latent dimensions, the Gaussian encoder $q_\phi(\bz|\bx)=\mathcal{N}(\bz|\bmu_z,\bSigma_z)$, and the Gaussian decoder $p_\theta(\bx|\bz)=\mathcal{N}(\bx|\bmu_x, \bSigma_x)$.
Moreover, the encoder moments are defined as $\bmu_z = f_{\mu_z}(\bx;\phi)$ and $\bSigma_z = \bS_z\bS_z^\top$ with $\bS_z = f_{S_z}(\bx;\phi)$. Likewise, the decoder moments are $\bmu_x = f_{\mu_x}(\bz;\theta)$ and $\bSigma_x = \gamma\bI$. Here $\gamma > 0$ is a tunable scalar, while $f_{\mu_z}$, $f_{S_z}$ and $f_{\mu_x}$ specify parameterized differentiable functional forms that can be arbitrarily complex, e.g., a deep neural network.
\end{definition}

Equipped with these definitions, we will now demonstrate that a $\kappa$-simple VAE, with $\kappa \ge r$, can achieve the optimality criteria (\emph{i}) and (\emph{ii}) from above.   In doing so, we first consider the simpler case where $r=d$, followed by the extended scenario with $r<d$.  The distinction between these two cases turns out to be significant, with practical implications to be explored in Section \ref{sec:model}.


\subsection{Manifold Dimension Equal to Ambient Space Dimension ($r = d$)} \label{sec:r_equals_d_case}

We first analyze the specialized situation where $r = d$. Assuming $p_{gt}(\bx) \triangleq \mu_{gt}(d\bx)/d\bx$ exists everywhere in $\mathbb{R}^d$, then $p_{gt}(\bx)$ represents the ground-truth probability density with respect to the standard Lebesgue measure in Euclidean space.  Given these considerations, the minimal possible value of (\ref{eqn:objective_general}) will necessarily occur if
\begin{equation} \label{eq:vae_optimality_conditions}
	\mathbb{KL} \left[ q_\phi(\bz|\bx) || p_\theta(\bz|\bx) \right] = 0 ~~~~ \text{and} ~~~~ p_\theta(\bx) = p_{gt}(\bx) \mbox{ almost everywhere}.
\end{equation}
This follows because by VAE design it must be that $\mathcal{L}(\theta,\phi) \geq -\int p_{gt}(\bx) \log p_{gt} (\bx) d\bx$, and in the present context, this lower bound is achievable iff the conditions from (\ref{eq:vae_optimality_conditions}) hold.  Collectively, this implies that the approximate posterior produced by the encoder $q_\phi(\bz|\bx)$ is in fact perfectly matched to the actual posterior $p_\theta(\bz|\bx)$, while the corresponding marginalized data distribution $p_\theta(\bx)$ is perfectly matched the ground-truth density $p_{gt}(\bx)$ as desired.  Perhaps surprisingly, a $\kappa$-simple VAE can actually achieve such a solution:

\begin{theorem}\label{thm:optima_r_eq_d}
Suppose that $r = d$ and there exists a density $p_{gt}(\bx)$ associated with the ground-truth measure $\mu_{gt}$ that is nonzero everywhere on $\mathbb{R}^d$.\footnote{This nonzero assumption can be replaced with a much looser condition.  Specifically, if there exists a diffeomorphism between the set $\{ \bx | p_{gt}(\bx) \neq 0 \}$ and $\mathbb{R}^d$, then it can be shown that Theorem \ref{thm:optima_r_eq_d} still holds even if $p_{gt}(\bx) = 0$ for some $\bx \in \mathbb{R}^d$.}. Then for any $\kappa \geq r$, there is a sequence of $\kappa$-simple VAE model parameters $\{ \theta_t^*, \phi_t^* \}$ such that
    \begin{equation}
        \lim_{t\to\infty}\mathbb{KL} \left[ q_{\phi_t^*}(\bz|\bx) || p_{\theta_t^*}(\bz|\bx) \right] = 0 ~~~~ \text{and} ~~~~
        \lim_{t\to\infty}p_{\theta_t^*}(\bx) = p_{gt}(\bx)  \mbox{ almost everywhere}.
    \end{equation}
\end{theorem}

All the proofs can be found in the appendices.  So at least when $r = d$, the VAE Gaussian assumptions need not actually prevent the optimal ground-truth probability measure from being recovered, as long as the latent dimension is sufficiently large (i.e., $\kappa \geq r$).  And contrary to popular notions, a richer class of distributions is not required to achieve this.  Of course Theorem \ref{thm:optima_r_eq_d} only applies to a restricted case that excludes $d>r$; however, later we will demonstrate that a key consequence of this result can nonetheless be leveraged to dramatically enhance VAE performance.


\subsection{Manifold Dimension Less Than Ambient Space Dimension ($r < d$)}

When $r < d$, additional subtleties are introduced that will be unpacked both here and in the sequel.  To begin, if both $q_\phi(\bz|\bx)$ and  $p_\theta(\bx|\bz)$ are arbitrary/unconstrained (i.e., not necessarily Gaussian), then $\inf_{\phi,\theta} \mathcal{L}(\theta,\phi) = -\infty$.  To achieve this global optimum, we need only choose $\phi$ such that $q_\phi(\bz|\bx) = p_\theta(\bz|\bx)$ (minimizing the KL term from (\ref{eqn:objective_general})) while selecting $\theta$ such that all probability mass collapses to the correct manifold $\bchi$.  In this scenario the density $p_\theta(\bx)$ will become unbounded on $\bchi$ and zero elsewhere, such that $\int_{\bchi} -\log p_\theta(\bx) \mu_{gt}(d\bx)$ will approach negative infinity.

But of course the stated Gaussian assumptions from the $\kappa$-simple VAE model could ostensibly prevent this from occurring by causing the KL term to blow up, counteracting the negative log-likelihood factor.  We will now analyze this case to demonstrate that this need not happen.  Before proceeding to this result, we first define a manifold density $\tilde{p}_{gt}(\bx)$ as the probability density (assuming it exists) of $\mu_{gt}$ with respect to the volume measure of the manifold $\bchi$.  If $d=r$ then this volume measure reduces to the standard Lebesgue measure in $\mathbb{R}^d$ and $\tilde{p}_{gt}(\bx) = p_{gt}(\bx)$; however, when $d>r$ a density $p_{gt}(\bx)$ defined in $\mathbb{R}^d$ will not technically exist, while $\tilde{p}_{gt}(\bx)$ is still perfectly well-defined.  We then have the following:



\begin{theorem}\label{thm:optima_r_less_d}
Assume $r<d$ and that there exists a manifold density $\tilde{p}_{gt}(\bx)$ associated with the ground-truth measure $\mu_{gt}$ that is nonzero everywhere on $\bchi$.  Then for any $\kappa \geq r$, there is a sequence of $\kappa$-simple VAE model parameters $\{ \theta_t^*, \phi_t^* \}$ such that

	\begin{enumerate}[label=(\roman*)]
		\item \itemEq{\lim_{t\to\infty}\mathbb{KL} \left[ q_{\phi^*_t}(\bz|\bx) || p_{\theta^*_t}(\bz|\bx) \right] = 0 ~~~~ \text{and} ~~~~\lim_{t\to\infty}\textstyle{\int_{\bchi} -\log p_{\theta^*_t}(\bx) \mu_{gt}(d\bx)= -\infty}, \label{eqn:achieve_optima}}

		\item \itemEq{\lim_{t\to\infty} \textstyle{\int_{\bx \in A} p_{\theta^*_t}(\bx) d\bx = \mu_{gt}(A\cap\bchi) } \label{eqn:match_distributiin}} for all measurable sets $A\subseteq\mathbb{R}^d$ with $\mu_{gt}(\partial A\cap \bchi)=0$, where $\partial A$ is the boundary of $A$.
	\end{enumerate}
\end{theorem}

Technical details notwithstanding, Theorem \ref{thm:optima_r_less_d} admits  a very intuitive interpretation.  First, (\ref{eqn:achieve_optima}) directly implies that the VAE Gaussian assumptions do not prevent minimization of $\mathcal{L}(\theta,\phi)$ from converging to minus infinity, which can be trivially viewed as a globally optimum solution.  Furthermore, based on (\ref{eqn:match_distributiin}), this solution can be achieved with a limiting density estimate that will assign a probability mass to most all measurable subsets of $\mathbb{R}^d$ that is indistinguishable from the ground-truth measure (which confines all mass to $\bchi$).  Hence this solution is more-or-less an arbitrarily-good approximation to $\mu_{gt}$ for all practical purposes.\footnote{Note that (\ref{eqn:match_distributiin}) is only framed in this technical way to accommodate the difficulty of comparing a measure $\mu_{gt}$ restricted to $\bchi$ with the VAE density $p_{\theta}(\bx)$ defined everywhere in $\mathbb{R}^d$.  See the appendices for details.}



Regardless, there is an absolutely crucial distinction between Theorem \ref{thm:optima_r_less_d} and the simpler case quantified by Theorem \ref{thm:optima_r_eq_d}.  Although both describe conditions whereby the $\kappa$-simple VAE can achieve the minimal possible objective, in the $r=d$ case achieving the lower bound (whether the specific parameterization for doing so is unique or not) necessitates that the ground-truth probability measure has been recovered almost everywhere.  But the $r<d$ situation is quite different because we have not ruled out the possibility that a different set of parameters $\{\theta,\phi\}$ could push $\calL(\theta,\phi)$ to $-\infty$ and yet not achieve (\ref{eqn:match_distributiin}).  In other words, the VAE could reach the lower bound but fail to closely approximate $\mu_{gt}$.  And we stress that this uniqueness issue is not a consequence of the VAE Gaussian assumptions per se; even if $q_{\phi}(\bz|\bx)$ were unconstrained the same lack of uniqueness can persist.

Rather, the intrinsic difficulty is that, because the VAE model does not have access to the ground-truth low-dimensional manifold, it must implicitly rely on a density $p_\theta(\bx)$ defined across \emph{all} of $\mathbb{R}^d$ as mentioned previously.  Moreover, if this density converges towards infinity on the manifold during training without increasing the KL term at the same rate, the VAE cost can be unbounded from below, even in cases where (\ref{eqn:match_distributiin}) is not satisfied, meaning incorrect assignment of probability mass.


To conclude, the key take-home message from this section is that, at least in principle, VAE Gaussian assumptions need not actually be the root cause of any failure to recover ground-truth distributions.  Instead we expose a structural deficiency that lies elsewhere, namely, the non-uniqueness of solutions that can optimize the VAE objective without necessarily learning a close approximation to $\mu_{gt}$.  But to probe this issue further and motivate possible workarounds, it is critical to further disambiguate these optimal solutions and their relationship with ground-truth manifolds.  This will be the task of Section \ref{sec:optima_property}, where we will explicitly differentiate the problem of locating the correct ground-truth manifold, from the task of learning the correct probability measure \emph{within} the manifold.


Note that the only comparable prior work we are aware of related to the results in this section comes from \citet{doersch2016tutorial}, where the implications of adopting Gaussian encoder/decoder pairs in the specialized case of $r = d = 1$ are briefly considered.  Moreover, the analysis there requires additional much stronger assumptions than ours, namely, that $p_{gt}(\bx)$ should be nonzero and infinitely differentiable everywhere in the requisite 1D ambient space. These requirements of course exclude essentially all practical usage regimes where $d = r > 1$ or $d > r$, or when ground-truth densities are not sufficiently smooth.


\section{Optimal Solutions and the Ground Truth Manifold}\label{sec:optima_property}

We will now more closely examine the properties of optimal $\kappa$-simple VAE solutions, and in particular, the degree to which we might expect them to at least reflect the true $\bchi$, even if perhaps not the correct probability measure $\mu_{gt}$ defined within $\bchi$.  To do so, we must first consider some \emph{necessary} conditions for VAE optima:

\begin{theorem}\label{thm:decoder_variance}
Let $\{ \theta^*_\gamma, \phi^*_\gamma \}$ denote an optimal $\kappa$-simple VAE solution (with $\kappa \geq r$) where the decoder variance $\gamma$ is fixed (i.e., it is the sole unoptimized parameter).  Moreover, we assume that $\mu_{gt}$ is not a Gaussian distribution when $d=r$.\footnote{This requirement is only included to avoid a practically irrelevant form of non-uniqueness that exists with full, non-degenerate Gaussian distributions.}  Then for any $\gamma > 0$, there exists a $\gamma^\prime<\gamma$ such that $\mathcal{L}(\theta_{\gamma^\prime}^*, \phi_{\gamma^\prime}^*) < \mathcal{L}(\theta_\gamma^*, \phi_\gamma^*)$.
\end{theorem}

This result implies that we can always reduce the VAE cost by choosing a smaller value of $\gamma$, and hence, if $\gamma$ is not constrained, it must be that $\gamma \rightarrow 0$ if we wish to minimize (\ref{eq:objective_general2}).  Despite this necessary optimality condition, in existing practical VAE applications, it is standard to fix $\gamma \approx 1$ during training. This is equivalent to simply adopting a non-adaptive squared-error loss for the decoder and, at least in part, likely contributes to unrealistic/blurry VAE-generated samples \citep{bousquetetal2017}.   Regardless, there are more significant consequences of this intrinsic favoritism for $\gamma \rightarrow 0$, in particular as related to reconstructing data drawn from the ground-truth manifold $\bchi$:

\begin{theorem}\label{thm:decoder_mean}
Applying the same conditions and definitions as in Theorem~\ref{thm:decoder_variance}, then for all $\bx$ drawn from $\mu_{gt}$, we also have that
	\begin{equation} \label{eq:perfect_reconstruction}
		\lim_{\gamma\to0} f_{\mu_x} \left[ f_{\mu_z}(\bx;\phi_\gamma^*) +  f_{S_z}(\bx;\phi_\gamma^*) \bvarepsilon ;~ \theta_\gamma^* \right] = \lim_{\gamma\to0} f_{\mu_x} \left[ f_{\mu_z}(\bx;\phi_\gamma^*); ~\theta_\gamma^* \right] = \bx, ~~~~ \forall \bvarepsilon \in \mathbb{R}^{\kappa}.
	\end{equation}
\end{theorem}


By design any random draw $\bz \sim q_{\phi_\gamma^*}\left(\bz|\bx \right)$ can be expressed as $f_{\mu_z}(\bx;\phi_\gamma^*) + f_{S_z}(\bx;\phi_\gamma^*) \bvarepsilon$ for some $\bvarepsilon \sim \calN(\bvarepsilon|{\bf 0},\bI)$.  From this vantage point then, (\ref{eq:perfect_reconstruction}) effectively indicates that any $\bx \in \bchi$ will be perfectly reconstructed by the VAE encoder/decoder pair at globally optimal solutions, achieving this necessary condition despite any possible stochastic corrupting factor $f_{S_z}(\bx;\phi_\gamma^*) \bvarepsilon$.

But still further insights can be obtained when we more closely inspect the VAE objective function behavior at arbitrarily small but explicitly nonzero values of $\gamma$.  In particular, when $\kappa = r$ (meaning $\bz$ has no superfluous capacity), Theorem \ref{thm:decoder_mean} and attendant analyses in the appendices ultimately imply that the squared eigenvalues of $f_{S_z}(\bx;\phi_\gamma^*)$ will become arbitrarily small at a rate proportional to $\gamma$, meaning $\tfrac{1}{\sqrt{\gamma}}f_{S_z}(\bx;\phi_\gamma^*) \approx O(1)$ under mild conditions.  It then follows that the VAE data term integrand from (\ref{eq:objective_general2}), in the neighborhood around optimal solutions, behaves as ~~~ $- 2\mathbb{E}_{q_{\phi_\gamma^*}\left(\bz|\bx \right)} \left[\log p_{ \theta_\gamma^*} \left(\bx | \bz  \right)  \right] ~ =$

\begin{equation} \label{eq:data_term_reduction}
2\mathbb{E}_{q_{\phi_\gamma^*}\left(\bz|\bx \right)} \left[ \tfrac{1}{\gamma} \left\| \bx - f_{\mu_x} \left[ \bz; ~\theta_\gamma^* \right] \right\|_2^2  \right] + d \log 2\pi\gamma \approx \mathbb{E}_{q_{\phi_\gamma^*}\left(\bz|\bx \right)} \left[ O(1)  \right] + d \log 2\pi\gamma =  d \log \gamma + O(1).
\end{equation}
This expression can be derived by excluding the higher-order terms of a Taylor series approximation of $f_{\mu_x} \left[ f_{\mu_z}(\bx;\phi_\gamma^*) +  f_{S_z}(\bx;\phi_\gamma^*) \bvarepsilon ;~ \theta_\gamma^* \right]$ around the point $f_{\mu_z}(\bx;\phi_\gamma^*)$, which will be relatively tight under the stated conditions. But because $2\mathbb{E}_{q_{\phi_\gamma^*}\left(\bz|\bx \right)} \left[ \tfrac{1}{\gamma} \left\| \bx - f_{\mu_x} \left[ \bz; ~\theta_\gamma^* \right] \right\|_2^2  \right] \geq 0$, a theoretical lower bound on (\ref{eq:data_term_reduction}) is given by $d \log 2\pi\gamma \equiv d \log \gamma + O(1)$.  So in this sense (\ref{eq:data_term_reduction}) cannot be significantly lowered further.


This observation is significant when we consider the inclusion of addition latent dimensions by allowing $\kappa > r$.  Clearly based on the analysis above, adding dimensions to $\bz$ \emph{cannot} improve the value of the VAE data term in any meaningful way.  However, it can have a detrimental impact on the the KL regularization factor in the $\gamma \rightarrow 0$ regime, where
\begin{equation} \label{eq:kl_term}
2\mathbb{KL}\left[ q_\phi(\bz|\bx) || p(\bz) \right] \equiv \mbox{trace}\left[\bSigma_z \right] + \|\bmu_z \|_2^2 - \log | \bSigma_z| \approx -\hat{r} \log \gamma + O(1).
\end{equation}
Here $\hat{r}$ denotes the number of eigenvalues $\{ \lambda_j(\gamma) \}_{j=1}^{\kappa}$ of $f_{S_z}(\bx;\phi_\gamma^*)$ (or equivalently $\bSigma_z$) that satisfy $\lambda_j(\gamma) \rightarrow 0$ if $\gamma \rightarrow 0$.  $\hat{r}$ can be viewed as an estimate of how many low-noise latent dimensions the VAE model is preserving to reconstruct $\bx$.  Based on (\ref{eq:kl_term}), there is obvious pressure to make $\hat{r}$ as small as possible, at least without disrupting the data fit.  The smallest possible value is $\hat{r} = r$, since it is not difficult to show that any value below this will contribute consequential reconstruction errors, causing $2\mathbb{E}_{q_{\phi_\gamma^*}\left(\bz|\bx \right)} \left[ \tfrac{1}{\gamma} \left\| \bx - f_{\mu_x} \left[ \bz; ~\theta_\gamma^* \right] \right\|_2^2  \right]$ to grow at a rate of $\Omega\left(\tfrac{1}{\gamma}\right)$, pushing the entire cost function towards infinity.\footnote{Note that $\inf_{\gamma > 0} \tfrac{C}{\gamma} + \log \gamma = \infty$ for any $C >0$.}

Therefore, \emph{in the neighborhood of optimal solutions the VAE will naturally seek to produce perfect reconstructions using the fewest number of clean, low-noise latent dimensions}, meaning dimensions whereby $q_{\phi}\left(\bz|\bx \right)$ has negligible variance.  For superfluous dimensions that are unnecessary for representing $\bx$, the associated encoder variance in these directions can be pushed to one.  This will optimize $\mathbb{KL}\left[ q_\phi(\bz|\bx) || p(\bz) \right]$ along these directions, and the decoder can selectively block the residual randomness to avoid influencing the reconstructions per Theorem \ref{thm:decoder_mean}.  So in this sense the VAE is capable of learning a minimal representation of the ground-truth manifold $\bchi$ when $r < \kappa$.

But we must emphasize that the VAE can learn $\bchi$ independently of the actual distribution $\mu_{gt}$ within $\bchi$.  Addressing the latter is a completely separate issue from achieving the perfect reconstruction error defined by Theorem \ref{thm:decoder_mean}.  This fact can be understood within the context of a traditional PCA-like model, which is perfectly capable of learning a low-dimensional subspace containing some training data without actually learning the distribution of the data within this subspace.  The central issue is that there exists an intrinsic bias associated with the VAE objective such that \emph{fitting the distribution within the manifold will be completely neglected whenever there exists the chance for even an infinitesimally better approximation of the manifold itself}.

Stated differently, if VAE model parameters have learned a near optimal, parsimonious latent mapping onto $\bchi$ using $\gamma \approx 0$, then the VAE cost will scale as $(d-r)\log \gamma$ regardless of $\mu_{gt}$.  Hence there remains a \emph{huge} incentive to reduce the reconstruction error still further, allowing $\gamma$ to push even closer to zero and the cost closer to $-\infty$.  And if we constrain $\gamma$ to be sufficiently large so as to prevent this from happening, then we risk degrading/blurring the reconstructions and widening the gap between $q_\phi(\bz|\bx)$ and $p_\theta(\bz|\bx)$, which can also compromise estimation of $\mu_{gt}$.  Fortunately though, as will be discussed next there is a convenient way around this dilemma by exploiting the fact that this dominanting $(d-r)\log \gamma$ factor goes away when $d = r$.

\section{From Theory to Practical VAE Enhancements}\label{sec:model}

Sections \ref{sec:impact_gaussian_assumption} and \ref{sec:optima_property} have exposed a collection of VAE properties with useful diagnostic value in and of themselves.  But the practical utility of these results, beyond the underappreciated benefit of learning $\gamma$, warrant further exploration.    In this regard, suppose we wish to develop a generative model of high-dimensional data $\bx \in \bchi$ where unknown low-dimensional structure is significant (i.e., the $r<d$ case with $r$ unknown).  The results from Section \ref{sec:optima_property} indicate that the VAE can partially handle this situation by learning a parsimonious representation of low-dimensional manifolds, but not necessarily the correct probability measure $\mu_{gt}$ within such a manifold.  In quantitative terms, this means that a decoder $p_\theta(\bx|\bz)$ will map all samples from an encoder $q_\phi(\bz|\bx)$ to the correct manifold such that the reconstruction error is negligible for any $\bx \in \bchi$.  But if the measure $\mu_{gt}$ on $\bchi$ has not been accurately estimated, then
\begin{equation}
	\textstyle{q_\phi(\bz) \triangleq \int_{\bchi} q_\phi(\bz|\bx) \mu_{gt}(d\bx) \not\approx \int_{\mathbb{R}^d} p_\theta(\bz|\bx) p_\theta(\bx) d\bx = \int_{\mathbb{R}^d} p_\theta(\bx|\bz)p(\bz) d\bx = \mathcal{N}(\bz | {\bf 0}, \bI)}, \label{eqn:mismatch}
\end{equation}
where $q_\phi(\bz)$ is sometimes referred to as the aggregated posterior \citep{makhzani2016}.  In other words, the distribution of the latent samples drawn from the encoder distribution, when averaged across the training data, will have lingering latent structure that is errantly incongruous with the original isotropic Gaussian prior.  This then disrupts the pivotal ancestral sampling capability of the VAE, implying that samples drawn from $\mathcal{N}(\bz | 0, I)$ and then passed through the decoder $p_\theta(\bx|\bz)$ will \emph{not} closely approximate $\mu_{gt}$.  Fortunately, our analysis suggests the following two-stage remedy:
\begin{enumerate}
	\item Given $n$ observed samples $\{\bx^{(i)}\}_{i=1}^n$, train a $\kappa$-simple VAE, with $\kappa \geq r$, to estimate the unknown $r$-dimensional ground-truth manifold $\bchi$ embedded in $\mathbb{R}^d$ using a minimal number of active latent dimensions.  Generate latent samples $\{\bz^{(i)}\}_{i=1}^n$ via $\bz^{(i)} \sim q_\phi(\bz|\bx^{(i)})$.  By design, these samples will be distributed as $q_\phi(\bz)$, but likely not $\calN(\bz|{\bf 0},\bI)$.
	\item Train a second $\kappa$-simple VAE, with independent parameters $\{\theta',\phi'\}$ and latent representation $\bu$, to learn the unknown distribution $q_\phi(\bz)$, i.e., treat $q_\phi(\bz)$ as a new ground-truth distribution and use samples $\{\bz^{(i)}\}_{i=1}^n$ to learn it.
\item Samples approximating the \emph{original} ground-truth $\mu_{gt}$ can then be formed via the extended ancestral process $\bu \sim \mathcal{N}(\bu | {\bf 0}, \bI)$, $\bz \sim  p_{\theta'}(\bz|\bu)$, and finally $\bx \sim p_{\theta}(\bx|\bz)$.
\end{enumerate}

The efficacy of the second-stage VAE from above is based on the following.  If the first stage was successful, then even though they will not generally resemble $\calN(\bz|{\bf 0},\bI)$, samples from $q_\phi(\bz)$ will nonetheless have nonzero measure across the full ambient space $\mathbb{R}^{\kappa}$.  If $\kappa = r$, this occurs because the entire latent space is needed to represent an $r$-dimensional manifold, and if $\kappa > r$, then the extra latent dimensions will be naturally filled in via randomness introduced along dimensions associated with nonzero eigenvalues of the decoder covariance $\bSigma_z$ per the analysis in Section \ref{sec:optima_property}.

Consequently, as long as we set $\kappa \geq r$,  \emph{the operational regime of the second-stage VAE is effectively equivalent to the situation described in Section \ref{sec:r_equals_d_case} where the manifold dimension is equal to the ambient dimension}.\footnote{Note that if a regular autoencoder were used to replace the first-stage VAE, then this would no longer be the case, so indeed a VAE is required for both stages unless we have strong prior knowledge such that we may confidently set $\kappa \approx r$.}  And as we have already shown there via Theorem \ref{thm:optima_r_eq_d}, the VAE can readily handle this situation, since in the narrow context of the second-stage VAE, $d=r=\kappa$, the troublesome $(d-r)\log \gamma$ factor becomes zero, and any globally minimizing solution is uniquely matched to the new ground-truth distribution $q_\phi(\bz)$.  Consequently, the revised aggregated posterior $q_{\phi'}(\bu)$ produced by the second-stage VAE should now closely resemble $\calN(\bu|{\bf 0},\bI)$.  And importantly, because we generally assume that $d \gg \kappa \geq r$, we have found that the second-stage VAE can be quite small.

It should also be emphasized that concatenating the two VAE stages and jointly training does \emph{not} generally improve the performance.  If trained jointly the few extra second-stage parameters can simply be hijacked by the dominant influence of the first stage reconstruction term and forced to work on an incrementally better fit of the manifold rather than addressing the critical mismatch between $q_\phi(\bz)$ and $\calN(\bu|{\bf 0},\bI)$.  This observation can be empirically tested, which we have done in multiple ways.  For example, we have tried fusing the respective encoders and decoders from the first and second stages to train what amounts to a slightly more complex single VAE model.  We have also tried merging the two stages including the associated penalty terms.  In both cases, joint training does not help at all as expected, with average performance no better than the first stage VAE (which contains the vast majority of parameters).  Consequently, although perhaps counterintuitive, separate training of these two VAE stages is actually critical to achieving high quality results as will be demonstrated next.


\section{Empirical Evaluation of VAE Two-Stage Enhancement}\label{sec:experiments}

In this section we first present quantitative evaluations of the proposed two-stage VAE against various GAN and VAE baselines.  We then describe experiments explicitly designed to corroborate our theoretical findings from Sections \ref{sec:impact_gaussian_assumption} and \ref{sec:optima_property}, including a demonstration of robustness to the choice of $\kappa$.  Finally, we include representative samples generated from our model.

\subsection{Quantitative Comparisons of Generated Sample Quality}

We first present quantitative evaluation of novel generated samples using the large-scale testing protocol of GAN models from \citep{lucic2018gans}.  In this regard, GANs are well-known to dramatically outperform existing VAE approaches in terms of the Fr\'echet Inception Distance (FID) score \citep{heusel2017gans} and related quantitative metrics.  For fair comparison, \citep{lucic2018gans} adopted a common neutral architecture for all models, with generator and discriminator networks based on InfoGAN \citep{chen2016infogan}; the point here is standardized comparisons, not tuning arbitrarily-large networks to achieve the lowest possible absolute FID values.  We applied the same architecture to our first-stage VAE decoder and encoder networks respectively for direct comparison.  For the low-dimensional second-stage VAE we used small, 3-layer networks contributing negligible additional parameters beyond the first stage (see the appendices for further design details).

We evaluated our proposed VAE pipeline, henceforth denoted as \emph{2-Stage VAE}, against three baseline VAE models differing only in the decoder output layer: a Gaussian layer with fixed $\gamma$, a Gaussian layer with a learned $\gamma$, and a cross-entropy layer as has been adopted in several previous applications involving images~\citep{chen2016variational}.  We also tested the Gaussian decoder VAE model (with learned $\gamma$) combined with an encoder augmented with normalizing flows~\citep{rezende2015variational}, as well as the recently proposed Wasserstein autoencoder (WAE)~\citep{tolstikhin2018wasserstein} which maintains a VAE-like structure.  All of these models were adapted to use the same neutral architecture from \citep{lucic2018gans}.  Note also that the WAE includes two variants, referred to as WAE-MMD and WAE-GAN because different Maximum Mean Discrepancy (MMD) and GAN regularization factors are involved.  We conduct experiments using the former because it does not involve potentially-unstable adversarial training, consistent with the other VAE baselines.\footnote{Later we compare against both WAE-MMD and WAE-GAN using the setup from \citep{tolstikhin2018wasserstein}.}  Additionally, we present results from \citep{lucic2018gans} involving numerous competing GAN models, including MM GAN~\citep{goodfellow2014generative}, WGAN~\citep{arjovsky2017wasserstein}, WGAN-GP~\citep{gulrajani2017improved}, NS GAN~\citep{fedus2017many}, DRAGAN~\citep{kodali2017convergence}, LS GAN~\citep{mao2017least} and BEGAN~\citep{berthelot2017began}.   Testing is conducted across four significantly different datasets: MNIST~\citep{lecun1998gradient}, Fashion MNIST~\citep{xiao2017/online}, CIFAR-10~\citep{krizhevsky2009learning} and CelebA~\citep{liu2015deep}.


For each dataset we executed $10$ independent trials and report the mean and standard deviation of the FID scores in Table~\ref{table:fid}.\footnote{All reported FID scores for VAE and GAN models were computed using TensorFlow (\url{https://github.com/bioinf-jku/TTUR}).  We have found that alternative PyTorch implementations (\url{https://github.com/mseitzer/pytorch-fid}) can produce different values in some circumstances.  This seems to be due, at least in part, to subtle differences in the underlying inception models being used for computing the scores.  Either way, a consistent implementation is essential for calibrating results across different scenarios.}  No effort was made to tune VAE training hyperparameters (e.g., learning rates, etc.); rather a single generic setting was first agnostically selected and then applied to all VAE-like models (including the WAE-MMD).  As an analogous baseline, we also report the value of the best GAN model (in terms of average FID across all four datasets) when trained without data-dependent tuning using suggested settings from the authors; see \citep{lucic2018gans}[Figure 4] for details.  Under these conditions, our single 2-Stage VAE is quite competitive with the best individual GAN model, while the other VAE baselines were not competitive.

Note that the relatively poor performance of the WAE-MMD on MNIST and Fashion MNIST data can be attributed to the sensitivity of this approach to the value of $\kappa$, which for consistency with other models was fixed at $\kappa = 64$ for all experiments.  This value is likely much larger than actually needed for these simpler data types (meaning $r \ll 64$), and the WAE-MMD model can potentially be more reliant on having some $\kappa \approx r$.  We will return to this issue in Section \ref{sec:testing_theory} below, where among other things, we empirically examine how performance varies with $\kappa$ across different modeling paradigms.

%

Table~\ref{table:fid} also displays FID scores from GAN models evaluated using hyperparameters obtained from a large-scale search executed independently across each dataset to achieve the best results; 100 settings per model per dataset, plus an optimal, data-dependent stopping criteria as described in \citep{lucic2018gans}.  Within this broader paradigm, cases of severe mode collapse were omitted when computing final GAN FID averages.  Despite these considerable GAN-specific advantages, the FID performance of the default 2-Stage VAE is within the range of the heavily-optimized GAN models for each dataset unlike the other VAE baselines.  Overall then, these results represent the first demonstration of a VAE pipeline capable of competing with GANs in the arena of generated sample quality using a shared/neutral architecture.

\begin{table}[t!]
	\begin{tabular}{c|c|cccc}
\hline
&  & MNIST & Fashion & CIFAR-10 & CelebA  \\
\hline
	\hline
	& MM GAN & $9.8\pm0.9$ & $29.6\pm1.6$ & $72.7\pm3.6$ & $65.6\pm4.2$ \\
	& NS GAN & $6.8\pm0.5$ & $26.5\pm1.6$ & $58.5\pm1.9$ & $55.0\pm3.3$  \\
optimized,	& LSGAN & $7.8\pm0.6$ & $30.7\pm2.2$ & $87.1\pm47.5$ & $53.9\pm2.8$  \\
data-dependent	& WGAN & $6.7\pm0.4$ & $21.5\pm1.6$ & $55.2\pm2.3$ & $41.3\pm2.0$   \\
	settings & WGAN GP & $20.3\pm5.0$ & $24.5\pm2.1$ & $55.8\pm0.9$ & $30.3\pm1.0$  \\
	& DRAGAN & $7.6\pm0.4$ & $27.7\pm1.2$ & $69.8\pm2.0$ & $42.3\pm3.0$  \\
	& BEGAN & $13.1\pm1.0$ & $22.9\pm0.9$ & $71.4\pm1.6$ & $38.9\pm0.9$  \\
	\hline\hline
& Best default GAN & $\sim$ 10 & $\sim$ 32 & $\sim$ 70 & $\sim 65$ \\
& VAE (cross-entr.) & $16.6\pm0.4$ & $43.6\pm0.7$ & $106.0\pm1.0$ & $53.3\pm0.6$ \\
default	& VAE (fixed $\gamma$) & $52.0\pm0.6$ & $84.6\pm0.9$ & $160.5\pm1.1$ & $55.9\pm0.6$  \\
settings	& VAE (learned $\gamma$) & $54.5\pm1.0$ & $60.0\pm1.1$ & $76.7\pm0.8$ & $60.5\pm0.6$ \\
& VAE + Flow & $54.8\pm2.8$ & $62.1\pm1.6$ & $81.2\pm2.0$ & $65.7\pm2.8$ \\
& WAE-MMD & $115.0\pm1.1$ & $101.7\pm0.8$ & $80.9\pm0.4$ & $62.9\pm0.8$ \\
& 2-Stage VAE (ours) & $12.6\pm1.5$ & $29.3\pm1.0$ & $72.9\pm0.9$ & $44.4\pm0.7$ \\
	\hline
	\hline
	\end{tabular}
\vspace*{0.3cm}
	\caption{\emph{FID score comparisons}.  For all GAN-based models listed in the top section of the table, reported values represent the optimal FID obtained across a large-scale hyperparameter search conducted separately for each dataset \citep{lucic2018gans}.  Outlier cases (e.g., severe mode collapse) were omitted, which would have otherwise increased these GAN FID scores.  In the lower section of the table, the label \emph{Best default GAN} indicates scores produced by the GAN model with the lowest FID (averaged across datasets) when trained using settings suggested by original authors; these approximate values were extracted from \citep{lucic2018gans}[Figure 4].  For all VAE results (including WAE), only a single default setting was adopted across all datasets and models (no tuning whatsoever), and no cases of mode collapse were removed.  Note that specialized architectures and/or random seed optimization can potentially improve the FID score for all models reported here.  Additionally, the relatively high FID scores obtained using the WAE-MMD model for MNIST and Fashion data is explained at the end of Section \ref{sec:testing_theory} where robustness to latent-space dimensionality is discussed.  And as a final caveat, it has recently come to our attention that the pre-processing of CelebA data used by \citep{lucic2018gans} involved a cropping step that appears to be a bit different than we used for producing VAE/WAE-based results in this table (please see the respective github implementations; \citep{lucic2018gans} itself does not discuss CelebA cropping).  Note though that the CelebA FID results from Table~\ref{table:fid_wae} and Figure~\ref{fig:fid_vs_latent} below all involve the separate protocol from \citep{tolstikhin2018wasserstein}, including a consistent cropping step across all methods.}
	\label{table:fid}
\end{table}

Of course there are still other ways of evaluating sample quality.  For example, although the FID score is a widely-adopted measure that significantly improves upon the earlier Inception Score \citep{salimans2016improved}, it has been shown to exhibit bias in certain circumstances~\citep{binkowski2018demystifying}. To address this issue, the recently-proposed Kernel Inception Distance (KID) applies a polynomial-kernel MMD measure to estimate the inception distance and is believed to enjoy better statistical properties \citep{binkowski2018demystifying}.  Note that we cannot evaluate all of the GAN baselines using the KID score; only the authors of (Lucic et al., 2018) could easily do this given the huge number of trained models involved that are not publicly available, and the need to retrain selected models multiple times to produce new average scores at optimal hyperparameter settings.  However, we can at least compare our trained 2-Stage VAE to other VAE/AE-based models.  Table~\ref{table:kid} presents these results, where we observe that the same improvement patterns reported with respect to FID are preserved when we apply KID instead, providing further confidence in our approach.

\begin{table}[t!]
\centering
\begin{tabular}{c|cccc}
\hline
& MNIST & Fashion & CIFAR-10 & CelebA \\
\hline
VAE (cross-entr.) & $10.5\pm0.3$ & $37.0\pm0.9$ & $89.1\pm1.3$ & $48.4\pm0.4$ \\
VAE (fixed $\gamma$) & $42.7\pm0.8$ & $54.9\pm0.8$ & $153.4\pm1.8$ & $53.9\pm0.7$ \\
VAE (learned $\gamma$) & $51.5\pm1.2$ & $62.8\pm1.7$ & $64.6\pm0.5$ & $63.6\pm1.1$ \\
VAE + Flow & $56.0\pm3.8$ & $66.9\pm1.6$ & $68.5\pm3.0$ & $67.2\pm3.4$ \\
WAE-MMD & $137.8\pm1.7$ & $107.3\pm1.5$ & $58.7\pm0.5$ & $59.7\pm0.8$ \\
2-Stage VAE (ours) & $6.7\pm0.3$ & $25.9\pm1.6$ & $59.3\pm0.9$ & $40.9\pm0.5$ \\
\hline
\end{tabular}
\caption{\emph{KID score comparisons} (corresponding values for GAN baselines are not presently available).}
\label{table:kid}
\end{table}

Thus far we have presented performance evaluations of VAE models without any special tuning on a neutral testing platform; however, we have not as of yet compared our approach against state-of-the-art VAE-like competition benefitting from an encoder-decoder architecture and training protocol adjusted for a specific dataset.  To the best of our knowledge, the WAE model involves the only published evaluation of this kind (at least at the time of our original posting of this work), where different architectures are designed for use with MNIST and CelebA datasets \citep{tolstikhin2018wasserstein}.  However, because actual FID scores are only reported for CelebA (which is far more complex than MNIST anyway), we focus our attention on head-to-head testing with this data.  In particular, we adopt the exact same encoder-decoder networks as the WAE models, and train using the same number of epochs.  We do not tune any hyperparameters whatsoever, and apply the same small second-stage VAE as used in previous experiments.  As before, the second-stage size is a small fraction of the first stage, so any benefit is not simply the consequence of a larger network structure.   Results are reported in Table~\ref{table:fid_wae}, where the 2-Stage VAE even outperforms the WAE-GAN model, which has the advantage of adversarial training tuned for this combination of data and network architecture.




\begin{table}[t!]
\centering
\begin{tabular}{c|cccc}
\hline
 & VAE & WAE-MMD & WAE-GAN & 2-Stage VAE (ours) \\
\hline
CelebA FID & 63 & 55 & 42 & 34 \\
\hline
\end{tabular}
\caption{FID scores on CelebA data obtained using the network structure and training protocol from \citep{tolstikhin2018wasserstein}. For the 2-Stage VAE, we apply the exact same architecture and training epochs without any tuning of hyperparameters.}
\label{table:fid_wae}
\end{table}


\subsection{Experimental Corroboration of Theoretical Results} \label{sec:testing_theory}
The true test of any theoretical contribution is the degree to which it leads to useful, empirically-testable predictions about behavior in real-world settings. In the present context, although our theory from Sections \ref{sec:impact_gaussian_assumption} and \ref{sec:optima_property} involves some unavoidable simplifying assumptions, it nonetheless makes predictions that can be tested under practically-relevant conditions where these assumptions may not strictly hold.  We now present the results of such tests, which provide strong confirmation of our previous analysis.  In particular, after providing validation of Theorems \ref{thm:decoder_variance} and \ref{thm:decoder_mean}, we explicitly demonstrate that the second stage of our 2-Stage VAE model can reduce the gap between $q(\bz)$ and $p(\bz)$, and that the combined VAE stages are quite robust to the value of $\kappa$ as predicted.

\begin{figure}
\centering
\begin{subfigure}[t]{0.35\textwidth}
    \centering
    \includegraphics[width=1\linewidth]{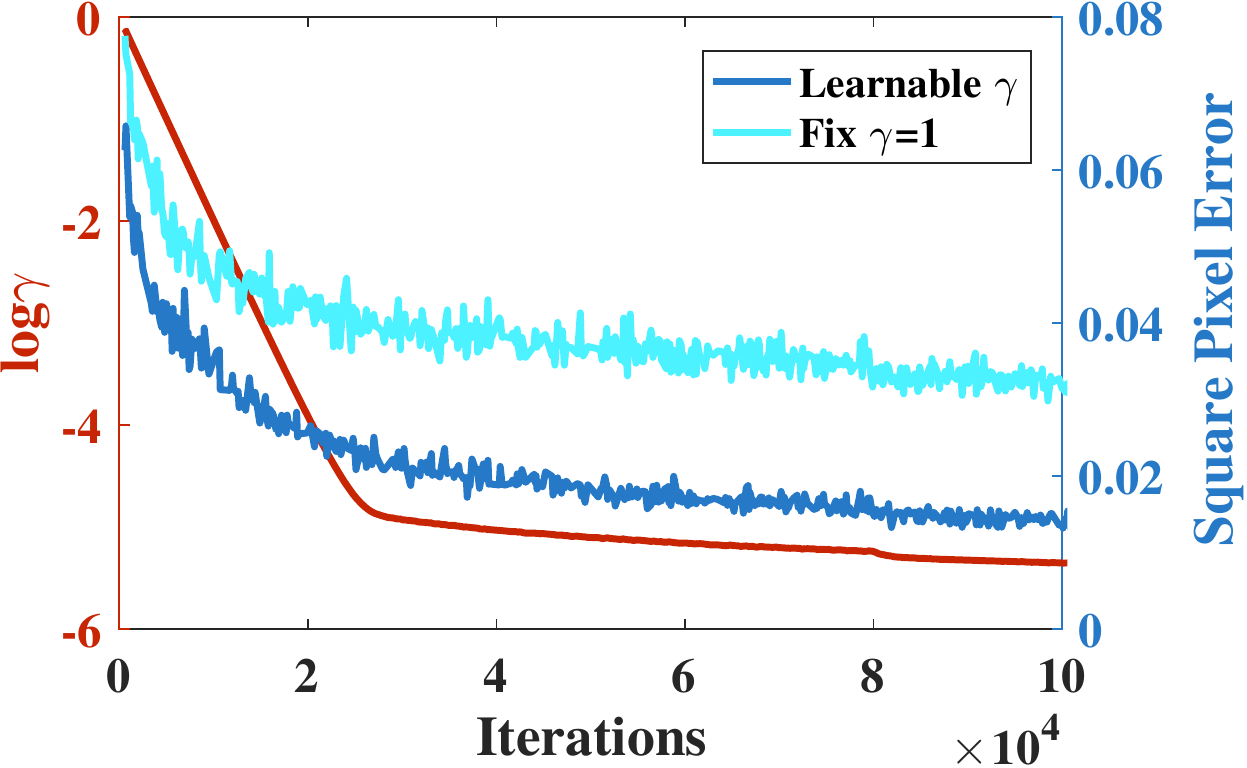}
    \caption{$\log\gamma$ and Reconstruction Error.}
    \label{fig:gamma_and_l2}
\end{subfigure}
\hspace{0.5cm}
\begin{subfigure}[t]{0.42\textwidth}
	\centering
	\includegraphics[width=1\linewidth]{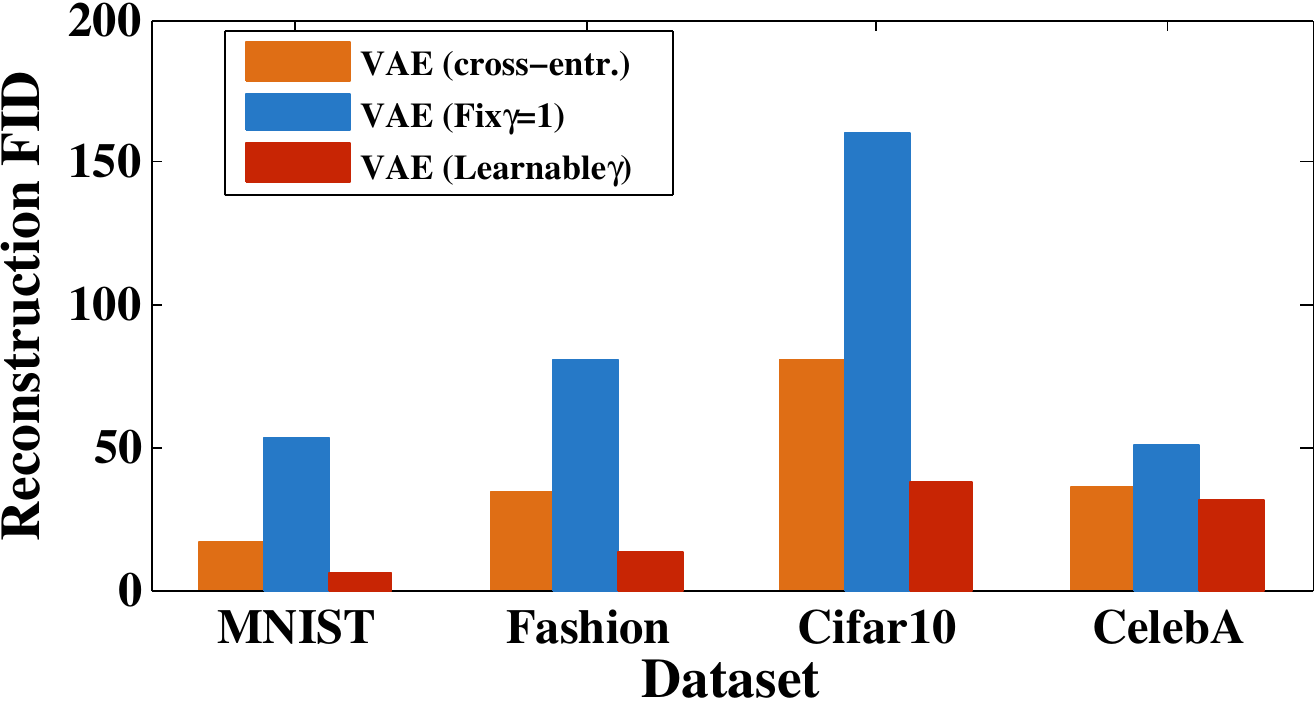}
	\caption{Reconstruction FID.}
	\label{fig:recon_fid}
\end{subfigure}
\caption{Validation of Theorem \ref{thm:decoder_variance}. (\emph{a}) The red line shows the evolution of $\log\gamma$, converging close to $0$ during training as expected. The two blue curves compare the associated pixel-wise reconstruction errors with $\gamma$ fixed at $1$ and with a learnable $\gamma$ respectively. (\emph{b}) The FID score obtained using \emph{reconstructed} images from various VAE models (reconstructed image FID is another way of evaluating reconstruction quality; it is distinct from measuring generated sample quality via FID scores).  In general, the VAE with learnable $\gamma$ produces the best reconstructions as expected.}
\label{fig:decoder_variance}
\end{figure}

\vspace*{0.4cm}
\noindent{\textbf{Validation of Theorem~\ref{thm:decoder_variance}:}} This theorem implies that $\gamma$ will converge to zero at any global minimum of the stated VAE objective under consideration. Figure~\ref{fig:gamma_and_l2} presents empirical support for this result, where indeed the decoder variance $\gamma$ does tend towards zero during training (red line).  This then allows for tighter image reconstructions (dark blue curve) with lower average squared error, i.e., a better manifold fit as expected.  Additionally, Figure~\ref{fig:recon_fid} compares the FID score computed using reconstructed training images from different VAE models (these are not new generated samples). The VAE with a learnable $\gamma$ achieves the lowest FID on all four datasets, implying that it also produces more realistic overall reconstructions as $\gamma$ goes to zero.  As will be discussed further in Section \ref{sec:discussion}, high-quality image reconstructions are half the battle in ultimately achieving good generative modeling performance.

\begin{figure}
\centering
\includegraphics[width=0.9\linewidth]{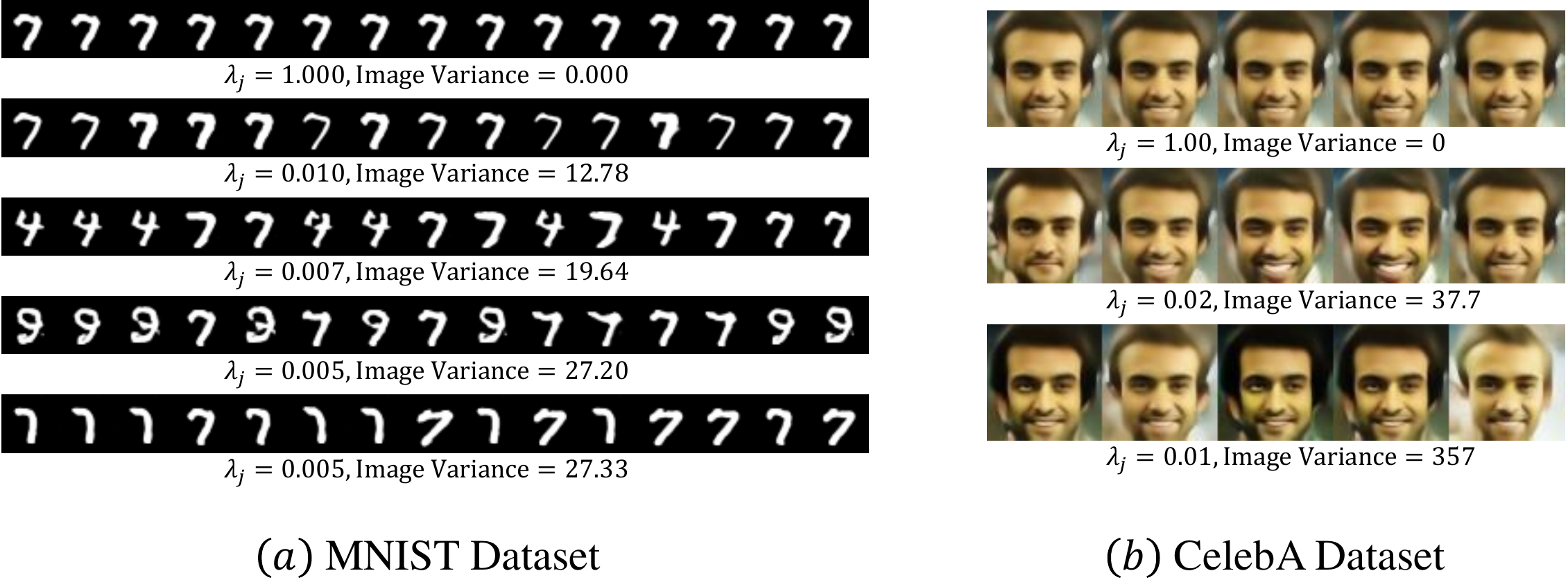}
\vspace{-0.2cm}
\caption{Validation of Theorem~\ref{thm:decoder_mean}. The $j$-th eigenvalue of $\bSigma_z$, denoted $\lambda_j$, should be very close to either $0$ or $1$ as argued in Section \ref{sec:optima_property}. When $\lambda_j$ is close to $0$, injecting noise along the corresponding direction will cause a large variance in the reconstructed image, meaning this direction is an informative one needed for representing the manifold. In contrast, if $\lambda_j$ is close to $1$, the addition of noise does not make any appreciable difference in the reconstructed image, indicating that the corresponding dimension is a superfluous one that has been ignored/blocked by the decoder.}
\label{fig:decoder_mean_main}
\end{figure}

\vspace*{0.4cm}
\noindent{\textbf{Validation of Theorem~\ref{thm:decoder_mean}:}} Figure~\ref{fig:decoder_mean_main} bolsters this theorem, and the attendant analysis which follows in Section \ref{sec:optima_property}, by showcasing the dissimilar impact of noise factors applied to different directions in the latent space before passage through the decoder mean network $f_{\mu_x}$. In a direction where an eigenvalue $\lambda_j$ of $\bSigma_z$ is large (i.e., a superfluous dimension), a random perturbation is completely muted by the decoder as predicted.  In contrast, in directions where such eigenvalues are small (i.e., needed for representing the manifold), varying the input causes large changes in the image space reflecting reasonable movement along the correct manifold.

\begin{figure}
\begin{floatrow}
\ffigbox[0.5\textwidth]{
	\includegraphics[width=0.65\linewidth]{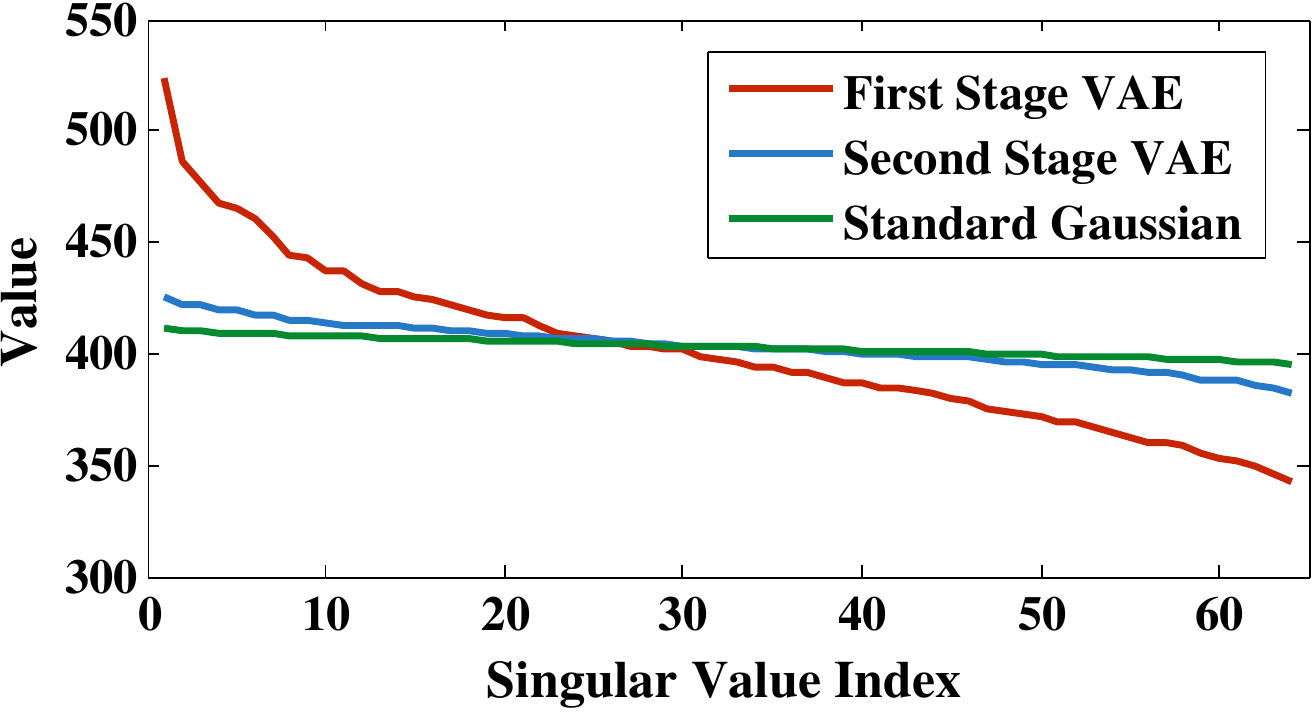}
}{
	\caption{Singular value spectrums of latent sample matrices drawn from $q_\phi(\bz)$ (first stage) and $q_{\phi^\prime}(\bu)$ (enhanced second stage).}\label{fig:svd}
}
\hspace*{0.3cm}
\capbtabbox{%
	\begin{tabular}{c|cc}
	\hline
	& First Stage & Second Stage \\
	\hline
	MNIST & $2.85$ & $0.43$ \\
	Fashion & $1.37$ & $0.40$ \\
	Cifar10 & $1.08$ & $0.00$ \\
	CelabA & $7.42$ & $0.29$ \\
	\hline
	\end{tabular}
}{
  \caption{Maximum mean discrepancy between $\mathcal{N}(0,\bI)$ and $q_\phi(\bz)$ (first stage); likewise for $q_{\phi^\prime}(\bu)$ (second stage).}\label{table:mmd}
}
\end{floatrow}
\end{figure}

\vspace*{0.4cm}
\noindent{\textbf{Reduced Mismatch between $q_{\phi}(\bz)$ and $p(\bz)$:}} Although the VAE with a learnable $\gamma$ can achieve high-quality reconstructions, the associated aggregated posterior is still likely not close to a standard Gaussian distribution as implied by (\ref{eqn:mismatch}).  This mismatch then disrupts the critical ancestral sampling process.  As we have previously argued, the proposed 2-Stage VAE has the ability to overcome this issue and achieve a standard Gaussian aggregated posterior, or at least nearly so.  As empirical evidence for this claim, Figure~\ref{fig:svd} displays the singular value spectrum of latent sample matrices $\bZ = \{ \bz^{(i)} \}_{i=1}^n$ drawn from $q_\phi(\bz)$ (first stage), and $\bU = \{ \bu^{(i)} \}_{i=1}^n$ drawn from $q_{\phi^\prime}(\bu)$ (enhanced second stage).  As expected, the latter is much closer to the spectrum from an analogous i.i.d. $\mathcal{N}(0,\bI)$ matrix. We also used these same sample matrices to estimate the MMD metric \citep{gretton2007kernel} between $\mathcal{N}(0,\bI)$ and the aggregated posterior distributions from the first and second stages in Table~\ref{table:mmd}.  Clearly the second stage has dramatically reduced the difference from $\mathcal{N}(0,\bI)$ as quantified by the MMD. Overall, these results indicate a superior latent representation, providing high-level support for our 2-Stage VAE proposal.

\vspace*{0.4cm}
\noindent{\textbf{Robustness to Latent-Space Dimensionality:}} According to the analysis from Section \ref{sec:optima_property}, the 2-Stage VAE should be relatively insensitive to having a good estimate of the ground-truth manifold dimension $r$.  As long as we choose $\kappa \geq r$, then our approach should in principle be able to compensate for any dimensionality mismatch.  To recap, the first VAE stage should fill in useless dimensions with random noise, such that $q_\phi(\bz)$ does \emph{not} lie on any lower-dimensional manifold.  The second stage then operates within the regime arbitrated by Theorem \ref{thm:optima_r_eq_d} such that we obtain a tractable means for sampling from $q_\phi(\bz)$.  This robustness to $\kappa$ need not be shared by alternative approaches, such as those predicated upon deterministic autoencoders or single-stage VAE models.  For example, both the WAE-MMD and WAE-GAN models are dependent on having a reasonable estimate for $\kappa \approx r$, at least for the deterministic encoder-decoder structures that were empirically tested in \citep{tolstikhin2018wasserstein}.  This is because, if $\kappa < r$, reconstructing the data on the manifold is not possible, and if $\kappa>r$, then $p(\bz) = \mathcal{N}(\bz | 0, I)$  and $q_\phi(\bz)$ cannot be matched since the latter is necessarily confined by the training data to an $r$-dimensional manifold within $\kappa$-dimensional space.  This likely explains the poor performance of WAE-MMD on MNIST and Fashion MNIST data reported in Table~\ref{table:fid}.

\begin{figure}
\centering
	\begin{subfigure}[t]{0.75\textwidth}
	\centering
	\includegraphics[width=1\linewidth]{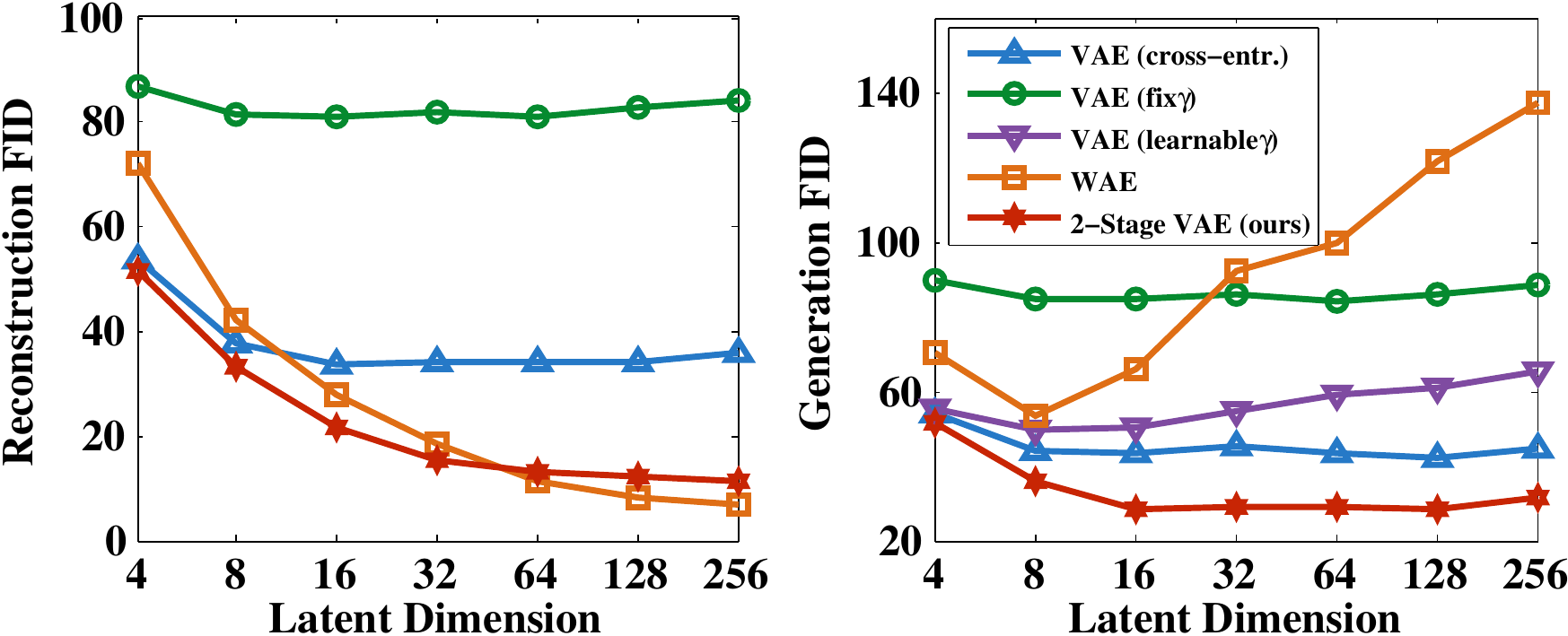}
	\caption{Fashion Dataset}
	\end{subfigure}
	\begin{subfigure}[t]{0.75\textwidth}
	\centering
	\includegraphics[width=1\linewidth]{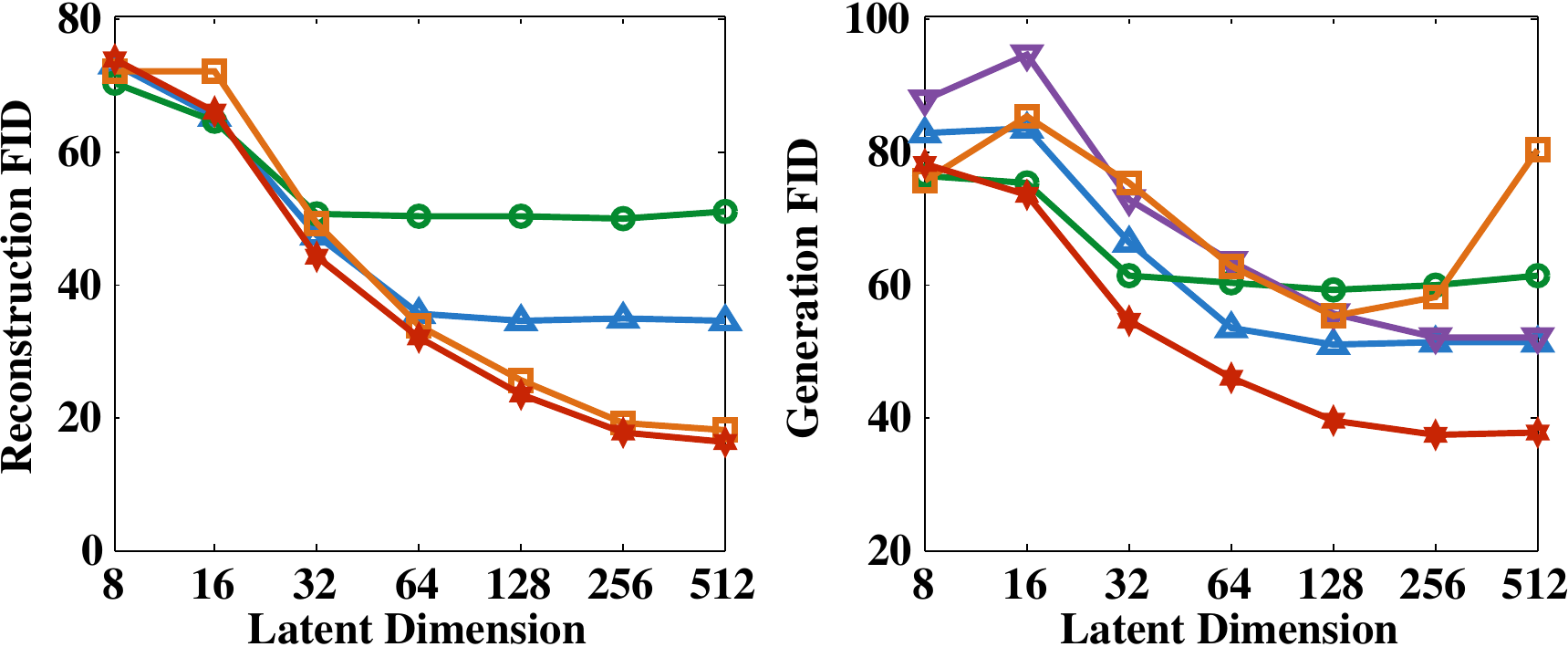}
	\caption{CelebA Dataset}
	\end{subfigure}
\caption{FID Score w.r.t. Different Latent Dimensions. \emph{(Left)} Reconstruction FID. \emph{(Right)} Generation FID.}
\label{fig:fid_vs_latent}
\end{figure}

To further probe these issues, we again adopt the neutral testing framework from \citep{lucic2018gans}, and retrain each model as $\kappa$ is varied.  We conduct this experiment using Fashion MNIST (relatively small/simple) and CelebA (more complex).  FID scores from both reconstructions of the training data, and novel generated samples are shown in Figure~\ref{fig:fid_vs_latent}.  From these plots (left column) we observe that both the 2-Stage VAE and WAE-MMD have a similar reconstruction FID values that become smaller as $\kappa$ increases (reconstructions should generally improve with increasing $\kappa$).  Note that the WAE-MMD relies on a Gaussian output layer with a small $\gamma$ value,\footnote{Unlike standard VAEs that are capable of self-calibration in some sense, with $\gamma$ and elements of $\bSigma_z$ jointly pushing towards small values, $\gamma$ may be difficult to learn for WAE models because there is also a required weighting factor balancing the MMD (or GAN) loss.  Additionally, the WAE code posted in association with \citep{tolstikhin2018wasserstein} does not attempt to learn $\gamma$.} and hence the reconstructions are unlikely to be significantly different from the 2-Stage VAE.  In contrast, the other baselines are considerably worse, especially when $\gamma = 1$ is fixed as is often done in practice.

The situation changes considerably however when we examine the FID scores obtained from generated samples.  While the VAE models remain relatively stable over a significant range of sufficiently large $\kappa$ values, some of which are likely to be considerably larger than necessary (e.g., $\kappa > 16$ on Fashion MNIST data), the WAE-MMD performance is much more sensitive.  Of course we readily concede that compensatory, data-dependent tuning of WAE-MMD hyperparameters might allow for some additional improvements in these curves, but this then further highlights the larger point:  the VAE models were not tuned in any way for these experiments and yet still maintain stable behavior, with our 2-Stage variant providing a sizeable advantage.

Of course obviously in practice if we set $\kappa$ to be far too large, then the training will likely become much more difficult, since in addition to learning the correct ground-truth manifold, we are also burdening the model to detect a much larger number of unnecessary dimensions.  But even so, the 2-Stage VAE is arguably quite robust to $\kappa$ within these experimental settings, and certainly we need not set $\kappa \approx r$ to achieve good results; a reasonable $\kappa \geq r$ appears to be sufficient.  Still, as a final caveat, we should mention that the VAE cannot automatically manage all forms of excess model capacity.  As discussed in \citep{dai2018jmlr}[Section 4], if the decoder mean network becomes inordinately complex/deep, then a useless degenerate solution involving the implicit memorization of training data can break even the natural VAE regularization mechanisms we have described herein (in principle, this can happen even with $\kappa = 1$ for a finite training dataset).  Obviously though, GAN models and virtually all other deep generative pipelines share a similar vulnerability.

\subsection{Qualitative Evaluation of Generated Samples}

Finally, we qualitatively evaluate samples generated via our 2-Stage VAE using a simple, convenient residual network structure (with fewer parameters than the InfoGAN architecture).  Details of this network are shown in the appendices.  Randomly generated samples from our 2-Stage VAE are shown in Figure~\ref{fig:gen_samples} for MNIST and CelebA data.  Additional samples can be found in the appendices.

\begin{figure}
\centering
\begin{subfigure}[t]{0.95\textwidth}
    \centering
    \includegraphics[width=1\linewidth]{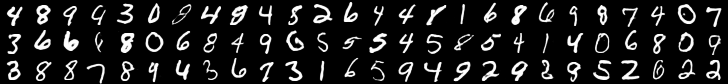}
    \caption{MNIST Data}
\end{subfigure}
\begin{subfigure}[t]{0.95\textwidth}
    \centering
    \includegraphics[width=1\linewidth]{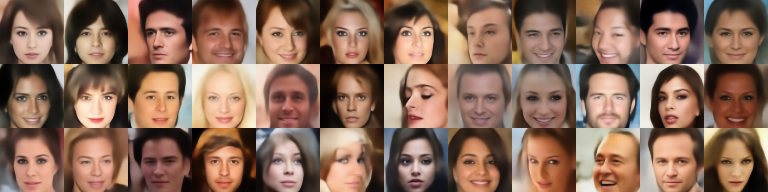}
    \caption{CelebA Data}
\end{subfigure}
\caption{Randomly Generated Samples (No cherry picking).}
\label{fig:gen_samples}
\end{figure}

\section{Discussion} \label{sec:discussion}
It is often assumed that there exists an unavoidable trade-off between the stable training, valuable attendant encoder network, and resistance to mode collapse of VAEs, versus the impressive visual quality of images produced by GANs.  While we certainly are not claiming that our two-stage VAE model is superior to the latest and greatest GAN-based architectures in terms of the realism of generated samples, we do strongly believe that this work at least narrows that gap substantially such that VAEs are worth considering in a broader range of applications.  We now close by situating our work within the context of existing VAE enhancements, as well as recent efforts to learn so-called disentangled representations.

\subsection{Connections with Existing VAE Enhancements} \label{sec:related_work}

Although a variety of alternative VAE renovations have been proposed, unlike our work, nearly all of these have focused on improving the log-likelihood scores assigned by the model to test data.  In particular, multiple elegant approaches involve replacing the Gaussian encoder network with a richer class of distributions instantiated through normalizing flows or related \citep{burda2015importance,kingma2016improved, rezende2015variational, van2018sylvester}.  While impressive log-likelihood gains have been demonstrated, this achievement is largely orthogonal to the goal of improving quantitative measures of visual quality \citep{Theis_ICLR2016}, which has been our focus herein.  Additionally, improving the VAE encoder does not address the uniqueness issue raised in Section \ref{sec:impact_gaussian_assumption}, and therefore, a second stage could potentially benefit these models too under the right circumstances.

Broadly speaking, if the overriding objective is generating realistic samples using an encoder-decoder-based architecture (VAE or otherwise), two important, well-known criteria must be satisfied:\footnote{In presenting these criteria, we are implicitly excluding models that employ powerful non-Gaussian decoders that can parameterize complex distributions even while excluding any signal from $\bz$, e.g., PixelCNN-based decoders or related \citep{van2016conditional}.  Such models are considerably different than a canonical AE and do not generally provide a way of accurately reconstructing the training data using a low-dimensional representation.}
\begin{enumerate}[label=(\roman*)]

\item Small reconstruction error when passing through the encoder-decoder networks, and

\item An aggregate posterior $q_\phi(\bz)$ that is close to some known distribution like $p(\bz) = \mathcal{N}(\bz | 0, I)$ that is easy to sample from.
\end{enumerate}
The first criteria can be naturally enforced by a deterministic AE, but also for the VAE as $\gamma$ becomes small as quantified by Theorem \ref{thm:decoder_mean}. Of course the second criteria is equally important.  Without it, we have no tractable way of generating random inputs that, when passed through the learned decoder, produce realistic output samples resembling the training data distribution.

Criteria (\emph{i}) and (\emph{ii}) can be addressed in multiple different ways.  For example, \citep{tomczak2018vae,zhao18} replace $\mathcal{N}(\bz | 0, I)$ with a parameterized class of prior distributions such that there exist more flexible pathways for pushing $p(\bz)$  and $q_\phi(\bz)$ closer together; a VAE-like objective is used for this purpose in \citep{tomczak2018vae}, while \citep{zhao18} employs adversarial training.  Consequently, even if $q_\phi(\bz)$ is not Gaussian, we can nonetheless sample from a known non-Gaussian alternative using either of these approaches.   This is certainly an interesting idea, but it has not as of yet been shown to improve FID scores.  For example, only log-likelihood values on relatively small black-and-white images are reported in \citep{tomczak2018vae}, while discrete data and associated evaluation metrics are the focus of \citep{zhao18}.

In fact, prior to our work the only competing encoder-decoder-based architecture that explicitly attempts to improve FID scores is the WAE model from \citep{tolstikhin2018wasserstein}, which can be viewed as a generalization of the adversarial autoencoder \citep{makhzani2016}.  For both WAE-MMD and WAE-GAN variants introduced and tested in Section \ref{sec:experiments} herein, the basic idea is to minimize an objective function composed of a reconstruction penalty for handling criteria (\emph{i}), and a Wassenstein distance measure between $p(\bz)$  and $q_\phi(\bz)$ (either MMD- or GAN-based) for addressing criteria (\emph{ii}).  Note that under the reported experimental design from \citep{tolstikhin2018wasserstein}, the WAE-GAN model more-or-less defaults to an adversarial autoencoder, although broader differentiating design choices are possible.  The adversarially-regularized autoencoder proposed in \citep{zhao18} can also be interpreted as a variant of the WAE-GAN customized to handle discrete data.




As with the approaches mentioned above, the two VAE stages we have proposed can also be motivated in one-to-one correspondence with criteria (\emph{i}) and (\emph{ii}). In brief, the first VAE stage addresses criteria (\emph{i}) by pushing both the encoder variance, and the decoder variances selectively, towards zero such that accurate reconstruction is possible using a minimal number of active latent dimensions.  However, our detailed analysis suggests that, although the resulting aggregate posterior $q_\phi(\bz)$ will occupy nonzero measure in $\kappa$-dimensional space (selectively filling out superfluous dimensions with random noise), it need not be close to $\mathcal{N}(\bz | 0, I)$.  This then implies that if we take samples from $\mathcal{N}(\bz | 0, I)$ and pass them through the learned decoder, the result may not closely resemble real data.


Of course if we could somehow directly sample from $q_\phi(\bz)$, then we would not need to use $\mathcal{N}(\bz | 0, I)$.  And fortunately, because the first-stage VAE ensures that $q_\phi(\bz)$ will satisfy the conditions of Theorem \ref{thm:optima_r_eq_d}, we know that a second VAE can in fact be learned to accurately sample from this distribution, which in turn addresses criteria (\emph{ii}).  Specifically, per the arguments from Section \ref{sec:model}, sampling $\bu \sim \mathcal{N}(\bu | {\bf 0}, \bI)$ and then $\bz \sim  p_{\theta'}(\bz|\bu)$ is akin to sampling $\bz \sim q_\phi(\bz)$ even though the latter is not available in closed form.  Such samples can then be passed through the first-stage VAE decoder to obtain samples of $\bx$.  Hence our framework provides a principled alternative to existing encoder-decoder structures designed to handle criteria (\emph{i}) and (\emph{ii}), leading to state-of-the-art results for this class of model in terms of FID scores under neutral testing conditions.


Finally, before closing this section, it is also informative to consider the vector-quantization VAE (VQ-VAE) from \citep{van2017neural}.  This model of discrete distributions involves first training a deterministic autoencoder with a latent space parameterized by a vector quantization operator.  Later, a PixelCNN \citep{van2016conditional} or related is trained to approximate the aggregated posterior of the corresponding discrete latent codes.  In this sense the VQ-VAE is effectively satisfying criteria (\emph{ii}), at least in the regime of discrete distributions.  However, in \citep{van2017neural} it is also mentioned that joint training of the VQ-VAE reconstruction module and the PixelCNN aggregated posterior estimate could improve performance.  While this may be true for discrete distributions (verification is left to future work in \citep{van2017neural}), at least for continuous data lying on a low-dimensional manifold as has been our focus, we reiterate that joint training of our two proposed continuous VAE stages is likely to degrade performance.  This is for the detailed reasons provided in Section \ref{sec:model}.  In this regard, there can seemingly exist non-trivial differences in the properties of generative models of discrete versus continuous data.

\subsection{Identifiability of Disentangled Representations}


Although definitions vary, if a latent $\bz$ is `disentangled' so to speak, then each element should ideally contain an interpretable, semantically-meaningful factor of variation that can act independently on the generative process \citep{chen2018isolating}.  For example, consider an MNIST image $\bx$ and a hypothetical latent representation $\bz = [z_1,\ldots,z_{\kappa}]^{\top} \in \mathbb{R}^{\kappa}$, whereby $z_1$ encodes the digit type (i.e., $0, 1, \ldots, 9$), $z_2$ is the stroke thickness, $z_3$ is a measure of the overall digit slant, and so on for $z_4, z_5, \ldots, z_{\kappa}$.  This $\bz$ is a canonical disentangled representation, with clearly interpretable elements that can be independently adjusted to generate samples with predictable differences, e.g., we may adjust $z_2$ while keeping other dimensions fixed to isolate variations in stroke thickness.

A number of VAE modifications have been putatively designed to encourage this type of disentanglement (likewise for other generative modeling paradigms).  These efforts can be partitioned into those that involve some degree of supervision, to both define and isolate specific ground-truth factors of variation (which may be application-specific), and those that aspire to be completely unsupervised.  For the latter, the VAE is typically refashioned to minimize, at least to the extent possible, some loose proxy for the degree of entanglement.  An influential example is the \emph{total correlation},\footnote{This naming convention is perhaps a misnomer, since total correlation measures statistical dependency beyond second-order correlations.} defined for the present context as
\begin{equation}
\mbox{TC} = \textstyle{\mbox{KL}\left[ q_\phi(\bz) || \prod_{j=1}^{\kappa} q_\phi(z_j) \right] \geq 0 },
\end{equation}
where $q_\phi(z_j)$ denotes the marginal distribution of $z_j$.  It follows that $\mbox{TC} = 0$ iff $q_\phi(\bz) = \prod_{j} q_\phi(z_j)$, meaning that the aggregate posterior distribution, which generates input samples to the VAE decoder, involves independent factors of variation.  It has been argued then that adjustments to the VAE objective that push the total correlation towards zero will favor desirable disentangled representations \citep{chen2018isolating,higgins2017}.  Note that if criteria (\emph{ii}) from Section \ref{sec:related_work} is satisfied with a factorial prior $p(\bz) = \prod_j p(z_j)$ such as $\mathcal{N}(\bz | 0, I)$, then we have actually already achieved $\mbox{TC} \approx 0$.  In contrast, models with more complex priors as used in \citep{tomczak2018vae} would require further alterations to enforce this goal.

While we agree that methods penalizing $\mbox{TC}$ (either directly or indirectly) will favor latent representations with independent dimensions, unfortunately this need not correspond with any semantically-meaningful form of disentanglement.  Intuitively, this means that independent latent factors need not correspond with interpretable concepts like stroke width or digit slope.  In fact, without at least some form of supervision or constraints on the space of possible representations, disentanglement defined with respect to any particular ground-truth semantic factors is not generally identifiable in the strict statistical sense.

As one easy way to see this, suppose that we have access to data generated as $\bx = f_{gt}(\bz)$, where $f_{gt}$ serves as an arbitrary ground-truth decoder and $\bz \sim \prod_j p_{gt}(z_j)$ represents ground-truth/disentangled latent factors of interest.  However, we could just as well generate identical data via $\bx = f_{gt}\left( \bD_1^{-1}\left[ \bR^{\top} \bD_2^{-1}\left( \widetilde{\bz} \right) \right] \right)$, where $\widetilde{\bz} \triangleq \bD_2 \left[ \bR \bD_1(\bz) \right]$ with associated factorial distribution $\prod_{j} \widetilde{p}(\widetilde{z}_j)$.  Here the operator $\bD_1$ converts $\bz$ to a standardized Gaussian distribution, $\bR$ is an arbitrary rotation matrix the mixes the factors but retains a factorial representation with $\mbox{TC} = 0$, and $\bD_2$ converts the resulting Gaussian to a new factorial distribution with arbitrary marginals.

By construction, each new latent factor $\widetilde{z}_j$ will be composed as a mixture of the original factors of interest, and yet they will also have zero total correlation, i.e., they will be statistically independent.  Therefore if we treat the composite operator $\widetilde{f}(\cdot) \triangleq f_{gt}\left( \bD_1^{-1}\left[ \bR^{\top} \bD_2^{-1}\left( \cdot \right) \right] \right)$ as the effective decoder, we have a new generative process $\bx = \widetilde{f}(\widetilde{\bz} )$ and $\widetilde{\bz} \sim  \prod_{j} \widetilde{p}(\widetilde{z}_j)$.  By construction, this process will produce identical observed data using a latent representation with zero total correlation, but with highly entangled factors with respect to the original ground truth delineation.  Note also that the revised decoder $\widetilde{f}$ need not necessarily be more complex than
$f_{gt}$, except in special circumstances.  For example, if we constrain our decoder architecture to be affine, then the model defaults to independent component analysis and the nonlinear operators $\bD_1$ and $\bD_2$ cannot be absorbed into an affine composite operator.  In this case, the model is identifiable up to an arbitrary permutation and scaling \citep{hyvarinen2000independent}.

Overall though, this lack of identifiability while searching for disentangled representations can exist even in trivially simple circumstances.  For example, consider a set of observed facial images generated by two ground-truth binary factors related to gender, male (M) or female (F), and age, young (Y) or old (O).  Both factors can be varied independently to modify the images with respect to either gender or age to create a set of four possible images, namely $\{\bx_{FY},\bx_{FO}, \bx_{MY}, \bx_{MO}\}$, where the subscripts indicate the latent factor values.  But if given only a set of such images without explicit labels as to the ground-truth factors of interest, finding a disentangled representation with respect to age and gender is impossible.

\begin{table}[t!]
\centering
\begin{tabular}{l|c|c}
& $\begin{array}{c} \mbox{Disentangled} \\ \mbox{Representation} \end{array}$ & $\begin{array}{c} \mbox{Entangled} \\ \mbox{Representation} \end{array}$ \\
& $z_1$  $z_2$ & $z_1$  $z_2$ \\
\hline
$\bx_{FY}$ & 0  0 & 0  0 \\
$\bx_{FO}$ & 0  1 & 1  1 \\
$\bx_{MY}$ & 1  0 & 0  1 \\
$\bx_{MO}$ & 1  1 & 1  0 \\
\hline
\end{tabular}
\caption{Illustration of identifiability issues.  Both candidate representations display zero total correlation and cannot be differentiated without supervision, but only the one on the left is disentangled with respect to the factors gender (male and female) and age (young and old).  In other words, with the disentangled representation, we can vary $z_1$ (with $z_2$ fixed) to change only gender, while we can vary $z_2$ (with $z_1$ fixed) to vary only age.  In contrast, with the entangled representation, if we vary $z_1$ (with $z_2$ fixed) both age \emph{and} gender will be changed simultaneously.}
\label{table:disentanglement_example}
\end{table}

To visualize this, let $\bz =[z_1,z_2]^{\top}$ denote a 2D binary vector that serves as the latent representation.   In Table \ref{table:disentanglement_example} we present two candidate encodings, both of which display zero total correlation between factors $z_1$ and $z_2$ (i.e., knowledge of $z_1$ tells us nothing about the value of $z_2$ and vice versa).  However, in terms of the semantically meaningful factors of gender and age, these two encodings are quite different.  In the first, varying $z_1$ alters gender independently of age, while $z_2$ varies age independently of gender.  In contrast, with the second encoding varying $z_1$ while keeping $z_2$ fixed changes both the age and gender, an \emph{entangled} representation per these attributes.  Without supervising labels to resolve such ambiguity, there is no possible way, even in principle, for any generative model (VAE or otherwise) to differentiate these two encodings such that the so-called disentangled representation is favored over the entangled one.

Although in some limited cases, it has been reported that disentangled representations can be at least partially identified, we speculate that this is a result of nuanced artifacts of the experimental design that need not translate to broader problems of interest.  Indeed, in our own empirical tests we have not been able to consistently reproduce any stable disentangled representation across various testing conditions as expected.  It was our original intent to include a thorough presentation of these results; however, in the process of preparing this manuscript we became aware of a contemporary work with extensive testing in this area \citep{locatello2018challenging}.  In full disclosure, the results from \citep{locatello2018challenging} are actually far more extensive than those that we have completed, so we simply defer the reader to this paper to examine the compelling battery of tests presented there.

\vspace*{1cm}

\appendix

\section{Comparison of Novel Samples Generated from our Model}\label{sec:experiment_gen}

\begin{figure}[t!]
    \centering
    \begin{subfigure}[t]{0.3\textwidth}
        \centering
        \includegraphics[width=1\linewidth]{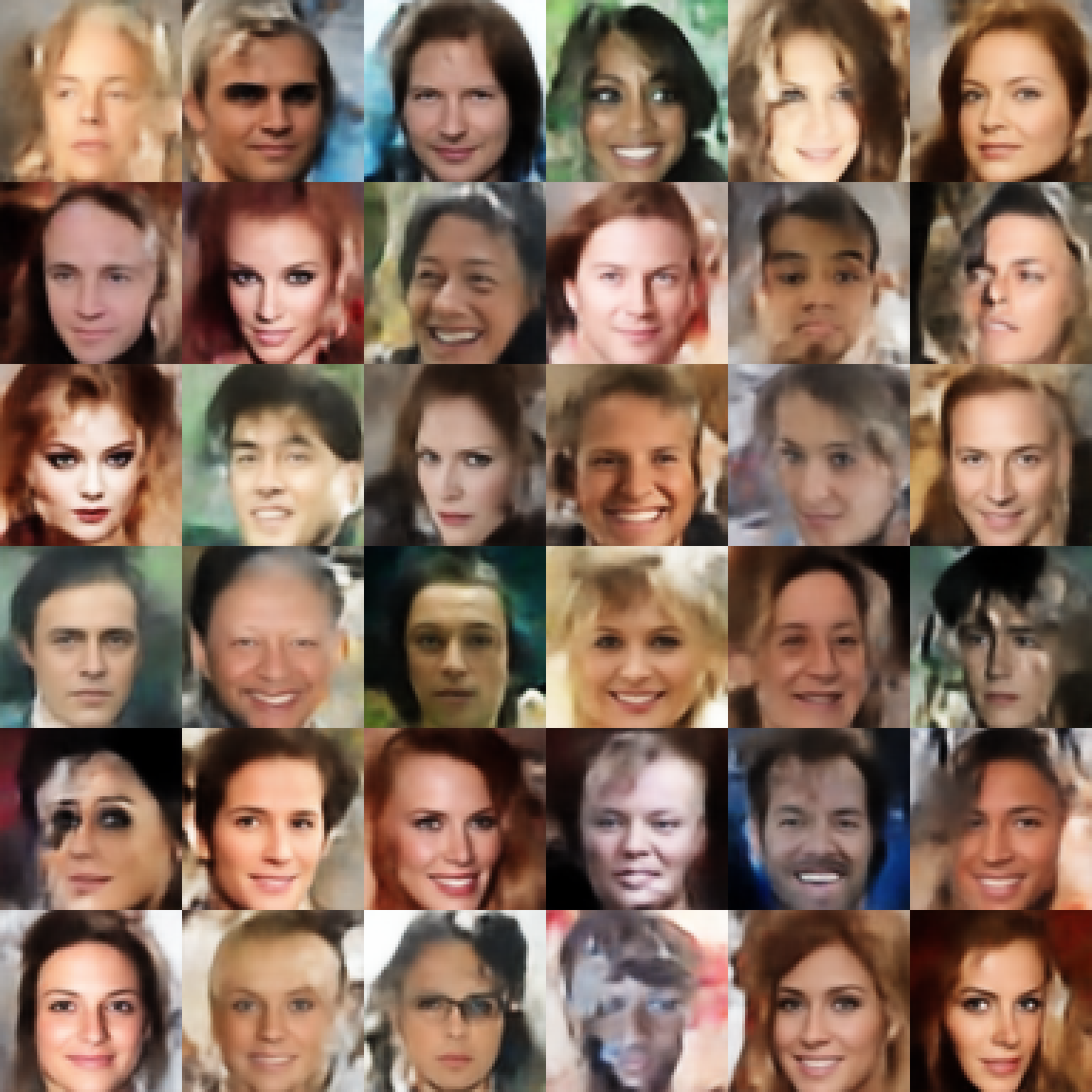}
        \caption{WAE-MMD}
    \end{subfigure}
    \begin{subfigure}[t]{0.3\textwidth}
        \centering
        \includegraphics[width=1\linewidth]{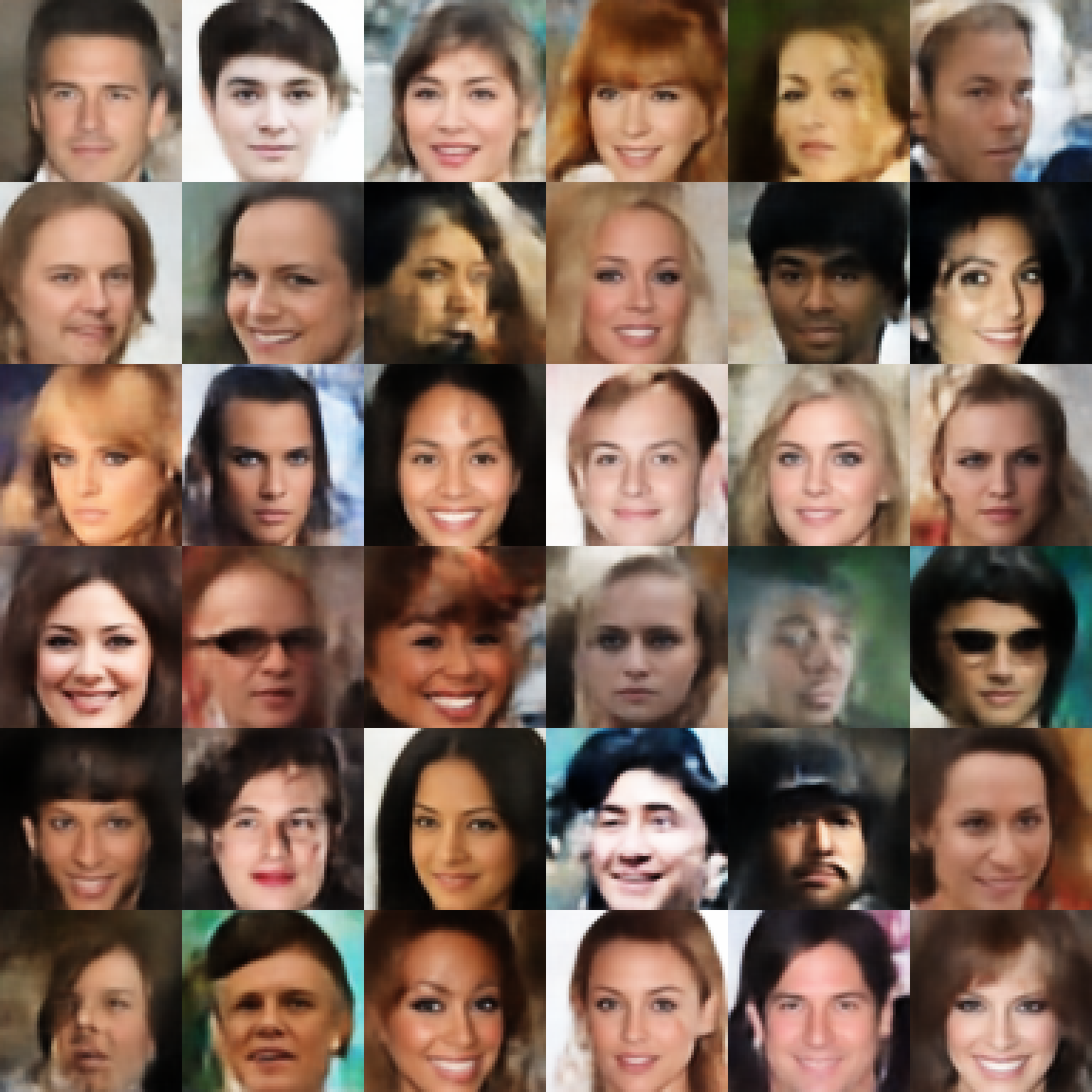}
        \caption{WAE-GAN}
    \end{subfigure}
    \begin{subfigure}[t]{0.3\textwidth}
        \centering
        \includegraphics[width=1\linewidth]{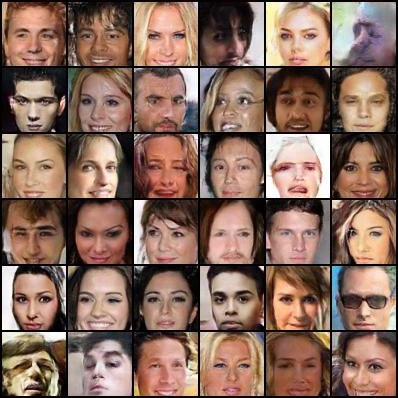}
        \caption{WGAN-GP}
    \end{subfigure}
    \begin{subfigure}[t]{0.3\textwidth}
        \centering
        \includegraphics[width=1\linewidth]{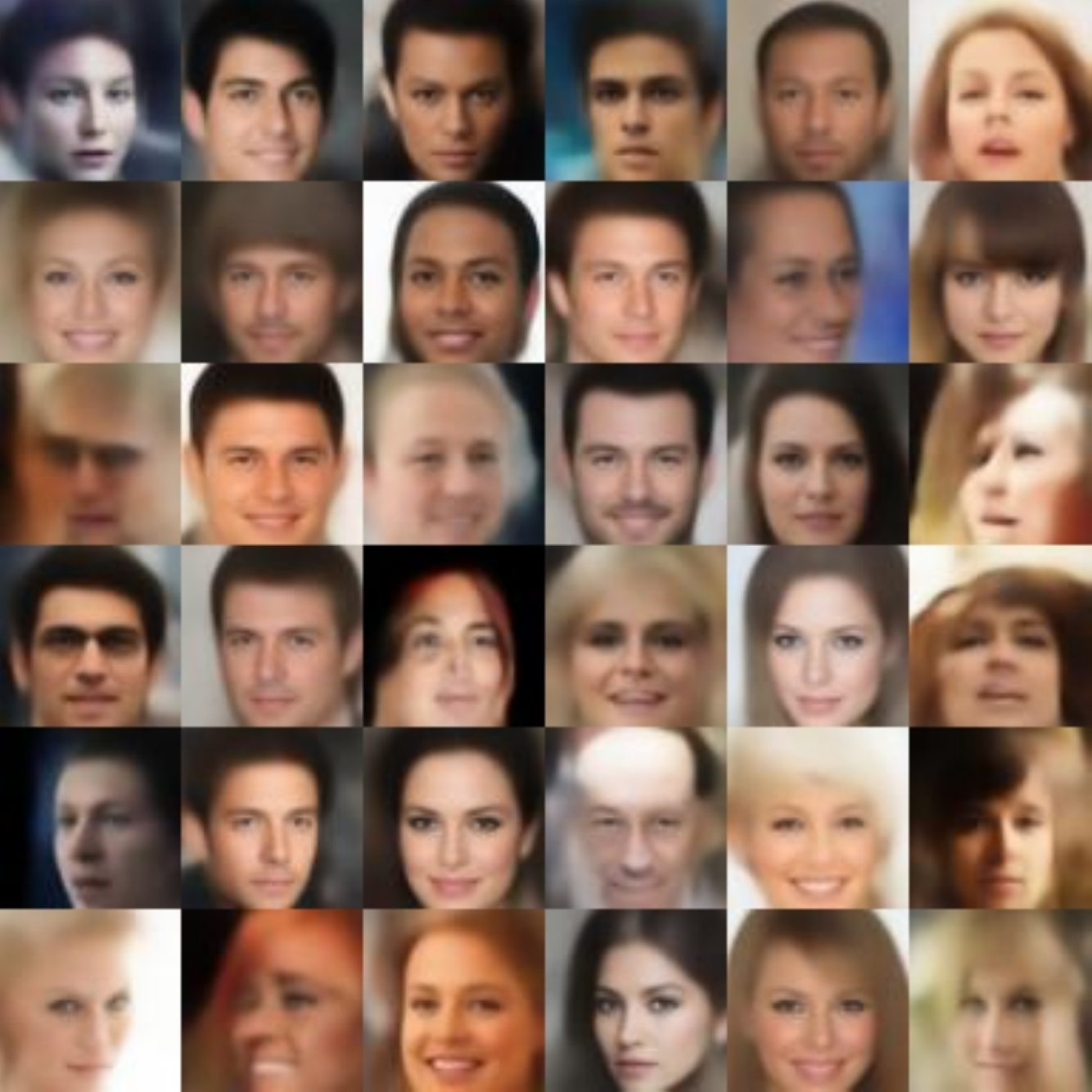}
        \caption{VAE (Fix $\gamma=1$)}
    \end{subfigure}
    \begin{subfigure}[t]{0.3\textwidth}
        \centering
        \includegraphics[width=1\linewidth]{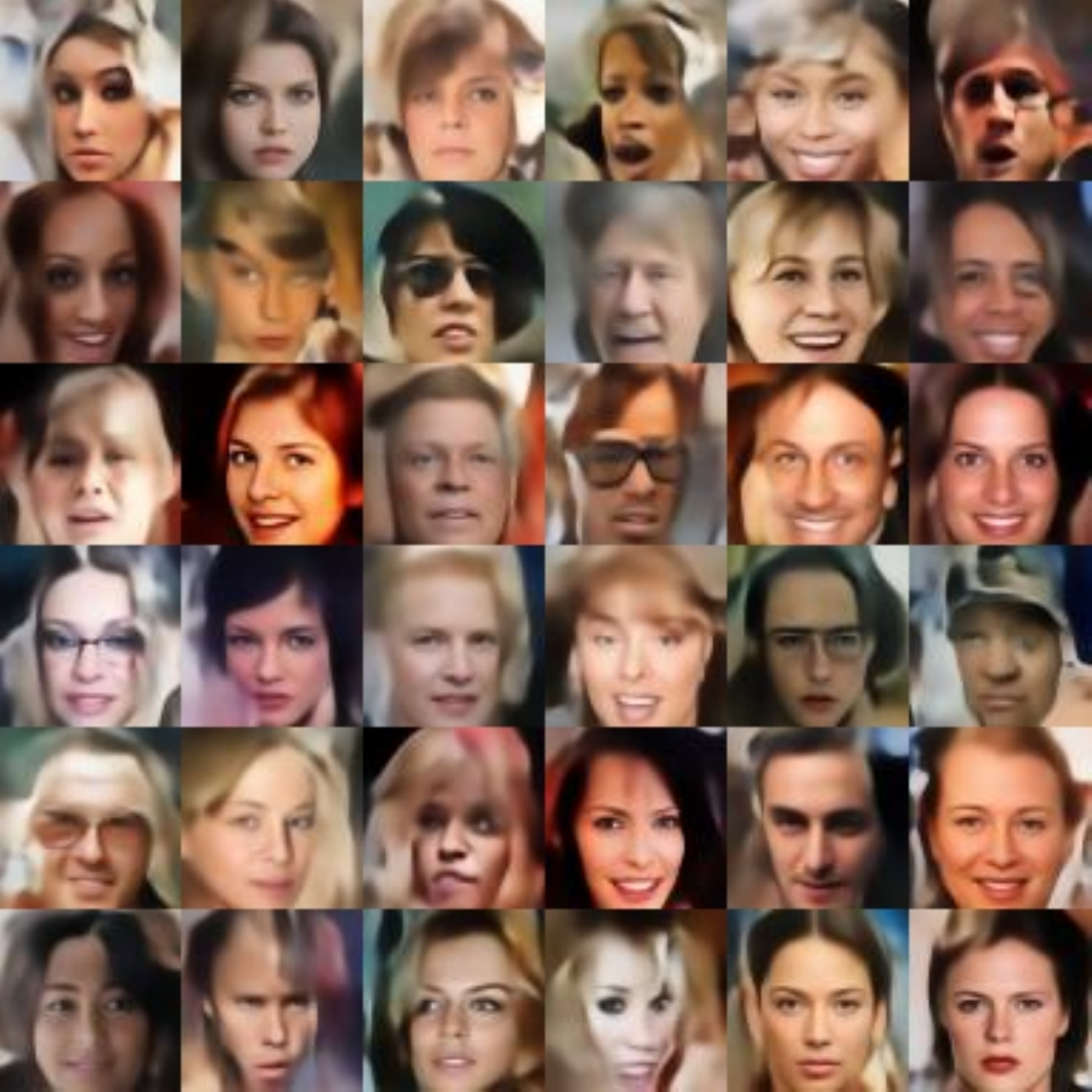}
        \caption{VAE (Learnable $\gamma$)}
    \end{subfigure}
    \begin{subfigure}[t]{0.3\textwidth}
        \centering
        \includegraphics[width=1\linewidth]{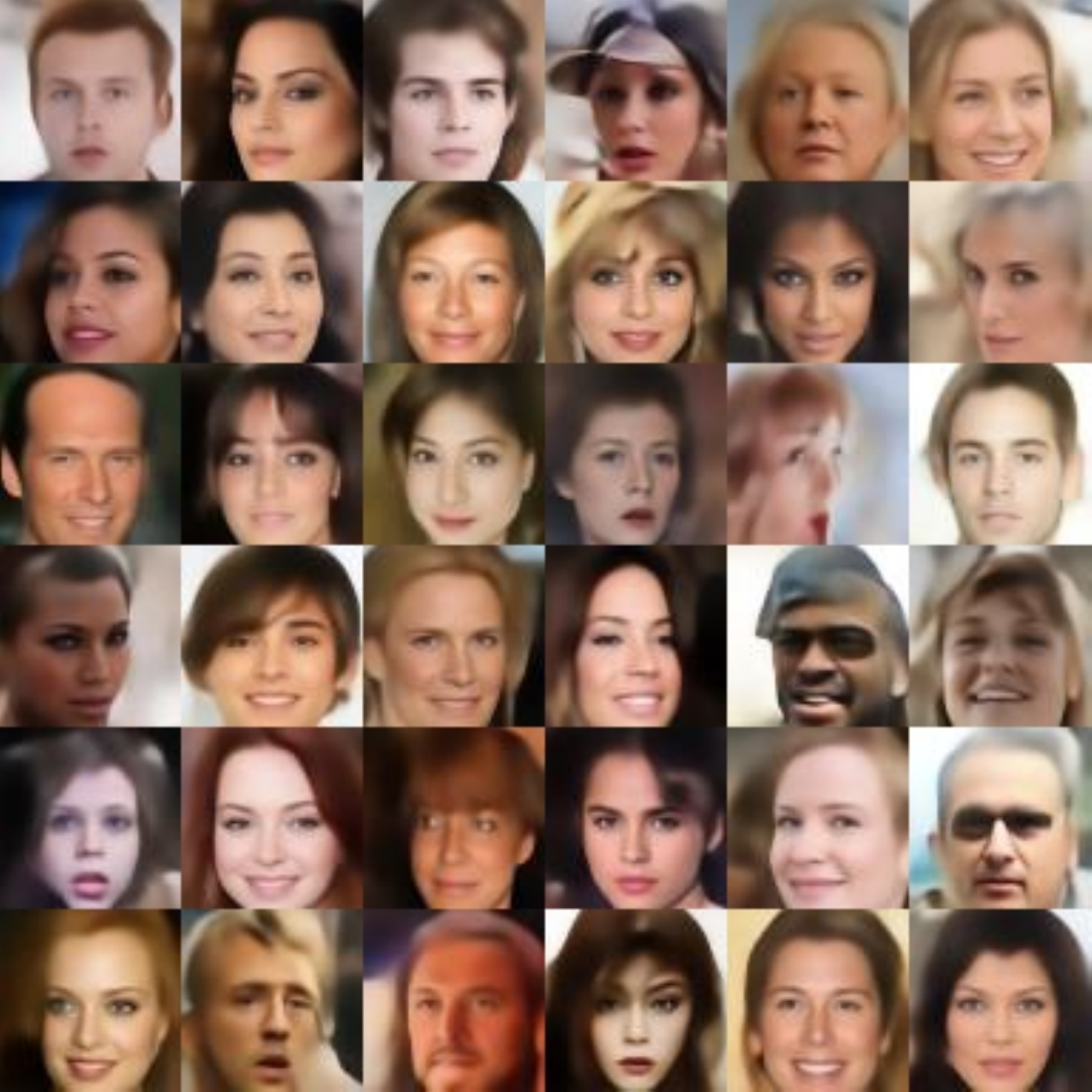}
        \caption{2-Stage VAE}
    \end{subfigure}
    \caption{Randomly generated samples on the CelebaA dataset (i.e., no cherry-picking).}
    \label{fig:gen_celeba}
\end{figure}

\begin{figure}[h]
    \centering
    \begin{subfigure}[t]{0.3\textwidth}
        \centering
        \includegraphics[width=1\linewidth]{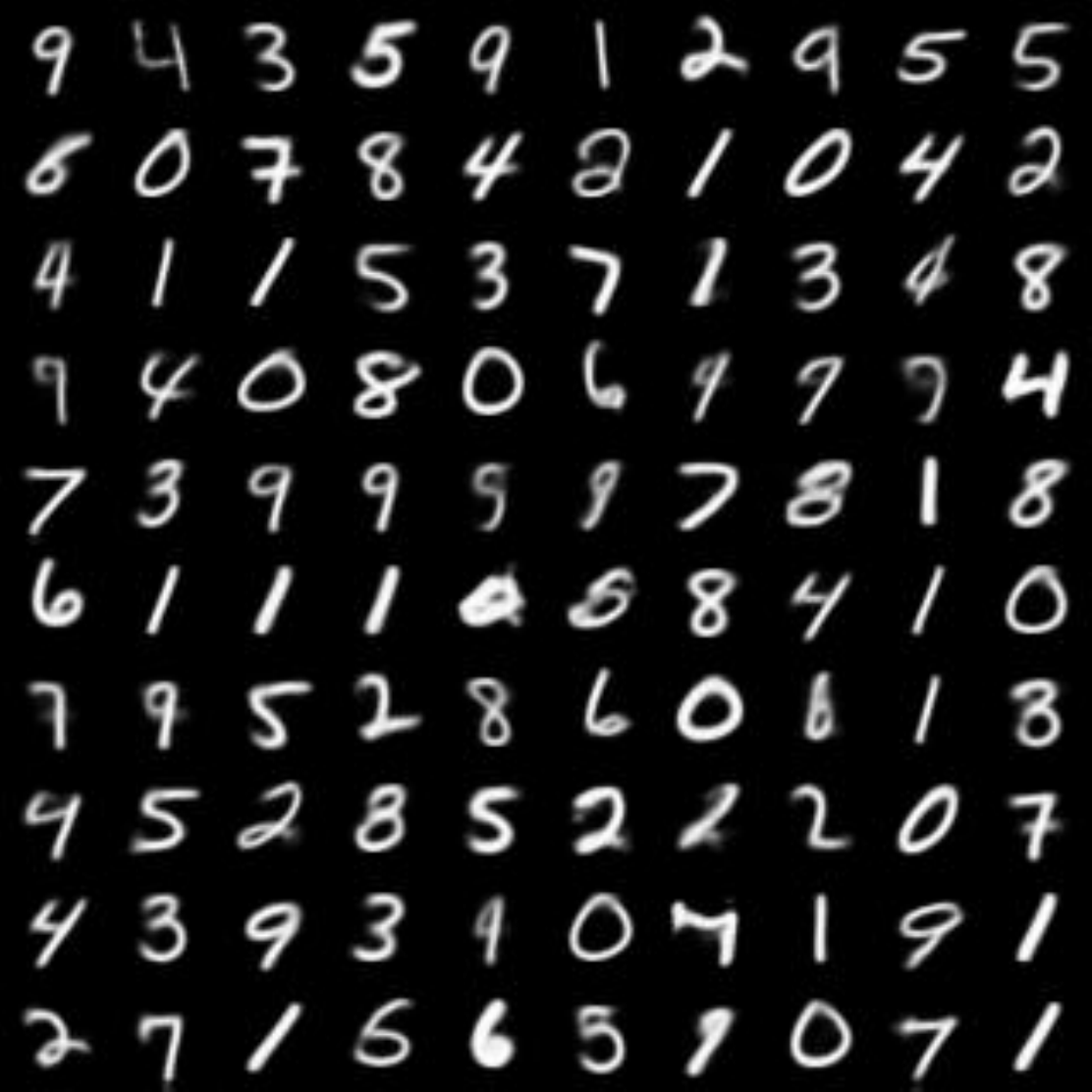}
        \caption{VAE (Fix $\gamma=1$)}
    \end{subfigure}
    \begin{subfigure}[t]{0.3\textwidth}
        \centering
        \includegraphics[width=1\linewidth]{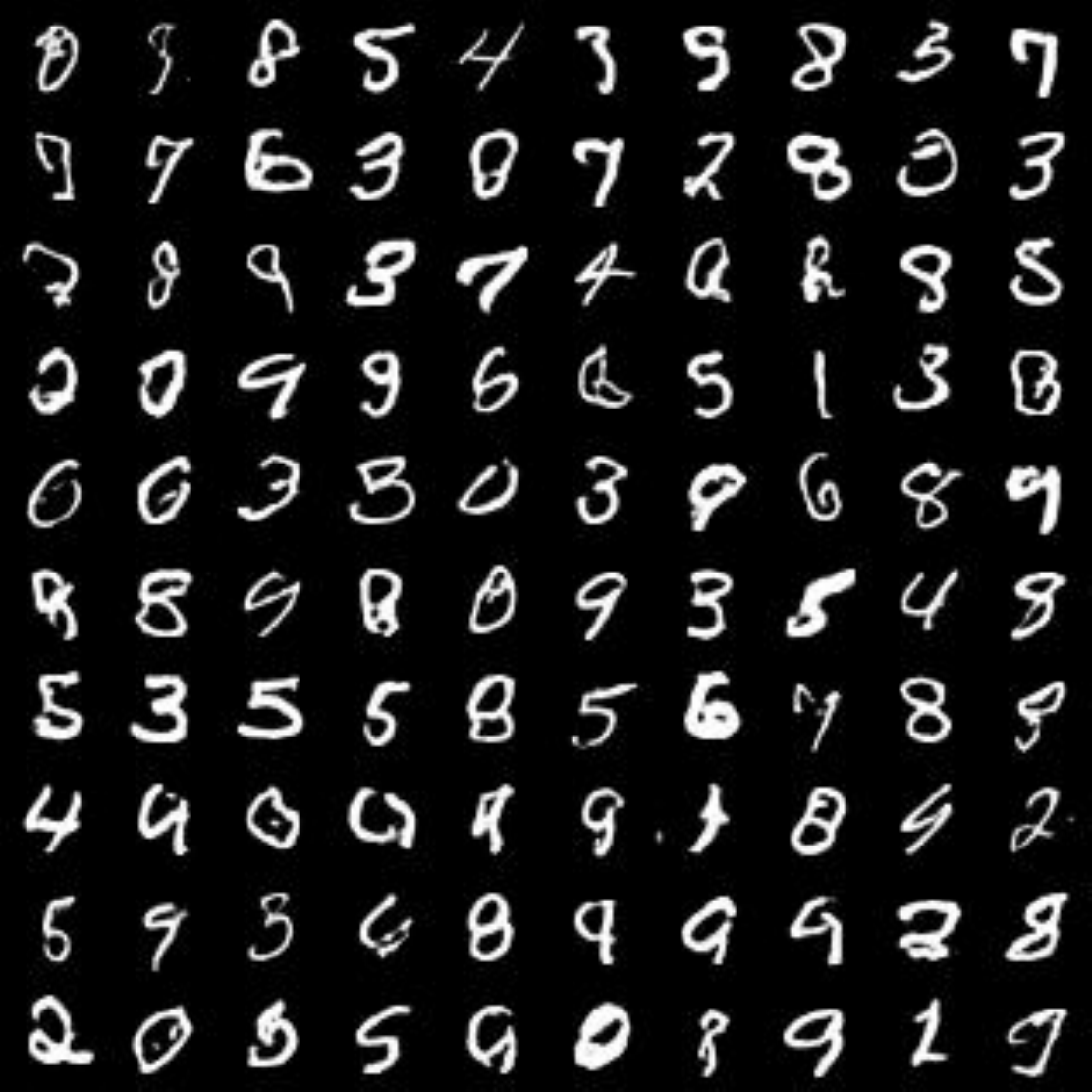}
        \caption{VAE (Learnable $\gamma$)}
    \end{subfigure}
    \begin{subfigure}[t]{0.3\textwidth}
        \centering
        \includegraphics[width=1\linewidth]{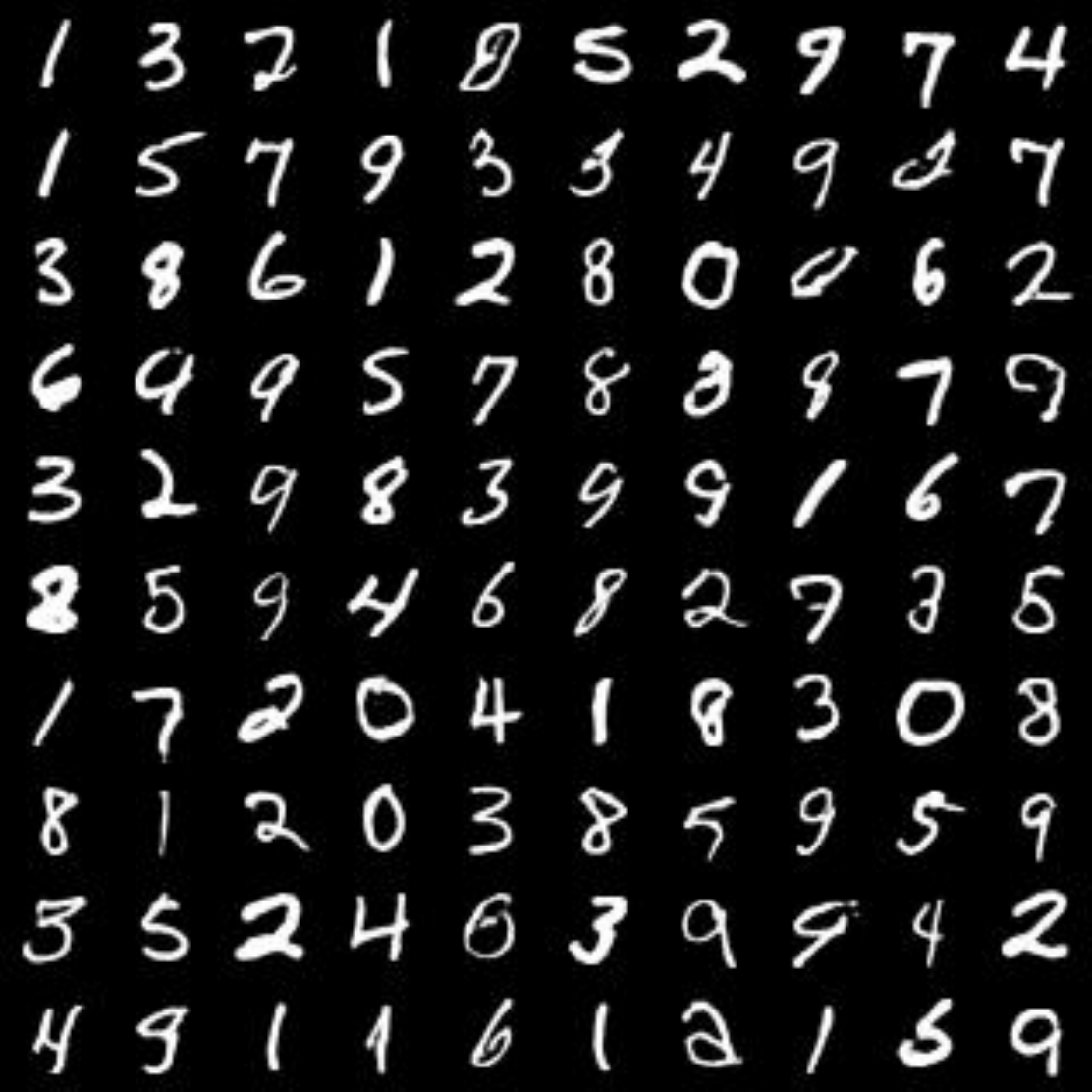}
        \caption{2-Stage VAE}
    \end{subfigure}
    \caption{Randomly generated samples on the MNIST dataset (i.e., no cherry-picking).}
    \label{fig:gen_mnist}
\end{figure}

\begin{figure}[t!]
    \centering
    \begin{subfigure}[t]{0.3\textwidth}
        \centering
        \includegraphics[width=1\linewidth]{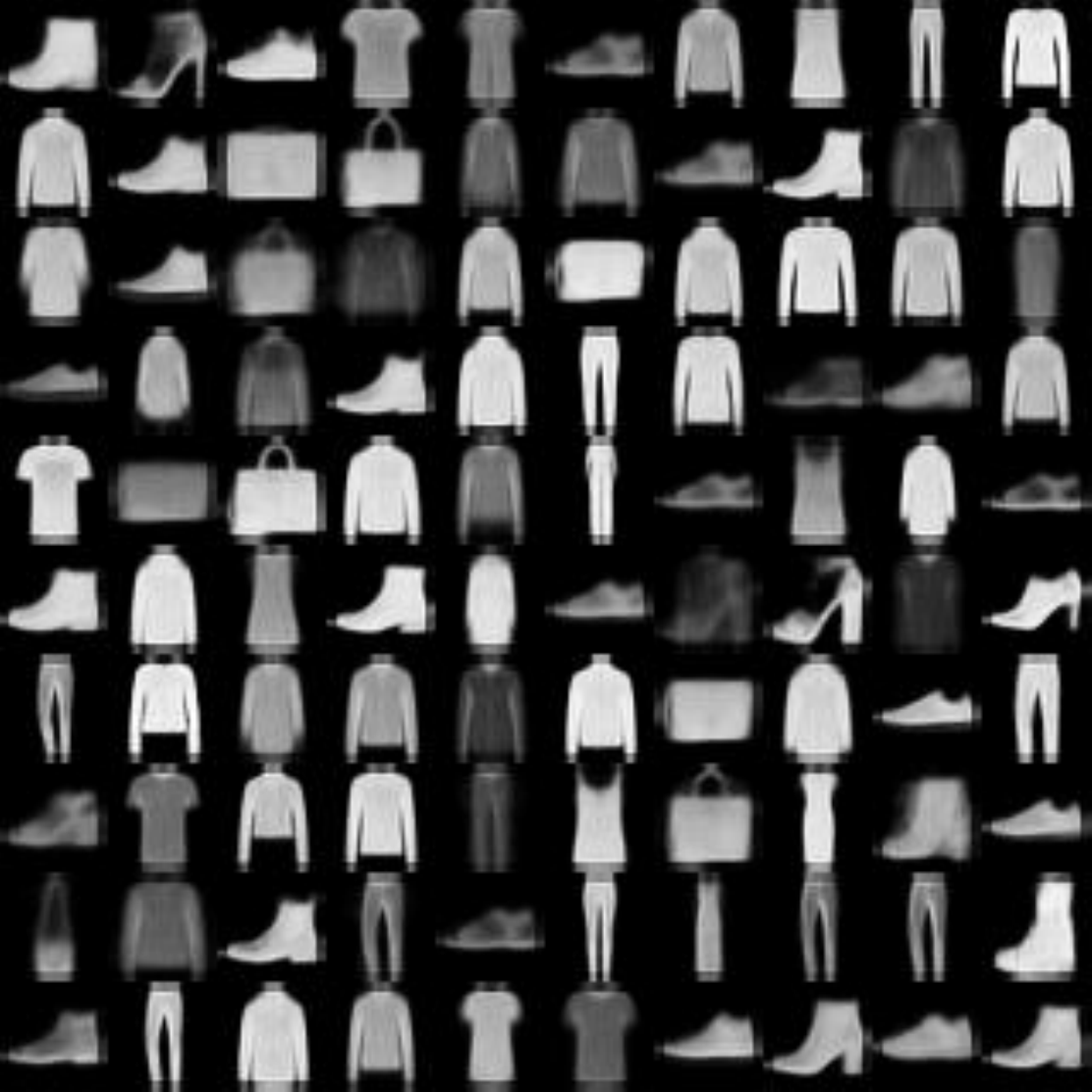}
        \caption{VAE (Fix $\gamma=1$)}
    \end{subfigure}
    \begin{subfigure}[t]{0.3\textwidth}
        \centering
        \includegraphics[width=1\linewidth]{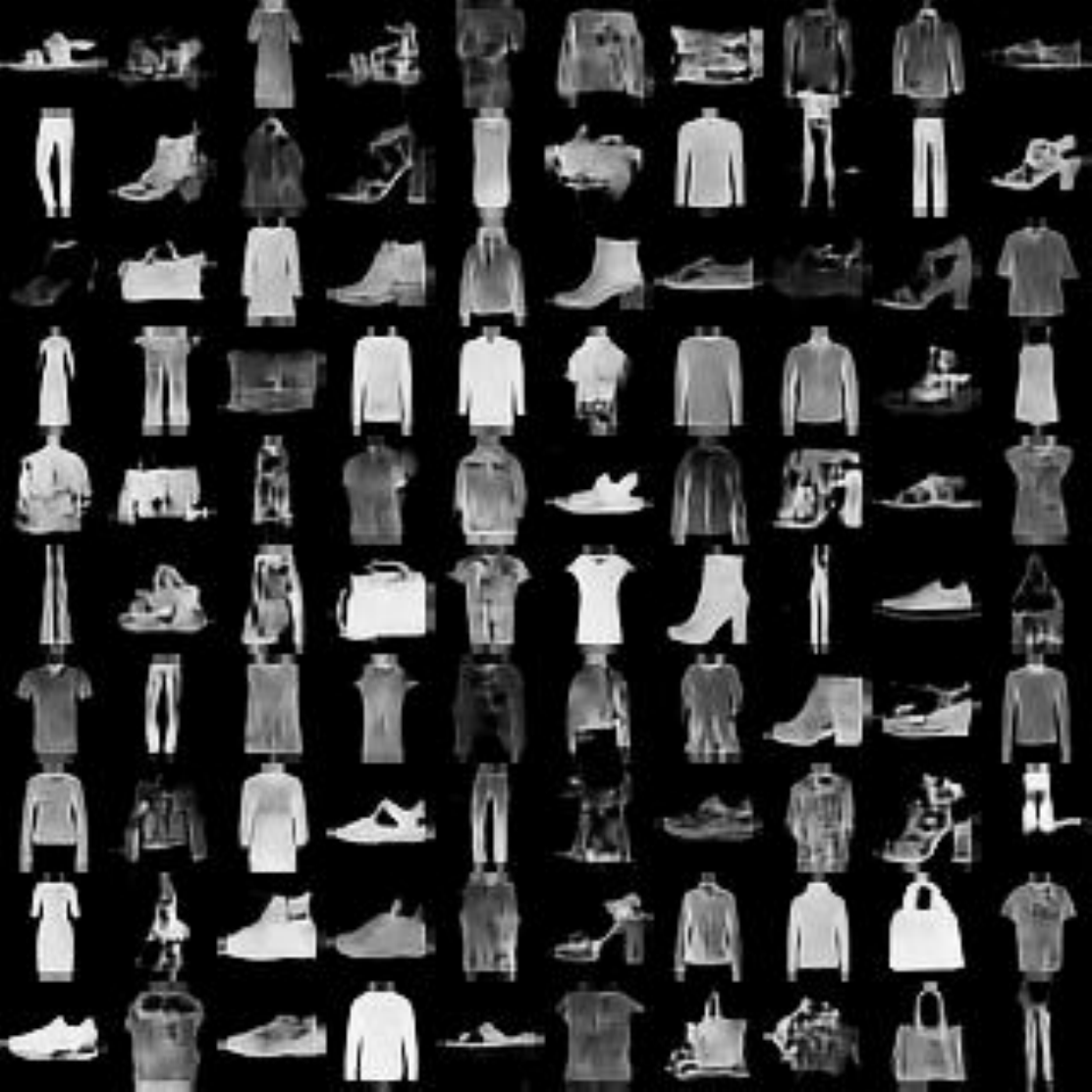}
        \caption{VAE (Learnable $\gamma$)}
    \end{subfigure}
    \begin{subfigure}[t]{0.3\textwidth}
        \centering
        \includegraphics[width=1\linewidth]{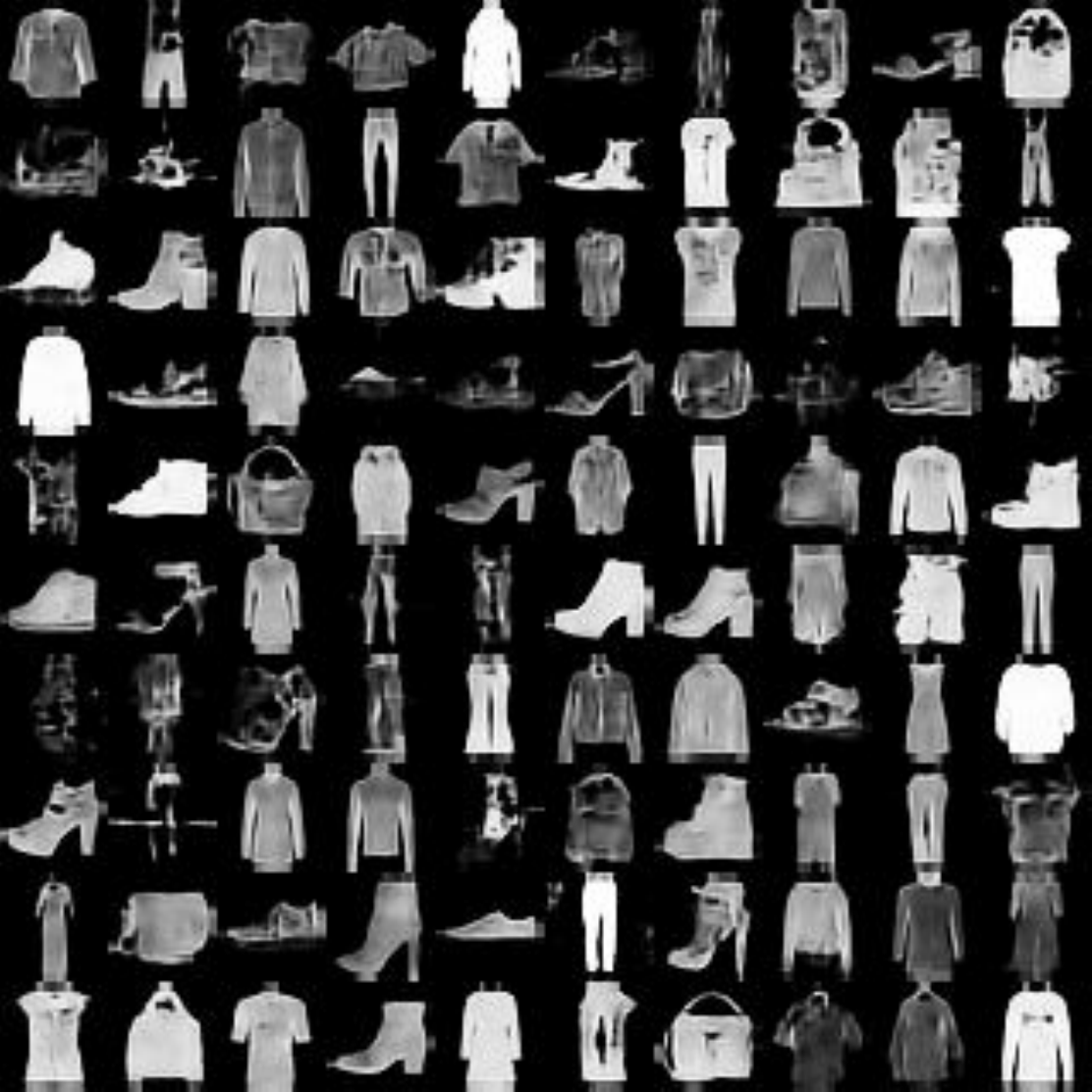}
        \caption{2-Stage VAE}
    \end{subfigure}
    \caption{Randomly Generated Samples on Fashion-MNIST Dataset (i.e., no cherry-picking).}
    \label{fig:gen_fashion}
\end{figure}

\begin{figure}[t!]
    \centering
    \begin{subfigure}[t]{0.3\textwidth}
        \centering
        \includegraphics[width=1\linewidth]{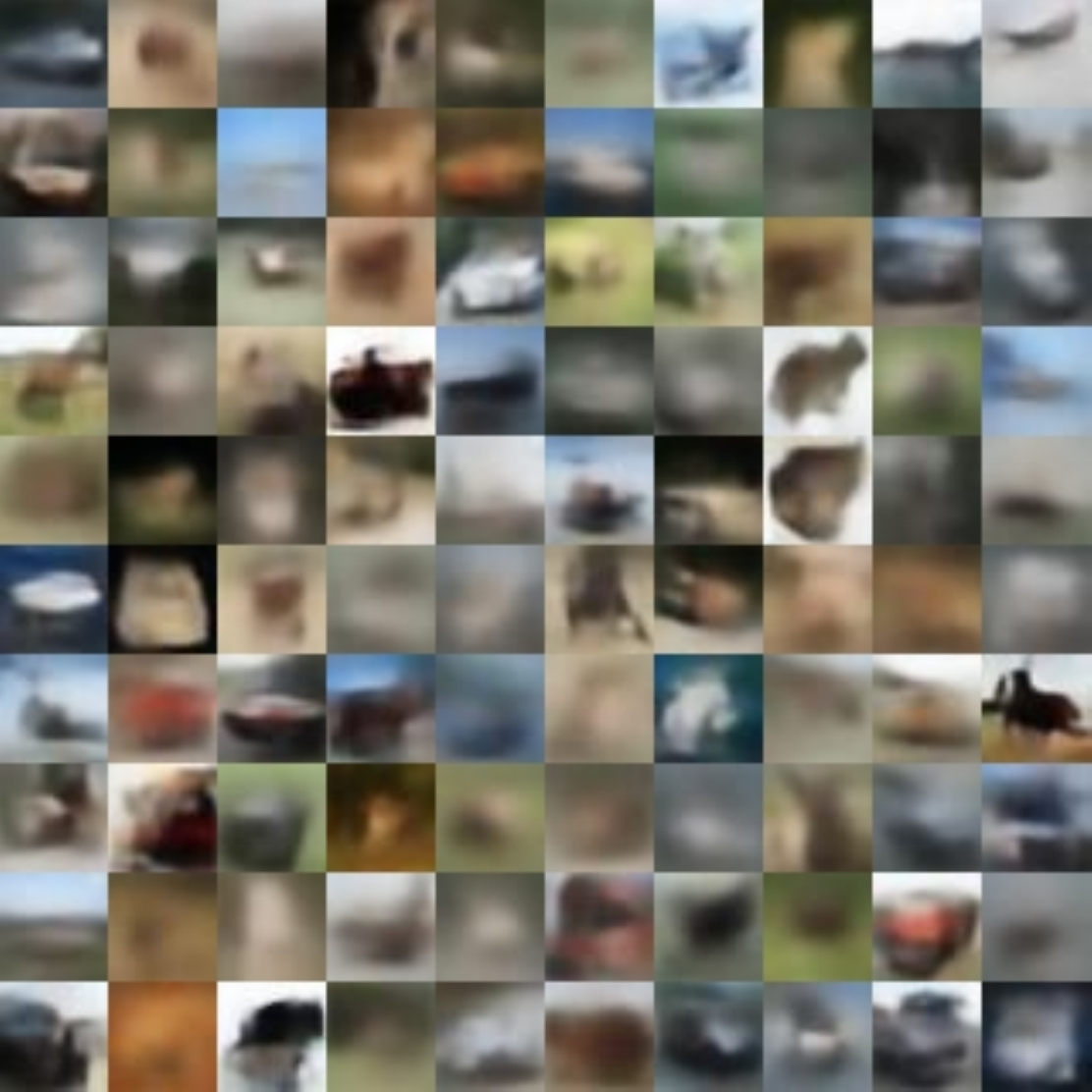}
        \caption{VAE (Fix $\gamma=1$)}
    \end{subfigure}
    \begin{subfigure}[t]{0.3\textwidth}
        \centering
        \includegraphics[width=1\linewidth]{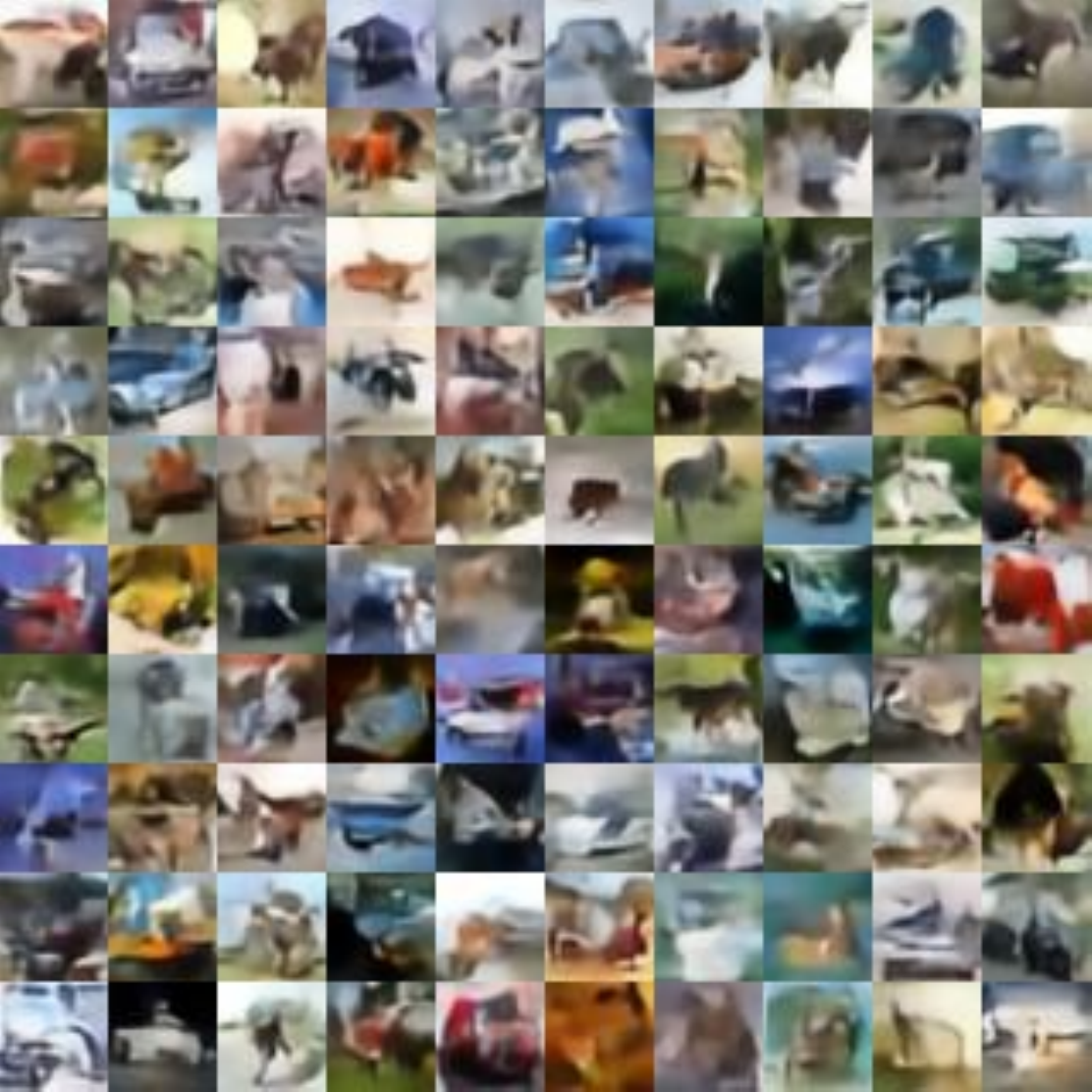}
        \caption{VAE (Learnable $\gamma$)}
    \end{subfigure}
    \begin{subfigure}[t]{0.3\textwidth}
        \centering
        \includegraphics[width=1\linewidth]{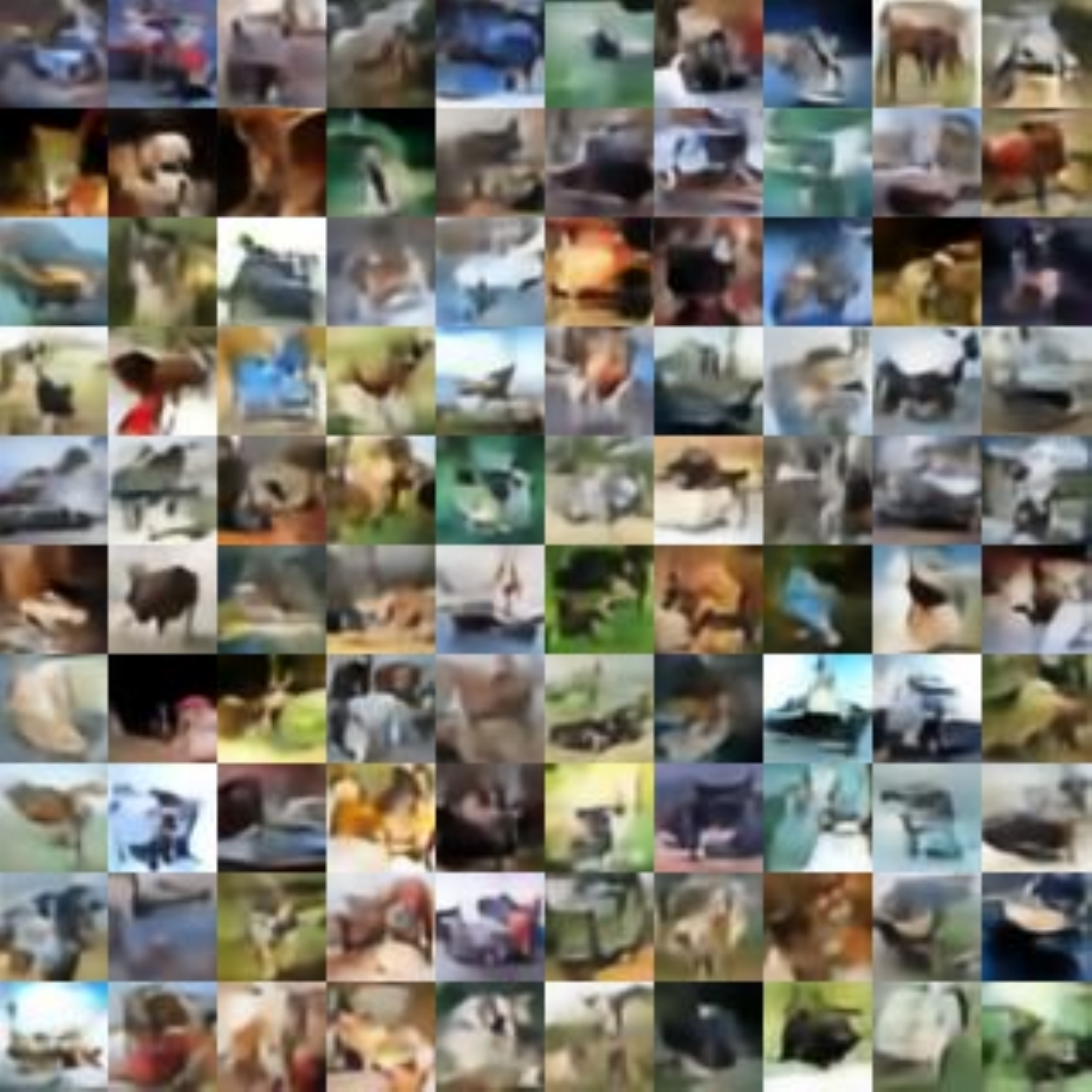}
        \caption{2-Stage VAE}
    \end{subfigure}
    \caption{Randomly Generated Samples on CIFAR-10 Dataset (i.e., no cherry-picking).}
    \label{fig:gen_cifar10}
\end{figure}

Generation results for CelebA, MNIST, Fashion-MNIST and CIFAR-10 datasets of different methods are shown in Figures~\ref{fig:gen_celeba}$-$\ref{fig:gen_cifar10} respectively. When $\gamma$ is fixed to be one, the generated samples are very blurry.  If a learnable $\gamma$ is used, the samples becomes sharper; however, there are many lingering artifacts as expected. In contrast, the proposed 2-Stage VAE can remove these artifacts and generate more realistic samples. For comparison purposes, we also show the results from WAE-MMD, WAE-GAN~\citep{tolstikhin2018wasserstein} and WGAN-GP~\citep{gulrajani2017improved} for the CelebA dataset.

\begin{figure}[h]
    \centering
    \begin{subfigure}[t]{0.3\textwidth}
        \centering
        \includegraphics[width=1\linewidth]{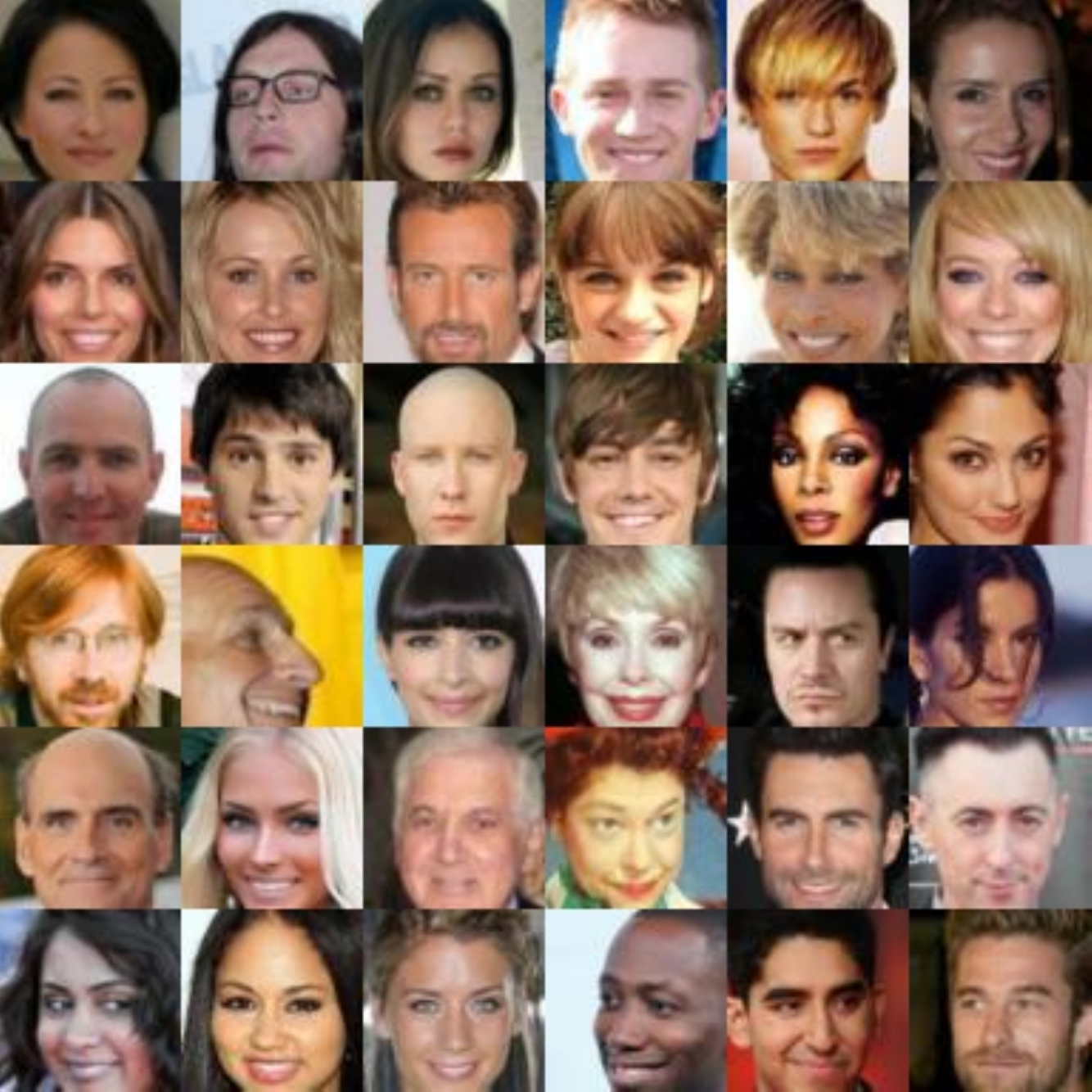}
        \caption{Ground Truth}
    \end{subfigure}
    \begin{subfigure}[t]{0.3\textwidth}
        \centering
        \includegraphics[width=1\linewidth]{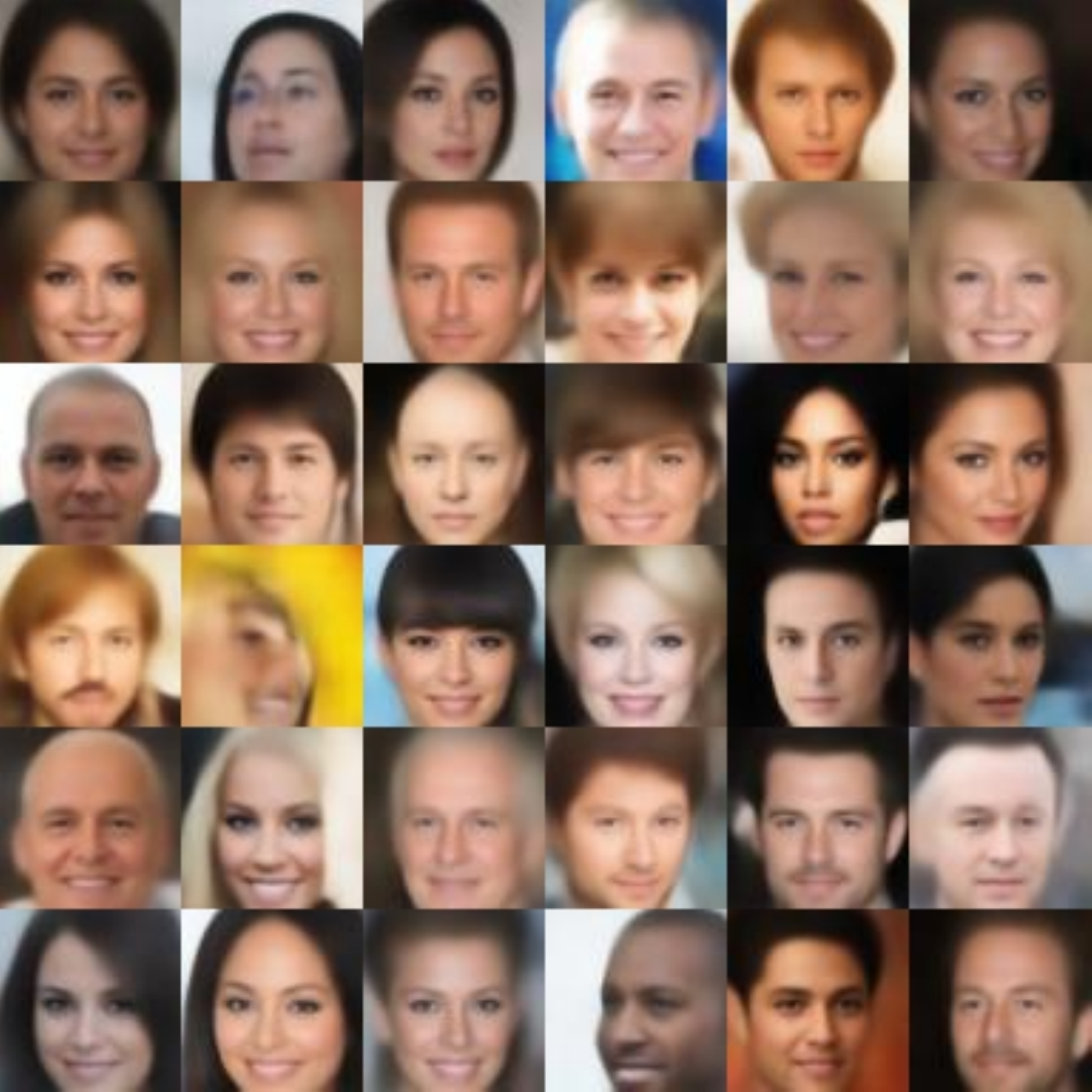}
        \caption{VAE (Fix $\gamma=1$)}
    \end{subfigure}
    \begin{subfigure}[t]{0.3\textwidth}
        \centering
        \includegraphics[width=1\linewidth]{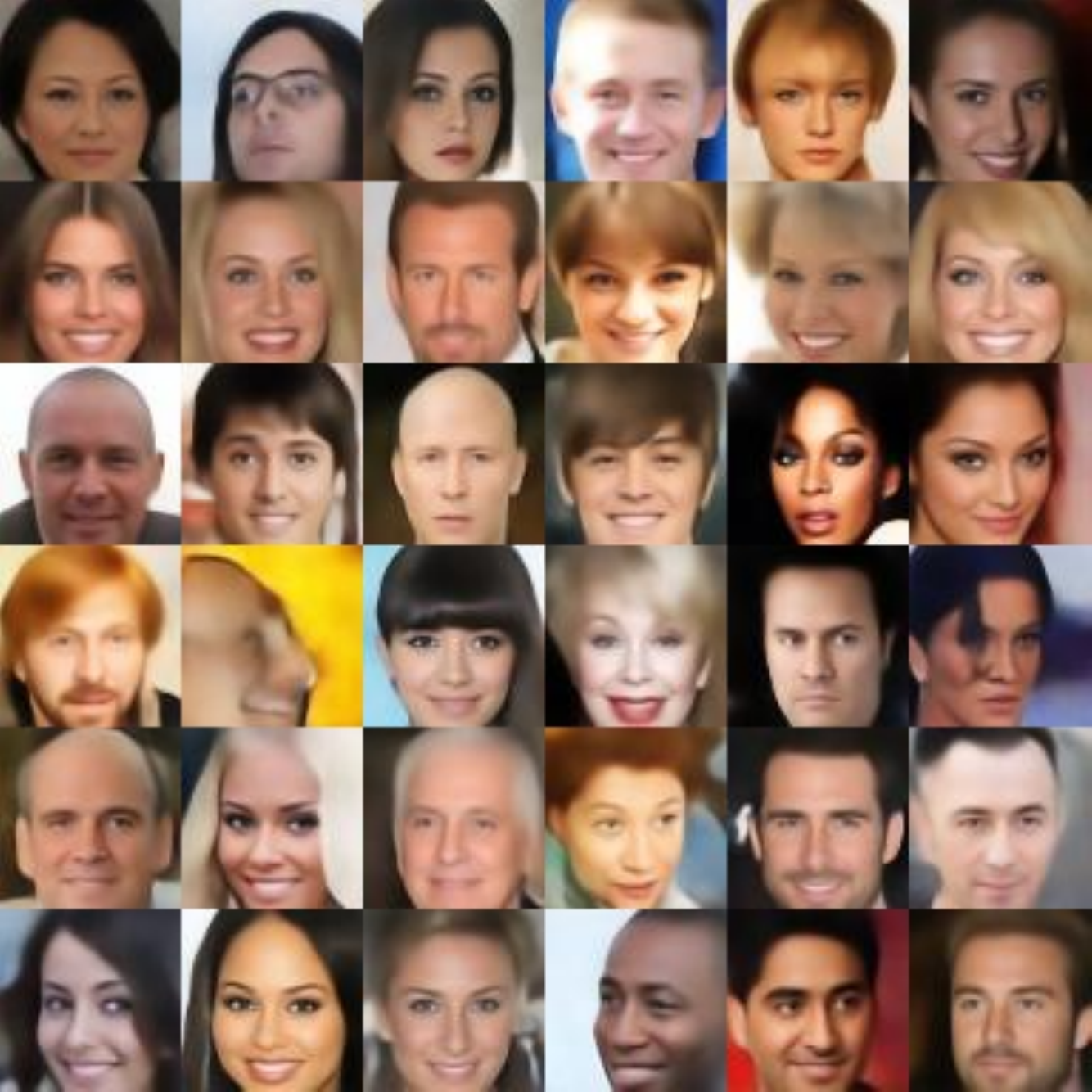}
        \caption{VAE (Learnable $\gamma$)}
    \end{subfigure}
    \caption{Reconstructions on CelebA Dataset.}
    \label{fig:recon_celeba}
\end{figure}


\begin{figure}[h]
    \centering
    \begin{subfigure}[t]{0.3\textwidth}
        \centering
        \includegraphics[width=1\linewidth]{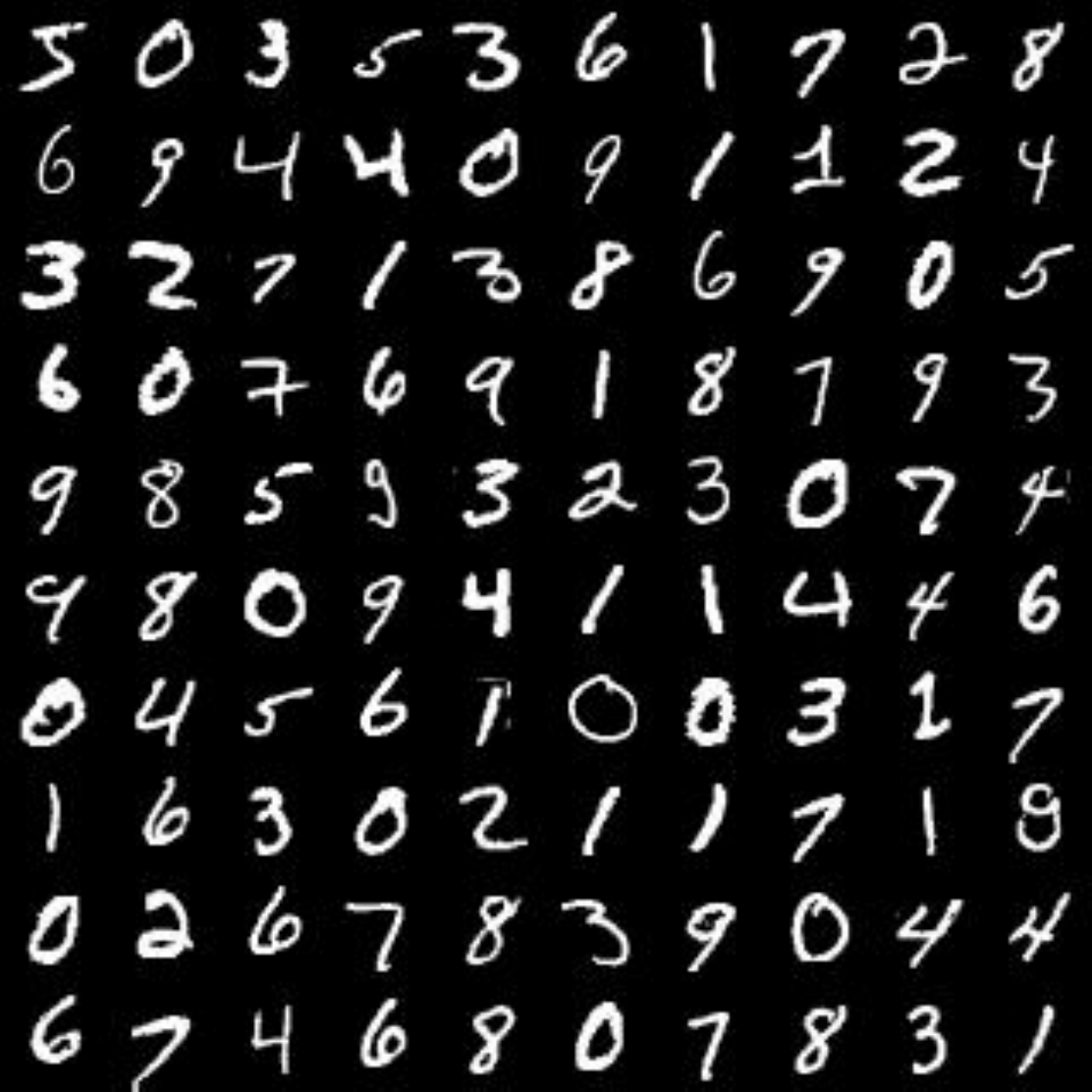}
        \caption{Ground Truth}
    \end{subfigure}
    \begin{subfigure}[t]{0.3\textwidth}
        \centering
        \includegraphics[width=1\linewidth]{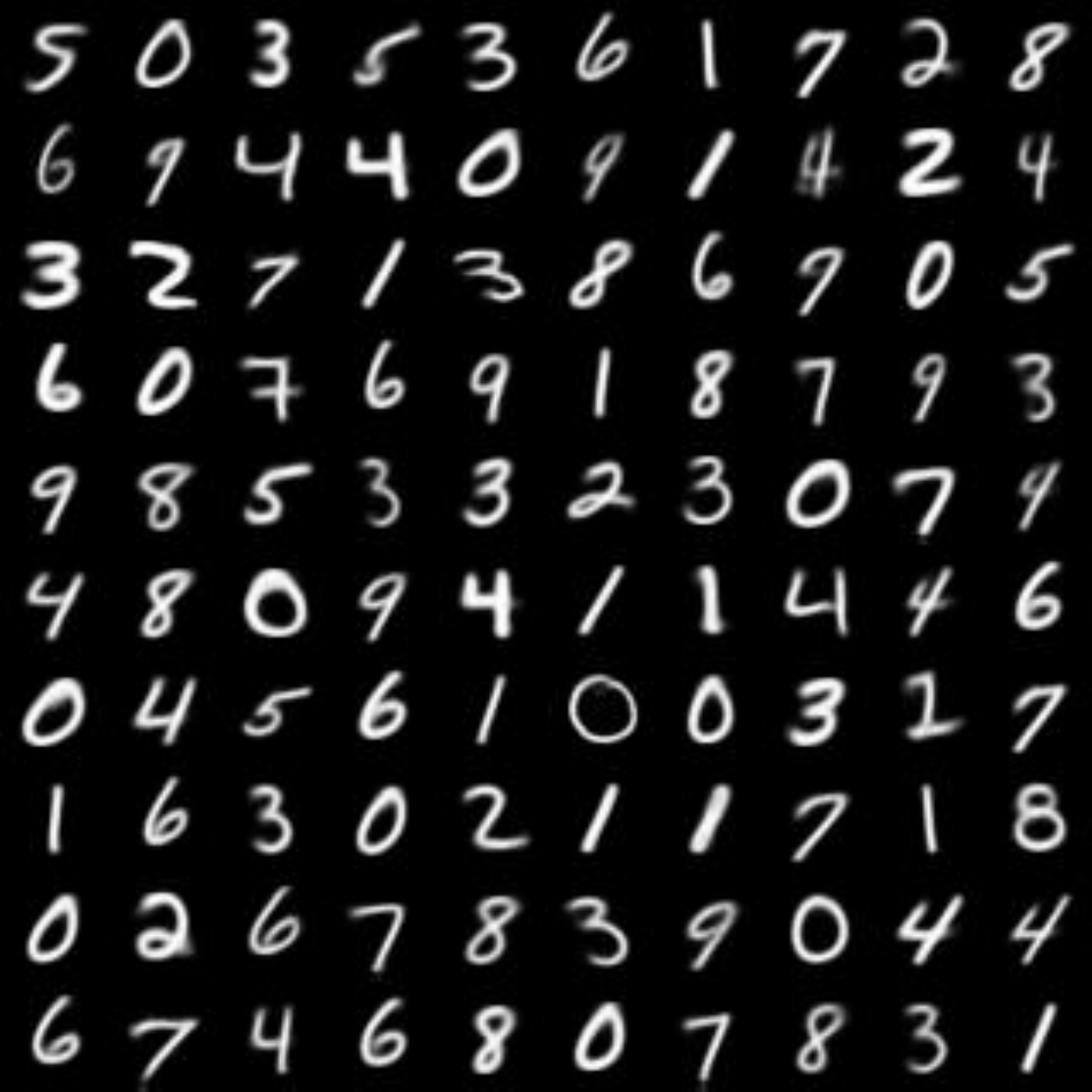}
        \caption{VAE (Fix $\gamma=1$)}
    \end{subfigure}
    \begin{subfigure}[t]{0.3\textwidth}
        \centering
        \includegraphics[width=1\linewidth]{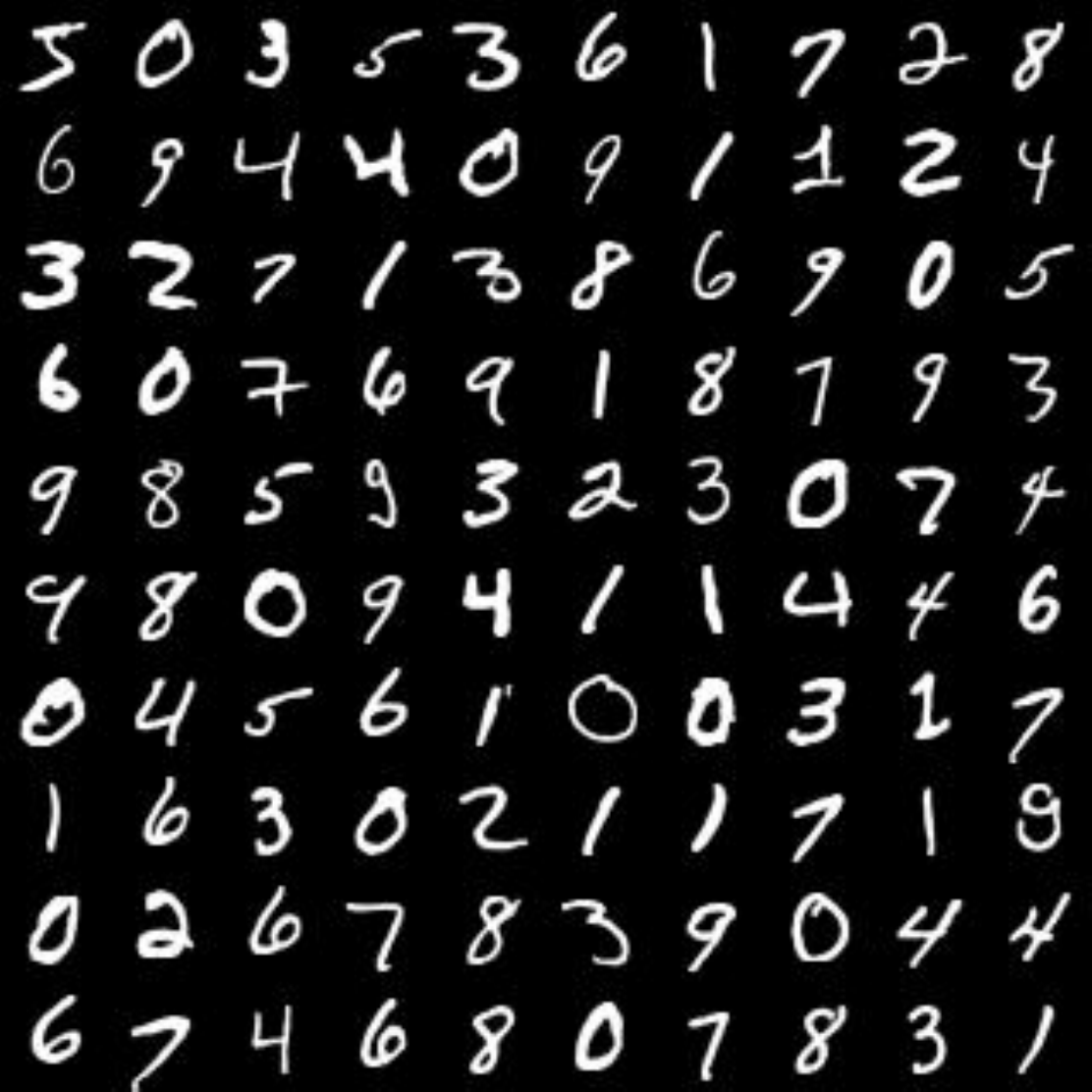}
        \caption{VAE (Learnable $\gamma$)}
    \end{subfigure}
    \caption{Reconstructions on MNIST Dataset.}
    \label{fig:recon_mnist}
\end{figure}

\section{Example Reconstructions of Training Data} \label{sec:experiment_recon}

\begin{figure}[t!]
    \centering
    \begin{subfigure}[t]{0.3\textwidth}
        \centering
        \includegraphics[width=1\linewidth]{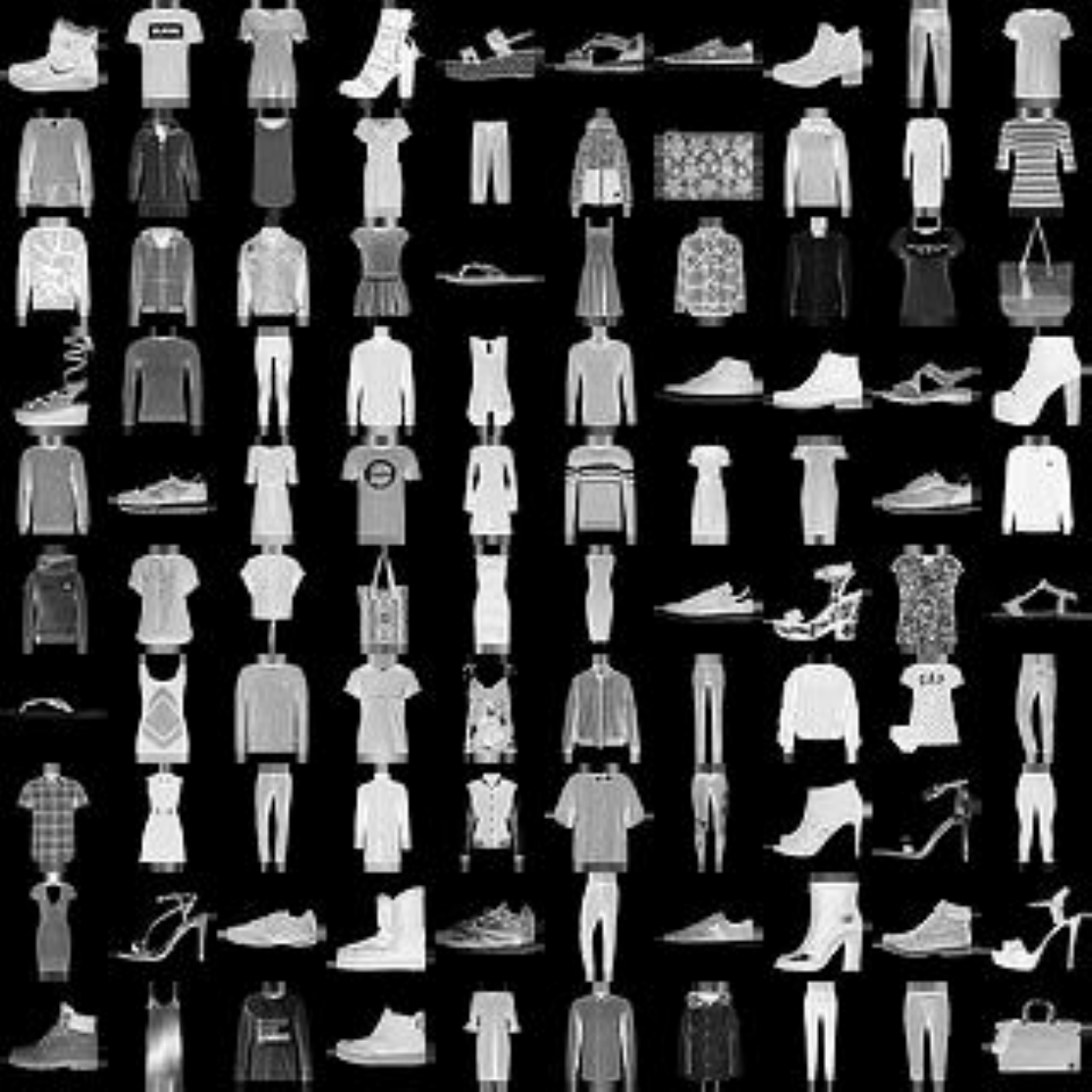}
        \caption{Ground Truth}
    \end{subfigure}
    \begin{subfigure}[t]{0.3\textwidth}
        \centering
        \includegraphics[width=1\linewidth]{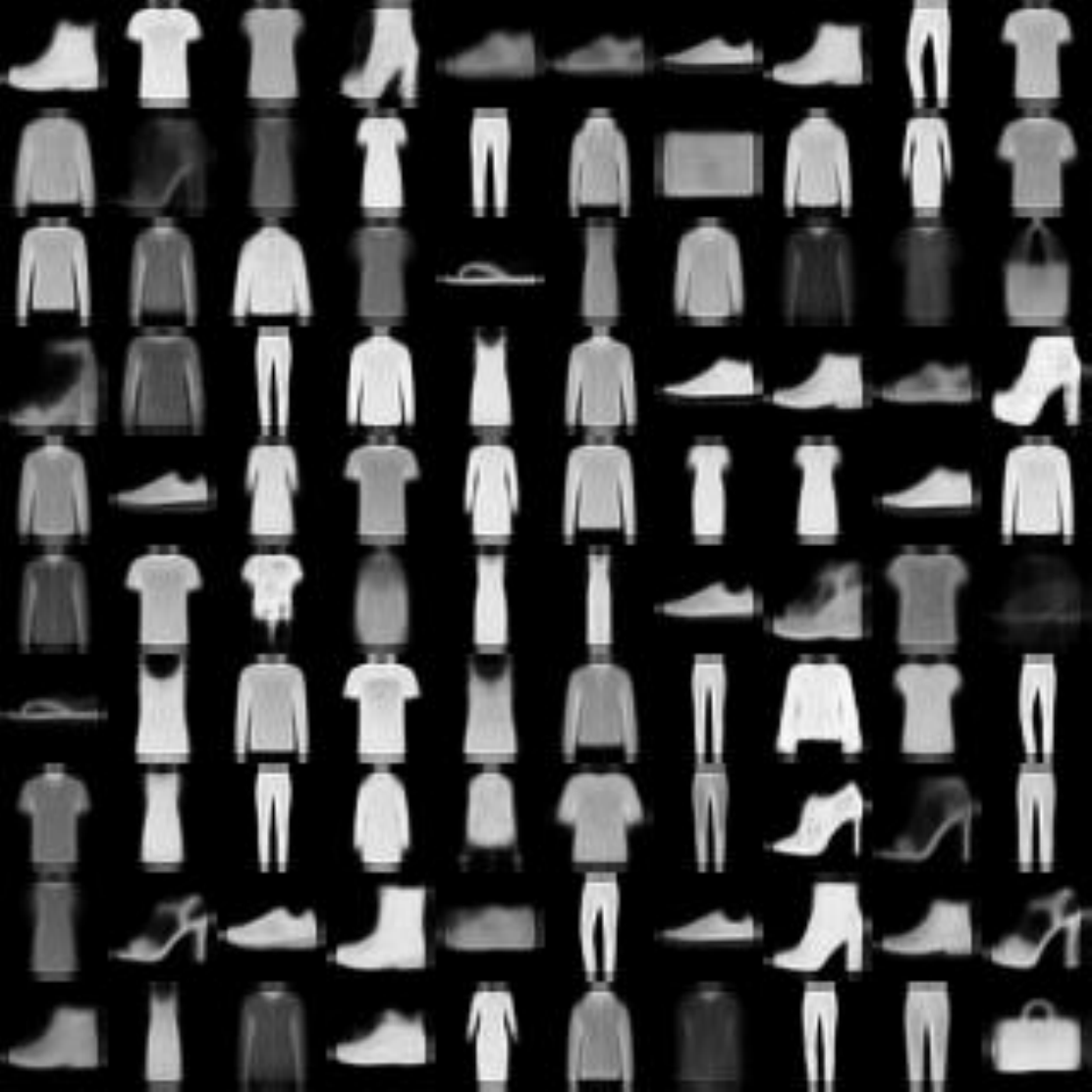}
        \caption{VAE (Fix $\gamma=1$)}
    \end{subfigure}
    \begin{subfigure}[t]{0.3\textwidth}
        \centering
        \includegraphics[width=1\linewidth]{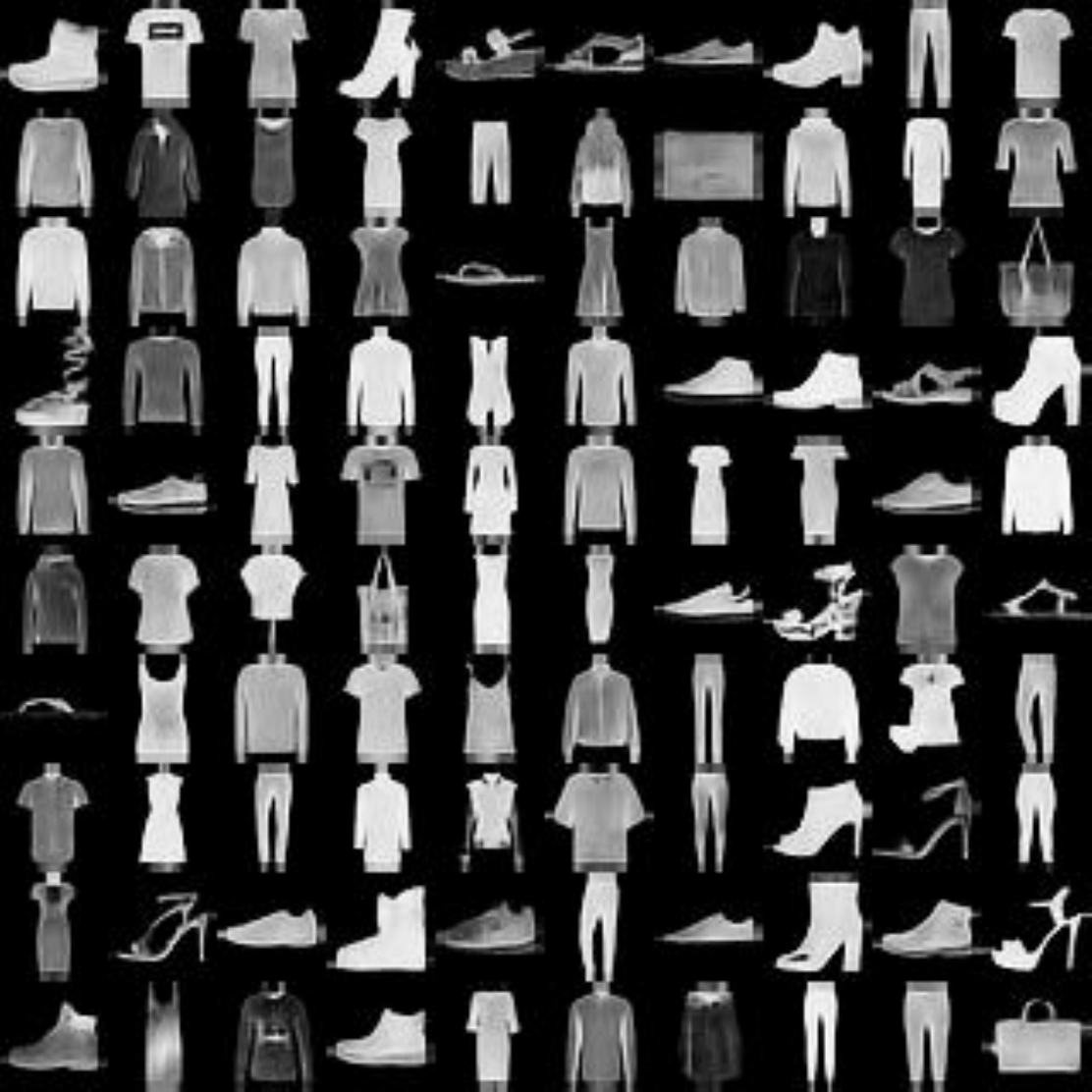}
        \caption{VAE (Learnable $\gamma$)}
    \end{subfigure}
    \caption{Reconstructions on Fashion-MNIST Dataset.}
    \label{fig:recon_fashion}
\end{figure}

\begin{figure}[t!]
    \centering
    \begin{subfigure}[t]{0.3\textwidth}
        \centering
        \includegraphics[width=1\linewidth]{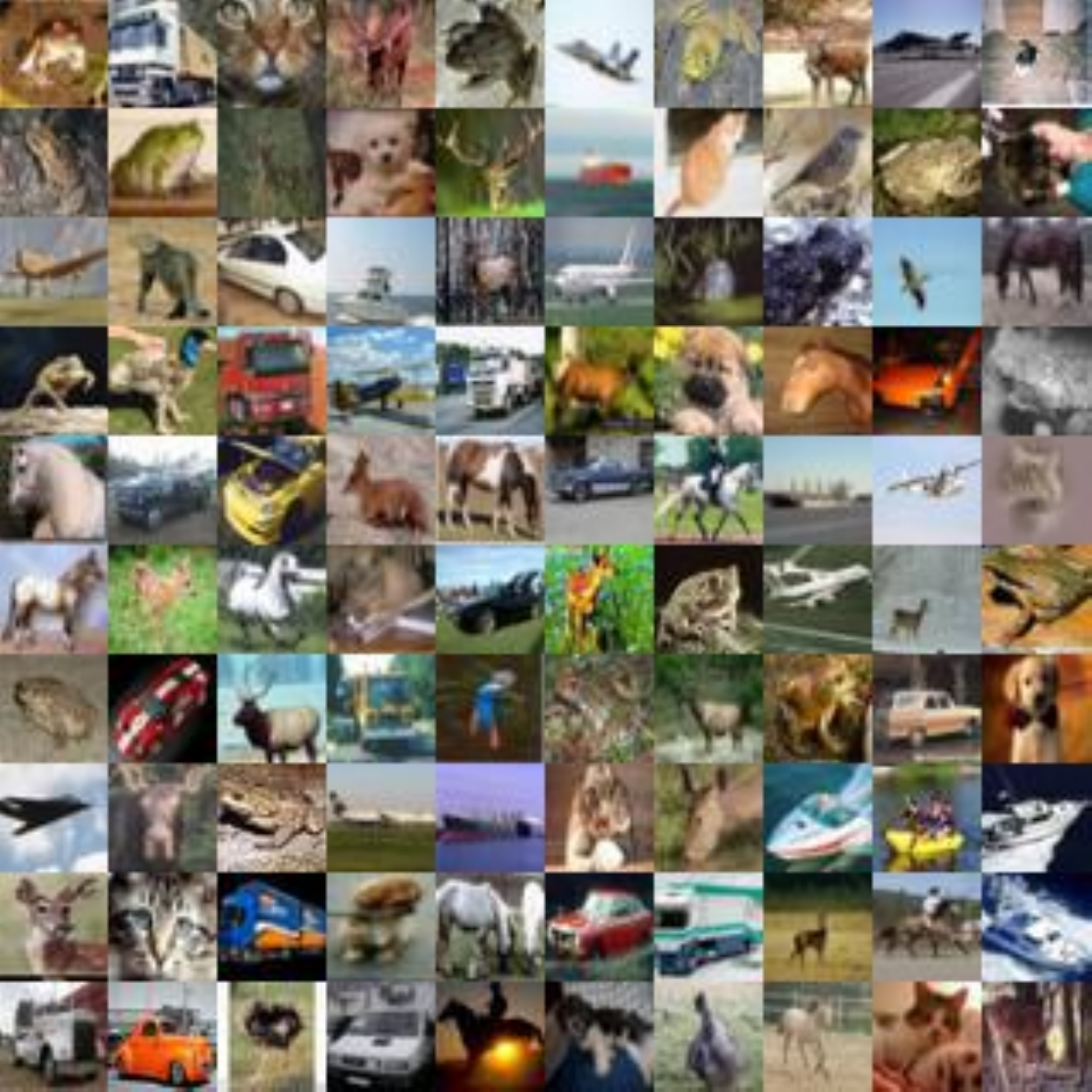}
        \caption{Ground Truth}
    \end{subfigure}
    \begin{subfigure}[t]{0.3\textwidth}
        \centering
        \includegraphics[width=1\linewidth]{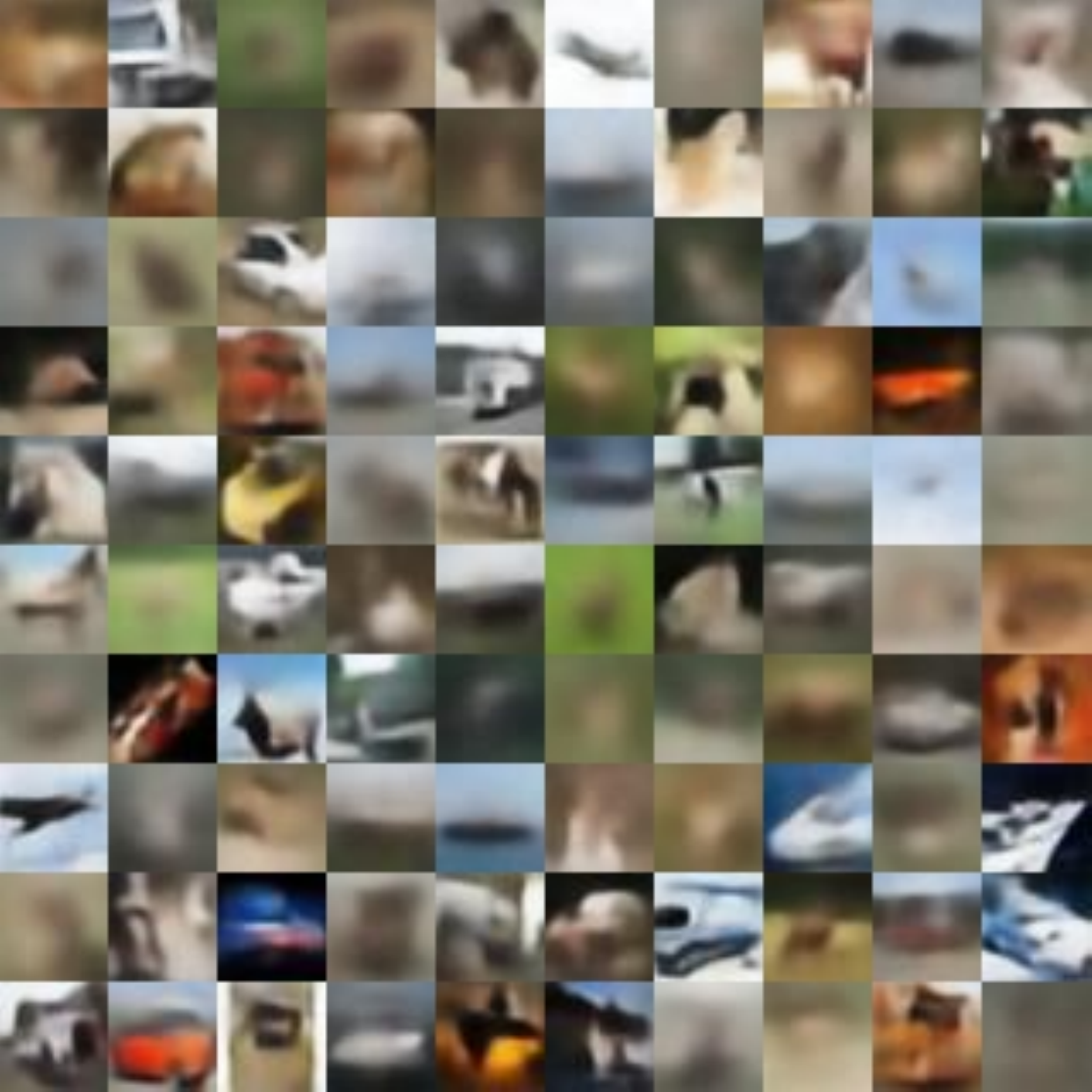}
        \caption{VAE (Fix $\gamma=1$)}
    \end{subfigure}
    \begin{subfigure}[t]{0.3\textwidth}
        \centering
        \includegraphics[width=1\linewidth]{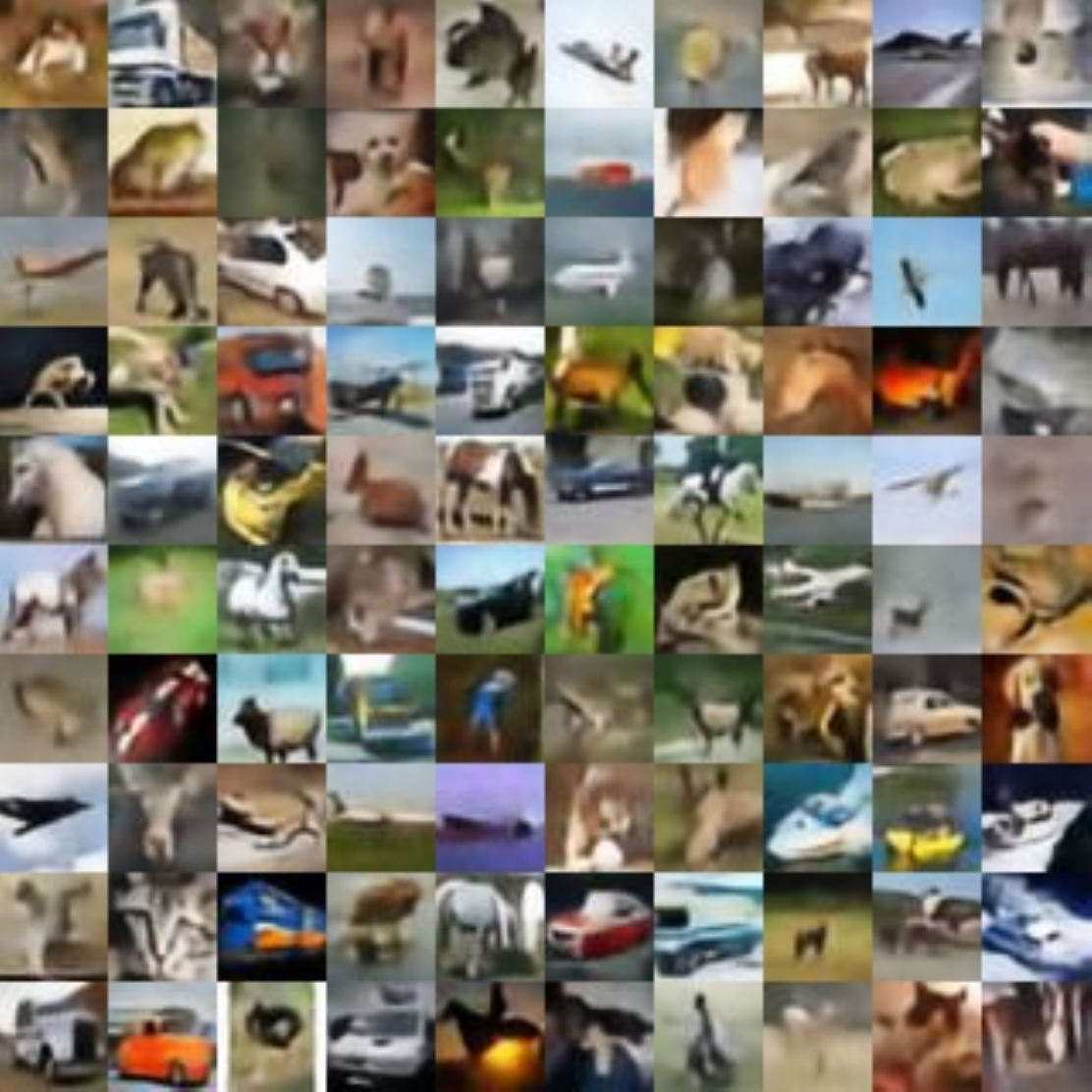}
        \caption{VAE (Learnable $\gamma$)}
    \end{subfigure}
    \caption{Reconstructions on CIFAR-10 Dataset.}
    \label{fig:recon_cifar10}
\end{figure}

Reconstruction results for MNIST, Fashion-MNIST, CIFAR-10 and CelebA datasets are shown in Figures~\ref{fig:recon_celeba}$-$\ref{fig:recon_cifar10} respectively. On relatively simple datasets like MNIST and Fashion-MNIST, the VAE with learnable $\gamma$ achieves almost exact reconstruction because of a better estimate of the underlying manifold consistent with theory. However, the VAE with fixed $\gamma = 1$  produces blurry reconstructions as expected. Note that the reconstruction of a 2-Stage VAE is the same as that of a VAE with learnable $\gamma$ because the second-stage VAE has nothing to do with facilitating the reconstruction task.

\clearpage

\begin{figure}[t!]
    \centering
    \begin{subfigure}[t]{0.45\textwidth}
        \centering
        \includegraphics[width=1\linewidth]{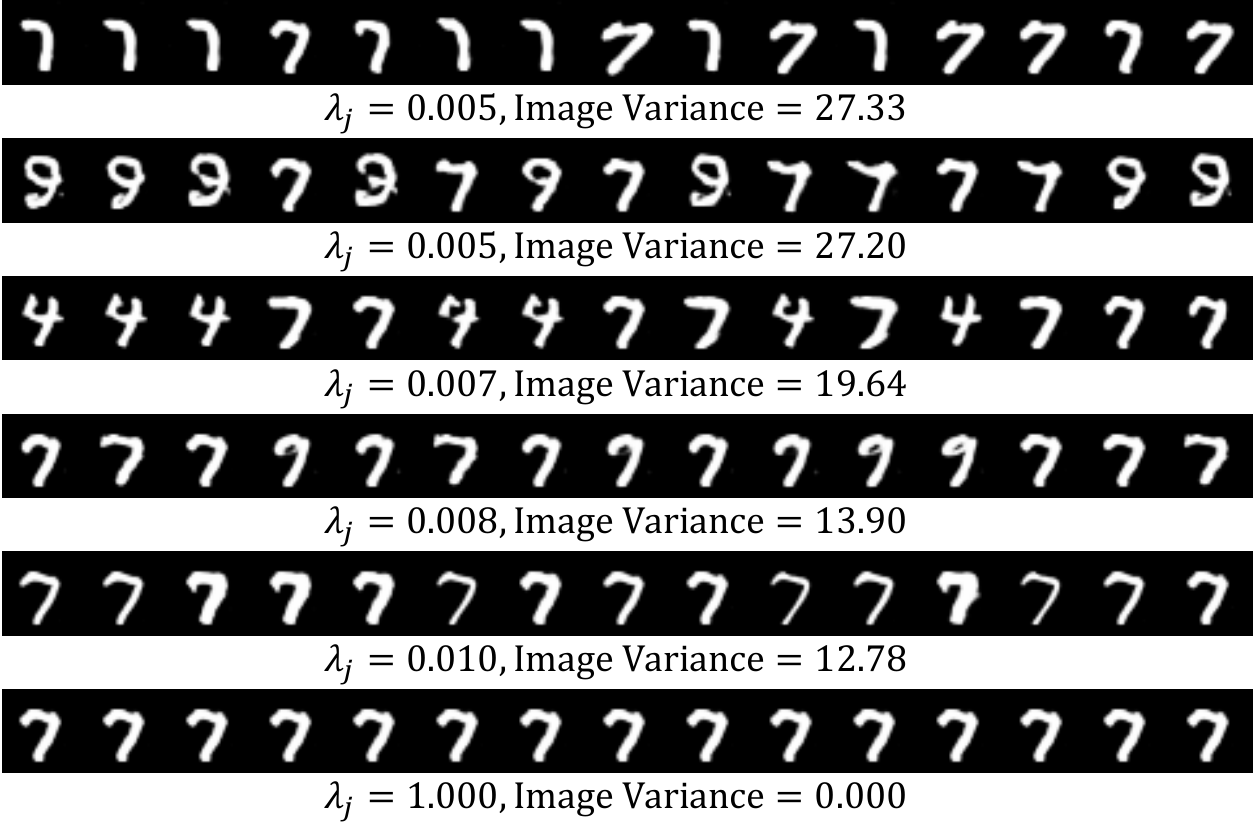}
        \caption{MNIST}
    \end{subfigure}
    \begin{subfigure}[t]{0.45\textwidth}
        \centering
        \includegraphics[width=1\linewidth]{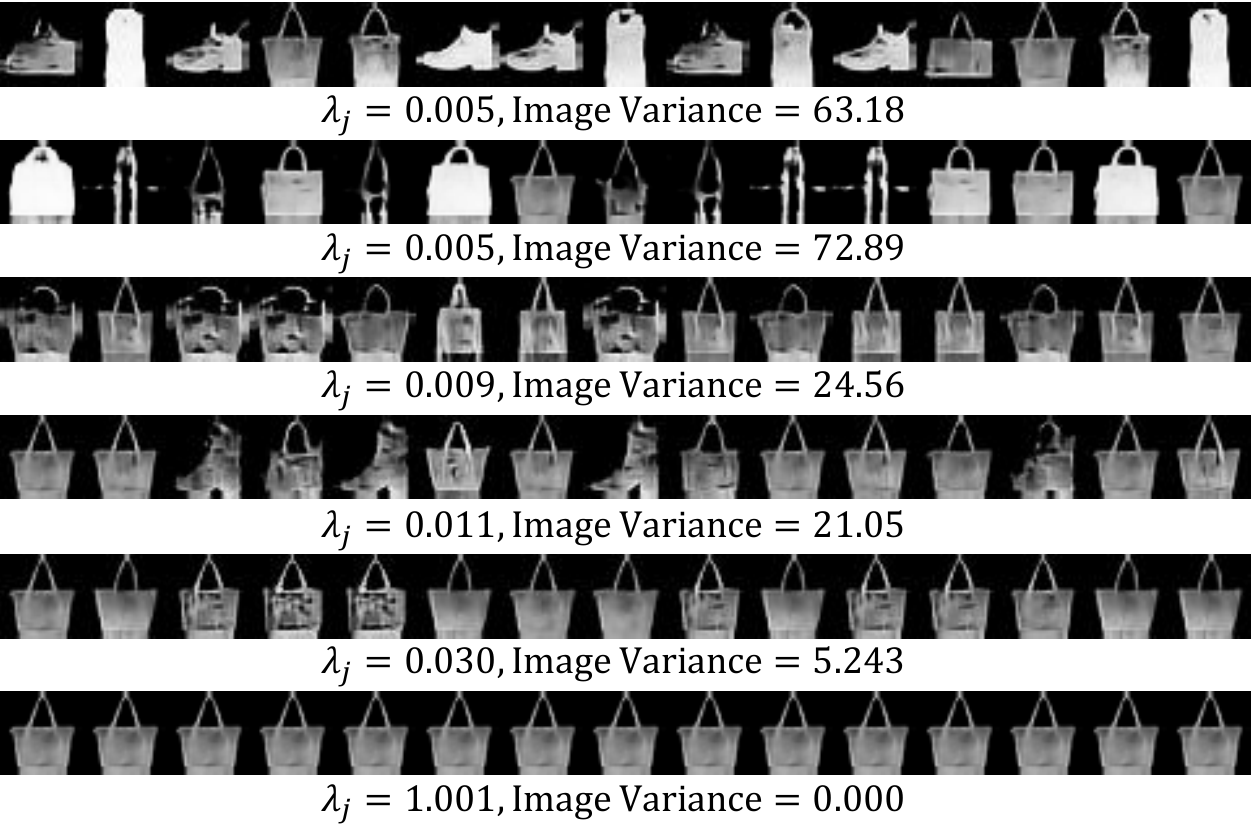}
        \caption{Fashion-MNIST}
    \end{subfigure}
    \begin{subfigure}[t]{0.9\textwidth}
        \includegraphics[width=1\linewidth]{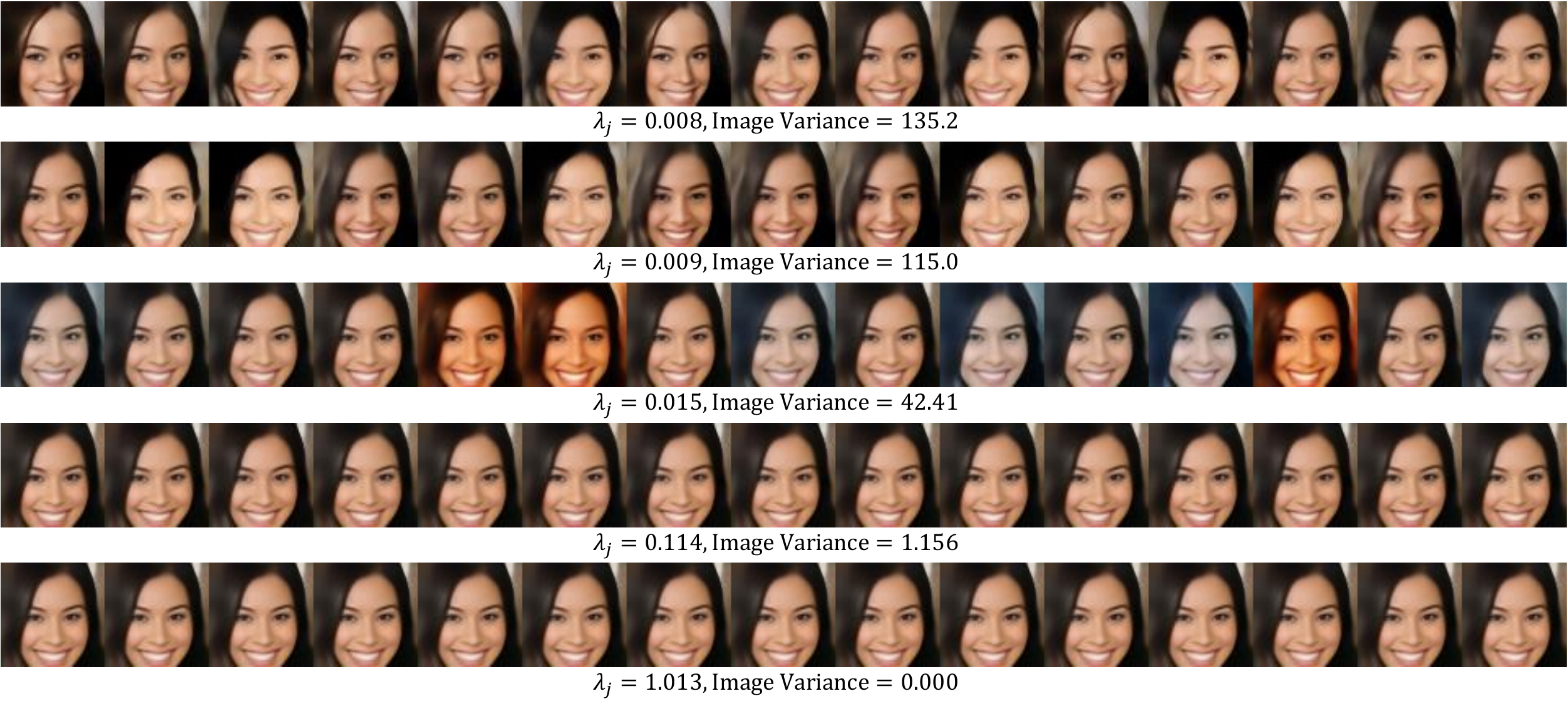}
        \caption{CelebA}
    \end{subfigure}
    \caption{More examples similar to Figure~\ref{fig:decoder_mean_main}.}
    \label{fig:decoder_mean}
\end{figure}

\clearpage

\section{Additional Experimental Results Validating Theoretical Predictions}\label{sec:experiment_theory}

We first present more examples similar to Figure~\ref{fig:decoder_mean_main} from the main paper. Random noise is added to $\bmu_z$ along different directions and the result is passed through the decoder network. Each row corresponds to a certain direction in the latent space and $15$ samples are shown for each direction. These dimensions/rows are ordered by the eigenvalues $\lambda_j$ of $\bSigma_z$. The larger $\lambda_j$ is, the less impact a random perturbation along this direction will have as quantified by the reported image variance values. In the first two or three rows, the noise generates some images from different classes/objects/identities, indicating a significant visual difference. For a slightly larger $\lambda_j$, the corresponding dimensions encode relatively less significant attributes as predicted. For example, the fifth row of both MNIST and Fashion-MNIST contains images from the same class but with a slightly different style. The images in the fourth row of the CelebA dataset have very subtle differences. When $\lambda_j=1$, the corresponding  dimensions become completely inactive and all the output images are exactly the same, as shown in the last rows for all the three datasets.

\begin{figure}[t!]
    \centering
    \begin{subfigure}[t]{0.45\textwidth}
        \centering
        \includegraphics[width=1\linewidth]{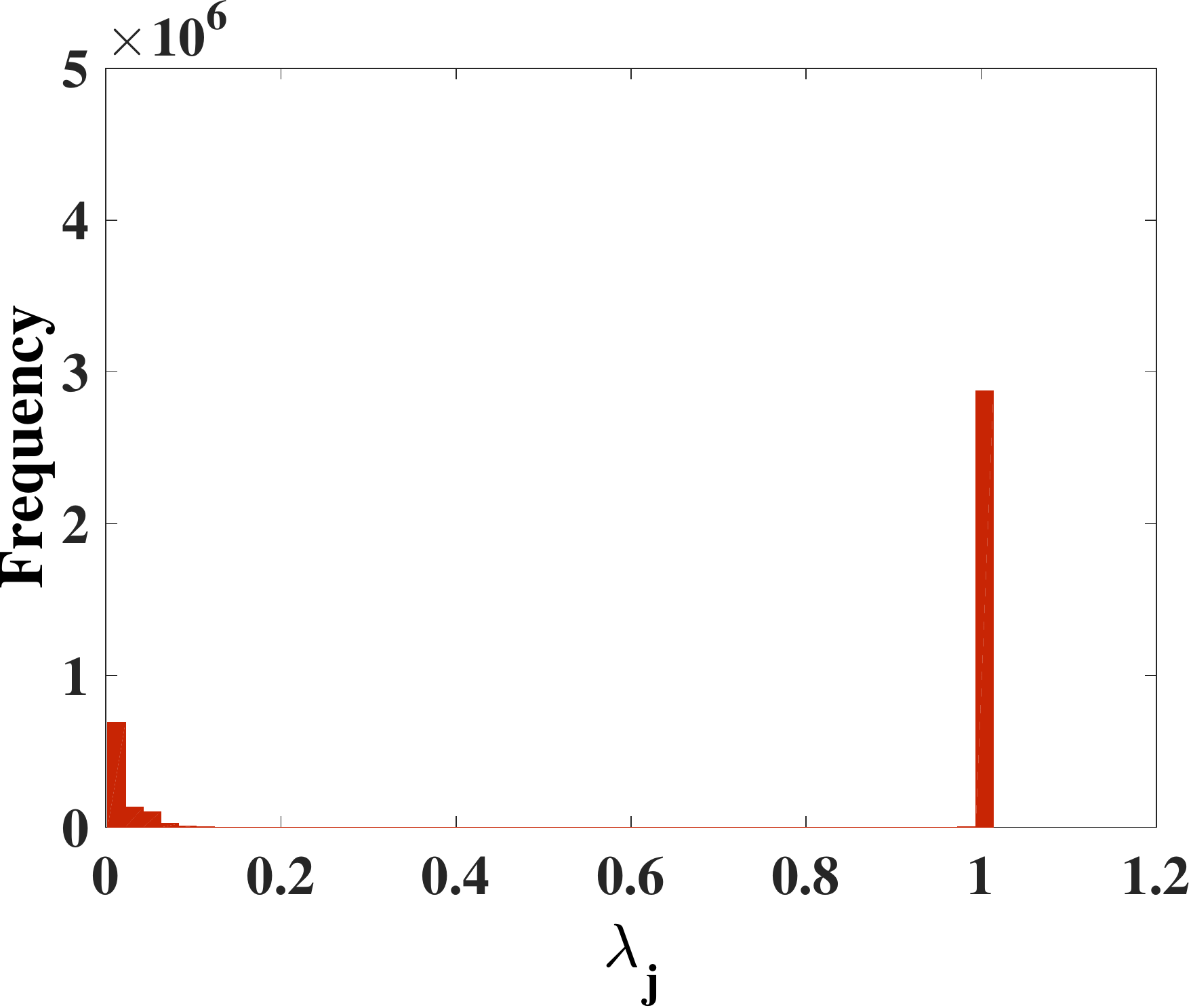}
        \caption{Hist of $\lambda_j$ on MNIST}
    \end{subfigure}
    \begin{subfigure}[t]{0.45\textwidth}
        \centering
        \includegraphics[width=1\linewidth]{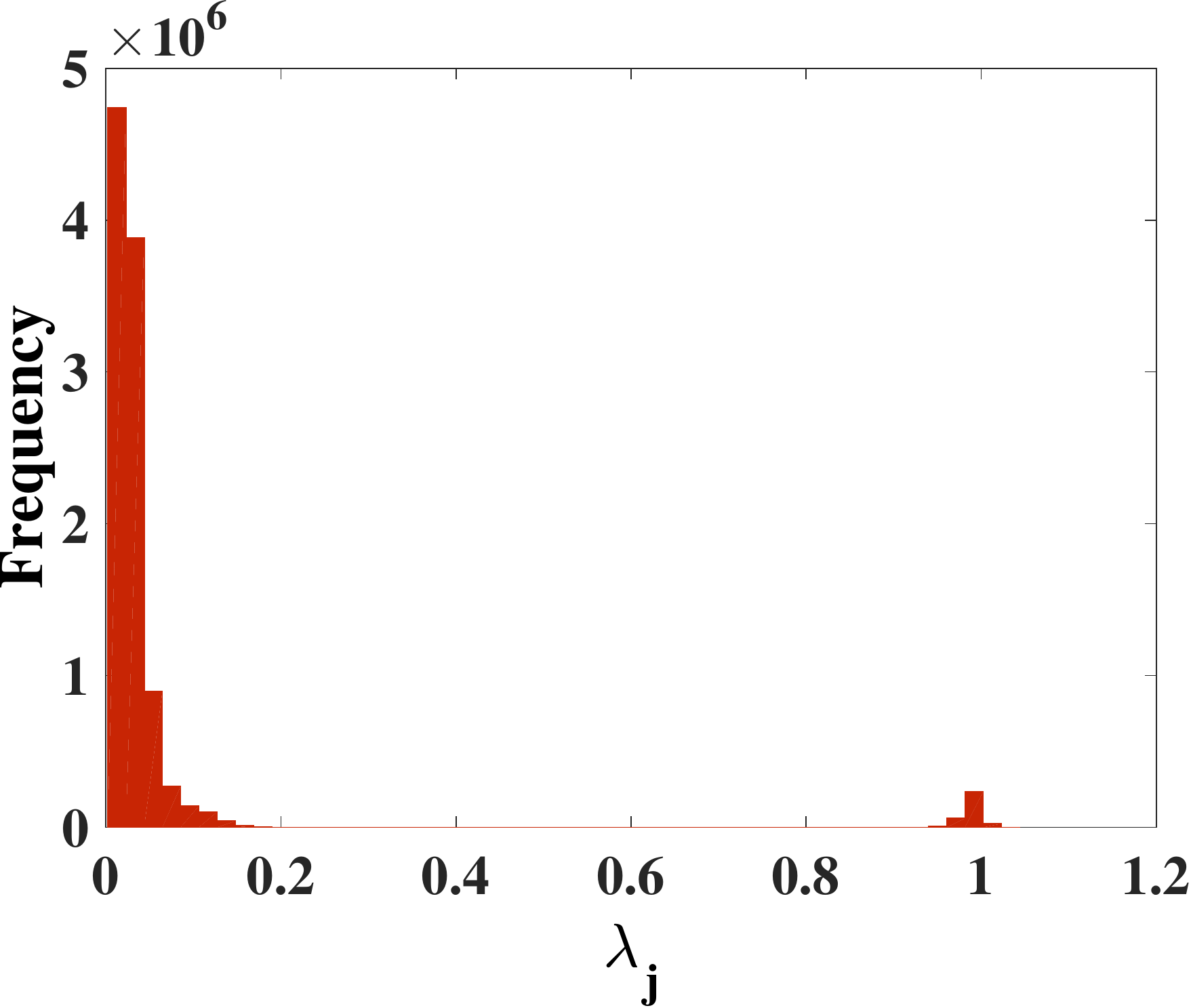}
        \caption{Hist of $\lambda_j$ on CelebA}
    \end{subfigure}
    \caption{Histogram of $\lambda_j$ values. There are more values around $0$ for CelebA because it is more complicated than MNIST and therefore requres more active dimensions to model the underlying manifold.}
    \label{fig:sd_hist}
\end{figure}

Additionally, as discussed in the main text and below in Section~\ref{sec:analysis_decoder_mean}, there are likely to be $r$ eigenvalues of $\bSigma_z$ converging to zero and $\kappa-r$ eigenvalues  converging to one.  We plot the histogram of $\lambda_j$ values for both MNIST and CelebA datasets in Figure~\ref{fig:sd_hist}. For both datasets, $\lambda_j$ approximately converges to either to zero or one. However, since CelebA is a more complicated dataset than MNIST, the ground-truth manifold dimension of CelebA is likely to be much larger than that of MNIST. So more eigenvalues are expected to be near zero for the CelebA dataset. This is indeed the case, demonstrating that VAE has the ability to detect the manifold dimension and select the proper number of latent dimensions in practical environments.

\section{Network Structure and Experimental Settings}\label{sec:experiments_setting}
\begin{figure}[t!]
    \centering
    \includegraphics[width=0.8\linewidth]{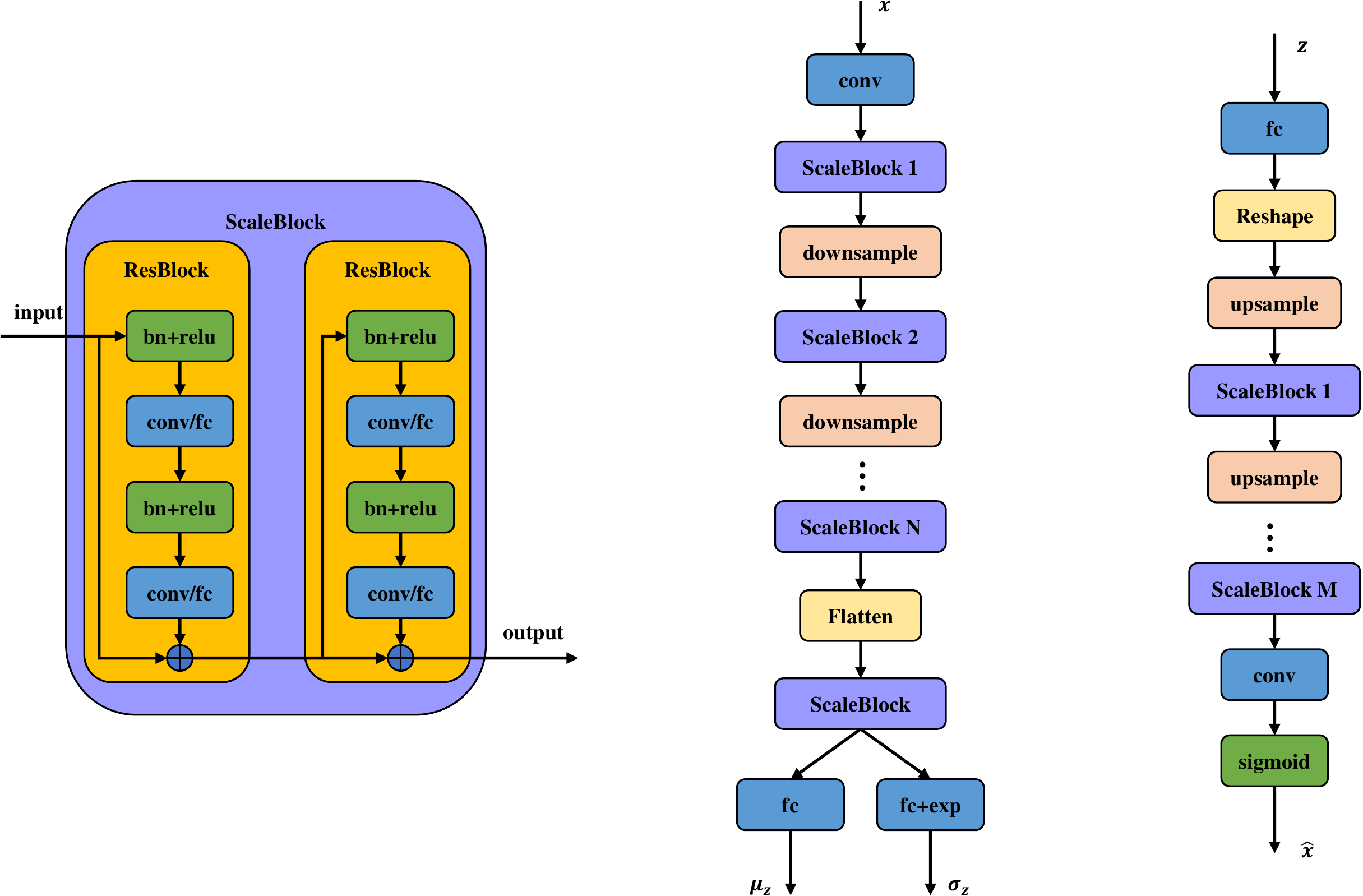}
    \caption{Network structure of the first-stage VAE used in producing Figure \ref{fig:gen_samples}, and for generating samples and reconstructions.  (\emph{Left}) The basic building block of the network called a Scale Block, which consists of two Residual Blocks. (\emph{Center}) The encoder network. For an input image $\bx$, we use a convolutional layer to transform it into $32$ channels. We then pass it to a Scale Block. After each Scale Block, we downsample using a convolutional layer with stride $2$ and double the channels. After $N$ Scale Blocks, the feature map is flattened to a vector. In our experiments, we used $N=4$ for CelebA dataset and $3$ for other datasets. The vector is then passed through another Scale Block, the convolutional layers of which are replaced with fully connected layers of $512$ dimensions. The output of this Scale Block is used to produce the $\kappa$-dimensional latent code, with $\kappa = 64$. (\emph{Right}) The decoder network. A latent code $\bz$ is first passed through a fully connected layer. The dimension is $4096$ for CelebA dataset and $2048$ for other datasets. Then it is reshaped to $2\times2$ resolution. We upsample the feature map using a transposed convolution layer and half the number of channels at the same time. It then goes through some Scale Blocks and upsampling layers until the feature map size becomes the desired value. Then we use a convolutional layer to transform the feature map, which should have $32$ channels, to $3$ channels for RGB datasets and $1$ channel for gray scale datasets.}
    \label{fig:network_structure}
\end{figure}

We first describe the network and training details used in producing Figure \ref{fig:gen_samples} from the main file, and for generating samples and reconstructions in the appendix. The first-stage VAE network is shown in Figure~\ref{fig:network_structure}. Basically we use two Residual Blocks for each resolution scale, and we double the number of channels when downsampling and halve it when upsampling. The specific settings such as the number of channels and the number of scales are specified in the caption.   The second VAE is much simpler. Both the encoder and decoder have three $2048$-dimensional hidden layers.  Finally, the training details are presented below.  Note that these settings were not tuned, we simply chose more epochs for more complex data sets and fewer for datasets with larger training samples.  For each dataset just a single setting was tested as follows:
\begin{itemize}
\item \textbf{MNIST and Fashion-MNIST: } The batch size is specified to be 100. We use the ADAM optimizer with the default hyperparameters in TensorFlow.  The first VAE is trained for $400$ epochs. The initial learning rate is $0.0001$ and we halve it every $150$ epochs. The second VAE is trained for $800$ epochs with the same initial learning rate, halved every $300$ epochs.

\item \textbf{CIFAR-10: } Since CIFAR-10 is more complicated than MNIST and Fashion-MNIST, we use more epochs for training. Specifically, we use $1000$ and $2000$ epochs for the two VAEs respectively and half the learning rate every $300$ and $600$ epochs for the two stages. The other settings are the same as that for MNIST.

\item \textbf{CelebA: } Because CelebA has many more examples, in the first stage we train $120$ epochs and half the learning rate every $48$ epochs. In the second stage, we train $300$ epochs and half the learning rate every $120$ epochs. The other settings are the same as that for MNIST, etc.
\end{itemize}

Finally, to fairly compare against various GAN models and VAE baselines using FID scores on a neutral architecture (i.e., the results from Table~\ref{table:fid}), we simply adopt the InfoGAN network structure consistent with the neutral setup from \citep{lucic2018gans} for the first-stage VAE.  For the second-stage VAE we just use three $1024$-dimensional hidden layers, which contribute less than $5\%$ to the total number of parameters.  
Note that the small number of additional parameters contributing to the second stage do not improve the other VAE baselines when aggregated and trained jointly.

\section{Proof of Theorem~\ref{thm:optima_r_eq_d}}\label{sec:proof_optima_r_eq_d}

\begin{figure}[h]
    \centering
    \includegraphics[width=0.7\linewidth]{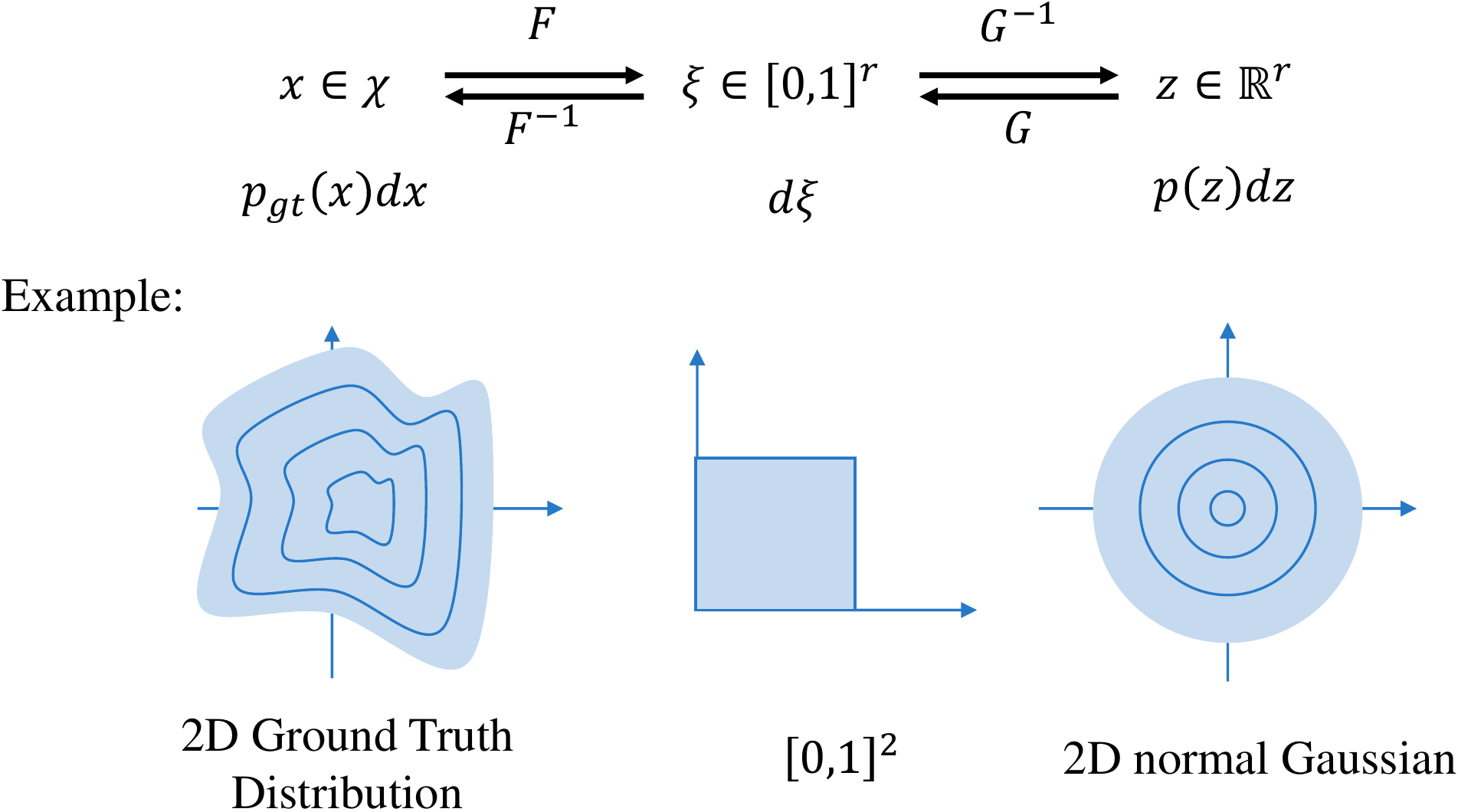}
    \caption{The relationship between different variables.}
    \label{fig:variable_graph}
\end{figure}

We first consider the case where the latent dimension $\kappa$ equals the manifold dimension $r$ and then extend the proof to allow for $\kappa>r$. The intuition is to build a bijection between $\bchi$ and $\mathbb{R}^r$ that transforms the ground-truth distribution $p_{gt}(\bx)$ to a normal Gaussian distribution. The way to build such a bijection is shown in Figure~\ref{fig:variable_graph}. We now fill in the details.

\subsection{Finding a Sequence of Decoders such that $p_{\theta^*_t}(\bx)$ Converges to $p_{gt}(\bx)$}


Define the function $F:\mathbb{R}^r\mapsto[0,1]^r$ as
\begin{eqnarray}
    F(\bx) &=& \left[F_1(\bx_1), F_2(\bx_2;\bx_1),...,F_r(\bx_r;\bx_{1:r-1})\right]^\top, \\
    F_i(\bx_i;\bx_{1:i-1}) &=& \int_{\bx_i^\prime=-\infty}^{\bx_i} p_{gt}(\bx_i^\prime|\bx_{1:i-1})d\bx_i^\prime.
\end{eqnarray}
Per this definition, we have that
\begin{equation}
	dF(\bx) = p_{gt}(\bx) d\bx.
\end{equation}
Also, since $p_{gt}(\bx)$ is nonzero everywhere, $F(\cdot)$ is invertible.  Similarly, we define another differentiable and invertible function $G:\mathbb{R}^r\mapsto[0,1]^r$ as
\begin{eqnarray}
    G(\bz) &=& [G_1(\bz_1),G_2(\bz_2),...,G_r(\bz_r)]^\top, \\
    G_i(\bz_i) &=& \int_{\bz_i^\prime=-\infty}^{\bz_i} \mathcal{N}(\bz_i|0,1) d\bz_i^\prime. \label{eqn:def_G}
\end{eqnarray}
Then
\begin{equation}
	dG(\bz) = p(\bz) d\bz = \mathcal{N}(\bz|0,\bI) d\bz.
\end{equation}
Now let the decoder be
\begin{eqnarray}
	f_{\mu_x}(\bz;\theta^*_t) &=& F^{-1} \circ G(z), \label{eqn:decoder_mean} \\
	\gamma_t^* &=& \frac{1}{t}. \label{eqn:decoder_var}
\end{eqnarray}
Then we have
\begin{equation}
    p_{\theta^*_t}(\bx) ~=~ \int_{\mathbb{R}^r} p_{\theta^*_t}(\bx|\bz)p(\bz) d\bz ~=~ \int_{\mathbb{R}^r} \mathcal{N}\left( \bx | F^{-1} \circ G(\bz), \gamma^*_t\bI \right) dG(\bz). \label{eqn:p_x}
\end{equation}
Additionally, let $\bxi=G(\bz)$ such that
\begin{equation}
	p_{\theta^*_t}(\bx) = \int_{[0,1]^r} \mathcal{N}\left( \bx | F^{-1} (\bxi), \gamma^*_t\bI \right) d\bxi,
\end{equation}
and let $\bx^\prime=F^{-1}(\bxi)$ such that $d\bxi=dF(\bx^\prime)=p_{gt}(\bx^\prime)d\bx^\prime$. Plugging this expression into the previous $p_{\theta^*}(\bx)$ we obtain
\begin{equation}
	p_{\theta^*_t}(\bx) = \int_{\mathbb{R}^r} \mathcal{N}\left( \bx | \bx^\prime, \gamma^*_t\bI \right) p_{gt}(\bx^\prime) d\bx^\prime. \label{eqn:p_x_trans}
\end{equation}
As $t\to\infty$, $\gamma^*_t$ becomes infinitely small and $\mathcal{N}\left(\bx|\bx^\prime,\gamma^*_t\bI\right)$ becomes a Dirac-delta function, resulting in
\begin{equation}
	\lim_{t\to\infty} p_{\theta^*_t}(\bx) ~=~ \int_{\bchi} \delta(\bx^\prime-\bx) p_{gt}(\bx^\prime) d\bx^\prime ~=~ p_{gt}(\bx). \label{eqn:lim_px}
\end{equation}

\subsection{Finding a Sequence of Encoders such that $\mathbb{KL}\left[q_{\phi^*_t}(\bz|\bx)||p_{\theta^*_t}(\bz|\bx)\right]$ Converges to $0$}\label{sec:proof_sufficient_full_kl}
Assume the encoder networks satisfy
\begin{eqnarray}
    f_{\mu_z}(\bx;\phi^*_t) &=& G^{-1}\circ F(\bx) = f_{\mu_x}^{-1}(\bx;\theta^*_t), \label{eqn:encoder_mean}\\
    f_{S_z}(\bx;\phi^*_t) &=& \sqrt{\gamma^*_t \left( f^\prime_{\mu_x}\left(f_{\mu_z}(\bx;\phi^*_t);\theta^*_t\right)^\top f^\prime_{\mu_x}\left(f_{\mu_z}(\bx;\phi^*_t);\theta^*_t\right) \right)^{-1}}, \label{eqn:encoder_var}
\end{eqnarray}
where $f^\prime_{\mu_x}(\cdot)$ is a $d\times r$ Jacobian matrix. We omit the arguments $\theta^*_t$ and $\phi^*_t$ in $f_{\mu_z}(\cdot)$, $f_{S_z}(\cdot)$ and $f_{\mu_x}(\cdot)$ hereafter to avoid unnecessary clutter. We first explain why $f_{\mu_x}(\cdot)$ is differentiable. Since $f_{\mu_x}(\cdot)$ is a composition of $F^{-1}(\cdot)$ and $G(\cdot)$ according to (\ref{eqn:decoder_mean}), we only need to explain that both functions are differentiable. For $F^{-1}(\cdot)$, it is the inverse of a differentiable function $F(\cdot)$. Moreover, the derivative of $F(\bx)$ is $p_{gt}(\bx)$, which is nonzero everywhere. So $F^{-1}(\cdot)$ and therefore $f_{\mu_x}(\cdot)$ are both differentiable.

The true posterior $p_{\theta^*_t}(\bz|\bx)$ and the approximate posterior are
\begin{eqnarray}
    p_{\theta^*_t}(\bz|\bx) &=& \frac{\mathcal{N}(\bz|0,\bI)\mathcal{N}(\bx|f_{\mu_x}(\bz),\gamma^*_t\bI)}{p_{\theta^*_t}(\bx)}, \\
    q_{\phi^*_t}(\bz|\bx) &=& \mathcal{N}\left( \bz|f_{\mu_z}(\bx), \gamma^*_t \left( f^\prime_{\mu_x}\left(f_{\mu_z}(\bx)\right)^\top f^\prime_{\mu_x}\left(f_{\mu_z}(\bx)\right) \right)^{-1} \right)
\end{eqnarray}
respectively. We first transform $\bz$ to $\bz^\prime$ via
\begin{equation}
    \bz^\prime = (\bz - \bz^*) / \sqrt{\gamma_t^*},
\end{equation}
where $\bz^*=f_{\mu_z}(\bx)$. The true and approximate posteriors $p_{\theta^*_t}(\bz|\bx)$ and $q_{\phi^*_t}(\bz|\bx)$ will then be transformed into another two distributions of $\bz^\prime$, namely $p^\prime_{\theta^*_t}(\bz^\prime|\bx)$ and $q^\prime_{\phi^*_t}(\bz|\bx)$. The KL divergence between $p_{\theta^*_t}(\bz|\bx)$ and $q_{\phi^*_t}(\bz|\bx)$ will be the same as that between $p^\prime_{\theta^*_t}(\bz^\prime|\bx)$ and $q^\prime_{\phi^*_t}(\bz|\bx)$.
We now prove that $q^\prime_{\phi^*_t}(\bz^\prime|\bx)/p^\prime_{\theta^*_t}(\bz^\prime|\bx)$ converges to a constant not related to $\bz^\prime$ as $t$ goes to $\infty$. If this is true, the constant must be $1$ since both $q^\prime_{\phi^*_t}(\bz^\prime|\bx)$ and $p^\prime_{\theta^*_t}(\bz^\prime|\bx)$ are probability distributions. Then the KL divergence between them converges to $0$ as $t\to\infty$.

We denote $\left( f^\prime_{\mu_x}\left(f_{\mu_z}(\bx)\right)^\top f^\prime_{\mu_x}\left(f_{\mu_z}(\bx)\right) \right)^{-1}$ as $\tilde{\bSigma}_z(\bx)$ for short. Given this definitions, it follows that
\begin{eqnarray}
    \frac{q^\prime_{\phi^*_t}(\bz^\prime|\bx)}{p^\prime_{\theta^*_t}(\bz^\prime|\bx)} &=& \frac{\mathcal{N}\left(\bz^* + \sqrt{\gamma_t^*}\bz^\prime | \bz^*, \gamma^*_t \tilde{\bSigma}_z \right) p_{\theta^*_t}(\bx)}{\mathcal{N}(\bz^*+\sqrt{\gamma_t^*}\bz^\prime|0,I) \mathcal{N}(\bx|f_{\mu_x}(\bz^*+\sqrt{\gamma_t^*}\bz^\prime),\gamma^*_t\bI)} \nonumber \\
    &=& (2\pi)^{d/2} {\gamma^*_t}^{(d-r)/2} \left|\tilde{\bSigma}_z\right|^{-1/2} \exp\left\{ - \frac{ { \bz^\prime }^\top \tilde{\bSigma}_z^{-1} { \bz^\prime } }{2\gamma^*_t} \right. \nonumber \\
    && ~~~~~~~~\left. + \frac{||\bz^*+\sqrt{\gamma_t^*}\bz^\prime||_2^2}{2} + \frac{||\bx-f_{\mu_x}(\bz^*+\sqrt{\gamma_t^*}\bz^\prime)||_2^2}{2\gamma^*_t} \right\} p_{\theta^*_t}(\bx). \label{eqn:q_p_ratio}
\end{eqnarray}

We use $C(\bx)$ to represent the terms not related to $\bz^\prime$, \emph{i.e.}, $(2\pi)^{d/2}{\gamma_t^*}^{(d-r)/2}|\tilde{\bSigma_z}|^{-1/2}p_{\theta_t^*}(\bx)$ and consider the limiting ratio 

\begin{eqnarray}
    \lim_{t\to\infty} \frac{q^\prime_{\phi^*_t}(\bz^\prime|\bx)}{p^\prime_{\theta^*_t}(\bz^\prime|\bx)} &=& \lim_{t\to\infty} C(\bx) \exp\left\{ -\frac{{\bz^\prime}^\top \tilde{\bSigma_z}^{-1} {\bz^\prime}}{2} + \frac{||\bz^*+\sqrt{\gamma^*_t}{\bz^\prime}||_2^2}{2} + \frac{|| \bx - f_{\mu_x}\left(\bz^*+\sqrt{\gamma^*_t}{\bz^\prime}\right) ||_2^2}{2\gamma_t^*}\right\} \nonumber \\
    &=& \lim_{t\to\infty} C(\bx) \exp\left\{ -\frac{{\bz^\prime}^\top \tilde{\bSigma_z}^{-1} {\bz^\prime}}{2} + \frac{||\bz^*||_2^2}{2} + \frac{||\sqrt{\gamma_t^*} f_{\mu_x}^\prime\left(\bz^*\right) {\bz^\prime} ||_2^2}{2\gamma_t^*}\right\} \nonumber \\
    &=& C(\bx) \exp\left\{ -\frac{{\bz^\prime}^\top \tilde{\bSigma_z}^{-1} {\bz^\prime}}{2} + \frac{||\bz^*||_2^2}{2} + \frac{{\bz^\prime}^\top f^\prime_{\mu_x}\left(\bz^*\right)^\top f^\prime_{\mu_x}\left(\bz^*\right) {\bz^\prime}}{2}\right\} \nonumber \\
    &=& C(\bx) \exp\left\{\frac{||\bz^*||_2^2}{2}\right\}.
\end{eqnarray}
The fourth equality comes from the fact that $f^\prime_{\mu_x}\left(\bz^*\right)^\top f^\prime_{\mu_x}\left(\bz^*\right) = f^\prime_{\mu_x}\left(f_{\mu_z}(\bx)\right)^\top f^\prime_{\mu_x}\left(f_{\mu_z}(\bx)\right) = \tilde{\bSigma}_z(\bx)^{-1}$. This expression is not related to $\bz^\prime$. Considering both $q^\prime_{\phi^*_t}(\bz^\prime|\bx)$ and $p^\prime_{\theta^*_t}(\bz^\prime|\bx)$ are probability distributions, the ratio should be equal to $1$. The KL divergence between them thus converges to $0$ as $t\to\infty$.

\subsection{Generalization to the Case with $\kappa>r$}\label{sec:proof_sufficient_full_generalize}
When $\kappa>r$, we use the first $r$ latent dimensions to build a projection between $\bz$ and $\bx$ and leave the remaining $\kappa-r$ latent dimensions unused. Specifically, let $f_{\mu_x}(\bz)=\tilde{f}_{\mu_x}(\bz_{1:r})$, where $\tilde{f}_{\mu_x}(\bz_{1:r})$ is defined as in (\ref{eqn:decoder_mean}) and $\gamma^*_t=1/t$. Again consider the case that $t\to\infty$. Then this decoder can also satisfy $\lim_{t\to\infty}p_{\theta^*_t}(\bx) ~=~ p_{gt}(\bx)$ because it produces exactly the same distribution as the decoder defined by (\ref{eqn:decoder_mean}) and (\ref{eqn:decoder_var}). The last $\kappa-r$ dimensions contribute nothing to the generation process.

Now define the encoder as
\begin{eqnarray}
    f_{\mu_z}(\bx)_{1:r} &=& \tilde{f}_{\mu_x}^{-1}(\bx) \\
    f_{\mu_z}(\bx)_{r+1:\kappa} &=& 0 \\
    f_{S_z}(\bx) &=& \begin{bmatrix}\tilde{f}_{S_z}(\bx) \\ \bn_{r+1}^\top \\ ... \\ \bn_{\kappa}^\top \end{bmatrix}
\end{eqnarray}
where $\tilde{f}_{S_z}(\bx)$ is defined as (\ref{eqn:encoder_var}). Denote $\{\bn_i\}_{i=r+1}^\kappa$ as a set of $\kappa$-dimensional column vectors satisfying
\begin{eqnarray}
    \tilde{f}_{S_z}(\bx)\bn_i &=& 0 \label{eqn:n_condition1} \\
    \bn_i^\top\bn_j &=& \bold{1}_{i=j} \label{eqn:n_condition2}
\end{eqnarray}
Such a set always exists because $\tilde{f}_{S_z}(\bx)$ is a $r\times\kappa$ matrix. So the dimension of the null space of $\tilde{f}_{S_z}(\bx)$ is at least $\kappa-r$. Assuming that $\{\bn_i\}_{i=r+1}^\kappa$ are $\kappa-r$ basis vectors of $\text{null}(\tilde{f}_{S_z})$, then the conditions (\ref{eqn:n_condition1}) and (\ref{eqn:n_condition2}) will be satisfied. The variance of the approximate posterior then becomes
\begin{equation}
    \bSigma_z ~=~ f_{S_z}(\bx)f_{S_z}(\bx)^\top ~=~ \begin{bmatrix} \tilde{f}_{S_z}(\bx)\tilde{f}_{S_z}(\bx)^\top & 0 \\ 0 & \bI_{\kappa-r} \end{bmatrix}
\end{equation}
The first $r$ dimensions can exactly match the true posterior as we have already shown. The remaining $\kappa-r$ dimensions follow a standardized Gaussian distribution. Since these dimensions contribute nothing to generating $\bx$, the true posterior should be the same as the prior, \emph{i.e.} a standardized Gaussian distribution. Moreover, any of these dimensions is independent of all the other dimensions, so the corresponding off-diagonal elements of the covariance of the true posterior should equal $0$. Thus the approximate posterior also matches the true posterior for the last $\kappa-r$ dimensions. As a result, we again have $\lim_{t\to\infty} ~\mathbb{KL}\left[q_{\phi^*_t}(\bz|\bx)||p_{\theta^*_t}(\bz|\bx)\right] ~=~0$.

\section{Proof of Theorem~\ref{thm:optima_r_less_d}}\label{sec:proof_optima_r_less_d}

\begin{figure}[h]
    \centering
    \includegraphics[width=0.9\linewidth]{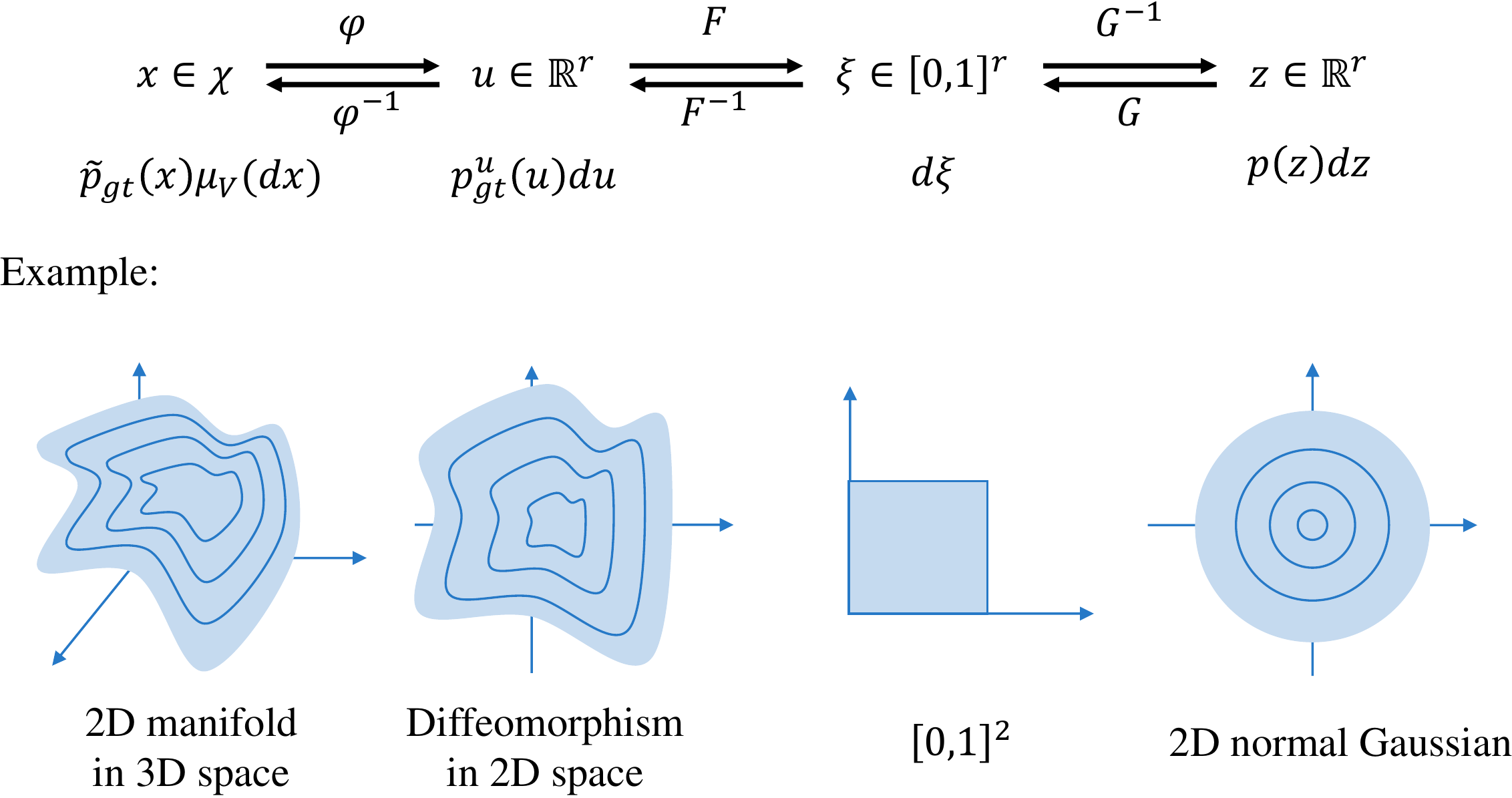}
    \caption{The relationship between different variables.}
    \label{fig:variable_graph2}
\end{figure}

Similar to Section~\ref{sec:proof_optima_r_eq_d}, we also construct a bijection between $\bchi$ and $\mathbb{R}^r$ which transforms the ground-truth measure $\mu_{gt}$ to a normal Gaussian distribution. But in this construction, we need one more step that bijects between $\bchi$ and $\mathbb{R}^r$ using the diffeomorphism $\varphi(\cdot)$, as shown in Figure~\ref{fig:variable_graph2}. We will now go into the details.

\subsection{Finding a Sequence of Decoders such that $-\log p_{\theta^*_t}(\bx)$ Converges to $-\infty$}

$\varphi(\cdot)$ is a diffeomorphism between $\bchi$ and $\mathbb{R}^r$. So it transforms the ground-truth probability distribution $\tilde{p}_{gt}(\bx)$ to another distribution $p_{gt}^u(\bu)$, where $\bu\in\mathbb{R}^r$. The relationship between the two distributions is
\begin{equation}
    p_{gt}^u(\bu)d\bu ~=~ \tilde{p}_{gt}(\bx)\mu_V(d\bx)|_{\bx=\varphi^{-1}(\bu)} ~=~ \mu_{gt}(d\bx),
\end{equation}
where $\mu_V(d\bx)$ is the volume measure with respect to $\calX$. Because $\varphi(\cdot)$ is a diffeomorphism, both $\varphi(\cdot)$ and $\varphi^{-1}(\cdot)$ are differentiable. Thus $d\bx/d\bu$ is nonzero everywhere on the manifold. Considering $\tilde{p}_{gt}(\bx)$ is also nonzero everywhere, $p_{gt}^u(\bu)$ is nonzero everywhere.

Analogous to the previous proof, define a function $F:\mathbb{R}^r\mapsto[0,1]^r$ as
\begin{eqnarray}
    F(\bu) &=& \left[F_1(\bu_1), F_2(\bu_2;\bu_1),...,F_r(\bu_r;\bu_{1:r-1})\right]^\top, \\
    F_i(\bu_i;\bu_{1:i-1}) &=& \int_{\bu_i^\prime=-\infty}^{\bu_i} p_{gt}^u(\bu_i^\prime|\bu_{1:i-1})d\bu_i^\prime.
\end{eqnarray}
According to this definition, we have
\begin{equation}
	dF(\bu) = p_{gt}^u(\bu) d\bu.
\end{equation}
Since $p_{gt}^u(\bu)$ is nonzero everywhere, $F(\cdot)$ is invertible.  We also define another differentiable and invertible function $G:\mathbb{R}^r\mapsto[0,1]^r$ as (\ref{eqn:def_G}).

Now let the decoder mean function be given by
\begin{eqnarray}
	f_{\mu_x}(\bz;\theta^*_t) &=& \varphi^{-1} \circ F^{-1} \circ G(z), \label{eqn:decoder_mean2} \\
	\gamma_t^* &=& \frac{1}{t}. \label{eqn:decoder_var2}
\end{eqnarray}
Then we have
\begin{eqnarray}
	p_{\theta^*_t}(\bx) &=& \int_{\mathbb{R}^r} p_{\theta^*_t}(\bx|\bz)p(\bz) d\bz \nonumber \\
	&=& \int_{\mathbb{R}^r} \mathcal{N}\left( \bx | \varphi^{-1}(\bu), \gamma^*_t\bI \right) p_{gt}^u(\bu) d\bu.
\end{eqnarray}
We next show that $p_{\theta^*_t}(\bx)$ diverges to infinite as $t\to\infty$ for any $\bx$.  For a given $\bx$, let $\bu^*=\varphi(\bx)$ and $B(\bu^*,\sqrt{\gamma^*_t})$ be the closed ball centered at $\bu^*$ with radius $\sqrt{\gamma^*_t}$. Then
\begin{eqnarray}
	p_{\theta^*_t}(\bx) &\ge& \int_{B(\bu^*,\sqrt{\gamma^*_t})} \mathcal{N}\left( \bx | \varphi^{-1}(\bu), \gamma^*_t\bI \right) p_{gt}^u(\bu) d\bu \nonumber \\
	&=& \int_{B(\bu^*,\gamma^*_t)} (2\pi\gamma^*_t)^{-d/2}\exp \left\{ -\frac{ \left|\left|\bx-\varphi^{-1}(\bu)\right|\right|_2^2 }{ 2\gamma^*_t } \right\} p_{gt}^u(\bu) d\bu. \label{eqn:p_x_bound}
\end{eqnarray}
According to the Lagrangian's mean value theorem, there exists a $\bu^\prime$ between $\bu$ and $\bu^*$ such that
\begin{equation}
	\varphi^{-1}(\bu) ~=~ \varphi^{-1}(\bu^*) + \frac{d\varphi^{-1}(\bu)}{d\bu}|_{\bu=\bu^\prime} (\bu-\bu^*) ~=~ \bx + \frac{d\varphi^{-1}(\bu)}{d\bu}|_{\bu=\bu^\prime} (\bu-\bu^*).
\end{equation}
If we denote $\Lambda(\bu^\prime) = \left( \frac{d\varphi^{-1}(\bu)}{d\bu}|_{\bu=\bu^\prime} \right)^\top \left( \frac{d\varphi^{-1}(\bu)}{d\bu}|_{\bu=\bu^\prime} \right)$, we then have that
\begin{eqnarray}
	\left|\left|\bx-\varphi^{-1}(\bu)\right|\right|_2^2 &=& (\bu-\bu^*)^\top \Lambda(\bu^\prime) (\bu-\bu^*) ~=~ \sum_{i,j} \Lambda(\bu^\prime)_{i,j} (\bu_i-\bu^*_i)^\top(\bu_j-\bu^*_j) \nonumber \\
	&\le& \sum_i \left(\sum_j \Lambda(\bu^\prime)_{i,j}\right)(\bu_i-\bu^*_i)^2 ~\le~ \left|\left| \Lambda(\bu^\prime) \right|\right|_1 \cdot ||\bu-\bu^*||_2^2.
\end{eqnarray}
And after defining
\begin{equation}
	D(\bu^*) ~=~ \max_{\bu\in B(\bu^*,1)} \left|\left| \Lambda(\bu) \right|\right|_1 ~\ge~ \max_{\bu\in B(\bu^*, \sqrt{\gamma_t^*})} \left|\left| \Lambda(\bu) \right|\right|,
\end{equation}
it also follows that
\begin{equation}
	\left|\left|\bx-\varphi^{-1}(\bu)\right|\right|_2^2 ~\le~ \left|\left| \Lambda(\bu^\prime) \right|\right|_1 \cdot ||\bu-\bu^*||_2^2 ~\le~ D(\bu^*)\gamma^*_t~~~~\forall\bu\in B(\bu^*,\sqrt{\gamma^*_t}).
\end{equation}
Plugging this inequality into (\ref{eqn:p_x_bound}) gives
\begin{eqnarray}
	p_{\theta^*_t}(\bx) &\ge& \int_{B(\bu^*,\sqrt{\gamma^*_t})} (2\pi\gamma^*_t)^{-d/2}\exp \left\{ -\frac{ D(\bu^*)\gamma^*_t }{ 2\gamma^*_t } \right\} p_{gt}^u(\bu) d\bu \nonumber \\
	&\ge& (2\pi\gamma^*_t)^{-d/2}\exp \left\{ -\frac{ D(\bu^*) }{ 2 } \right\} \left( \min_{\bu\in B(\bu^*,1)} p_{gt}^u(\bu) \right) \int_{B(\bu^*, \sqrt{\gamma^*_t})}d\bu \nonumber \\
	&=& (2\pi\gamma^*_t)^{-d/2}\exp \left\{ -\frac{ D(\bu^*) }{ 2 } \right\} \left( \min_{\bu\in B(\bu^*,1)} p_{gt}^u(\bu) \right) V\left(B(\bu^*,\sqrt{\gamma^*_t})\right),
\end{eqnarray}
where $V\left(B(\bu^*,\sqrt{\gamma^*_t})\right)$ is the volume of the $r$-dimensional ball $B(\bu^*,\sqrt{\gamma^*})$. The volume should be $a_r{\gamma^*}^{r/2}$ where $a_r$ is a constant related to the dimension $r$. So
\begin{equation}
	p_{\theta^*_t}(\bx) \ge (2\pi)^{-d/2} {\gamma^*}^{-(d-r)/2} a_r \exp \left\{ -\frac{ D(\bu^*) }{ 2 } \right\} \left( \min_{\bu\in B(\bu^*,1)} p_{gt}^u(\bu) \right).
\end{equation}
Since $\varphi(\cdot)$ defines a diffeomorphism, $D(\bu^*)<\infty$. Moreover, $\left( \min_{\bu\in B(\bu^*,1)} p_{gt}^u(\bu) \right)>0$ because $p_{gt}^u(\bu)$ is nonzero and continuous everywhere. We may then conclude that
\begin{equation}
	\lim_{t\to\infty} -\log p_{\theta^*}(\bx) = -\infty.
\end{equation}
for $\bx \in \calX$.  This then implies that the stated average across $\calX$ with respect to $\mu_{gt}$ will also be $-\infty$.

\subsection{Finding a Sequence of Encoders such that $\mathbb{KL}\left[q_{\phi^*_t}(\bz|\bx)||p_{\theta^*_t}(\bz|\bx)\right]$ Converges to $0$}
Similar to (\ref{eqn:encoder_mean}) and (\ref{eqn:encoder_var}), let the encoder be
\begin{eqnarray}
    f_{\mu_z}(\bx;\phi^*_t) &=& G^{-1}\circ F\circ \varphi (\bx) = f_{\mu_x}^{-1}(\bx;\theta^*_t), \label{eqn:encoder_mean2}\\
    f_{S_z}(\bx;\phi^*_t) &=& \sqrt{\gamma^*_t \left( f^\prime_{\mu_x}\left(f_{\mu_z}(\bx;\phi^*_t);\theta^*_t\right)^\top f^\prime_{\mu_x}\left(f_{\mu_z}(\bx;\phi^*_t);\theta^*_t\right) \right)^{-1}}. \label{eqn:encoder_var2}
\end{eqnarray}
Following the proofs in Section~\ref{sec:proof_sufficient_full_kl}, we can prove the KL divergence between $q_{\phi^*_t}(\bz|\bx)$ and $p_{\theta^*_t}(\bz|\bx)$ converges to $0$.

\subsection{The Relationship between $\lim_{t\to\infty}p_{\theta^*_t}(\bx)$ and $\mu_{gt}(\bx)$}
We then prove our construction from (\ref{eqn:decoder_mean2}) and (\ref{eqn:decoder_var2}) satisfies (\ref{eqn:match_distributiin}). Unlike the case $d=r$ where we can compare $p_{\theta^*_t}(\bx)$ and $p_{gt}(\bx)$ directly, here $p_{\theta^*_t}(\bx)$ is a density defined everywhere in $\mathbb{R}^d$ while $\mu_{gt}$ is a probability measure defined only on the $r$-dimensional manifold $\bchi$. Consequently, to assess $p_{\theta_t^*}(\bx)$ relative to $\mu_{gt}$, we evaluate the respective probability mass assigned to any measurable subset of $\mathbb{R}^d$ denoted as $A$. For $p_{\theta_t^*}(\bx)$, we integrate the density over $A$ while for $\mu_{gt}$ we compute the measure of the intersection of $A$ with $\bchi$, i.e., $\mu_{gt}$ confines all mass to the manifold.

We begin with the probability distribution given by $p_{\theta_t^*}(\bx)$:
\begin{align}
	p_{\theta_t^*}(\bx) &= \int_{\mathbb{R}^r} p_{\theta_t^*}(\bx|\bz) p(\bz) d\bz = \int_{\mathbb{R}^r} \mathcal{N}\left( \bx | \varphi^{-1} \circ F^{-1} \circ G(z), \gamma_t^*\bI \right) dG(\bz) \nonumber \\
	&= \int_{[0,1]^r} \mathcal{N}\left( \bx | \varphi^{-1} \circ F^{-1}(\bxi), \gamma_t^*\bI \right) d\bxi \nonumber \\
	&= \int_{\mathbb{R}^r} \mathcal{N}\left( \bx | \varphi^{-1}(\bu), \gamma_t^*\bI \right) p_{gt}^u(\bu) d\bu \nonumber \\
	&= \int_{\bx^\prime\in\bchi} \mathcal{N}\left( \bx | \bx^\prime, \gamma_t^* \right) \mu_{gt}(d\bx^\prime).
\end{align}
Consider a measurable set $A\in\mathbb{R}^d$,
\begin{align}
	\lim_{t\to\infty} \int_{\bx\in A} p_{\theta_t^*}(\bx) d\bx &= \lim_{t\to\infty}\int_{\bx\in A} \left[ \int_{\bx^\prime\in\bchi} \mathcal{N}\left( \bx | \bx^\prime, \gamma_t^* \right) \mu_{gt}(d\bx^\prime) \right] d\bx \nonumber \\
	&= \lim_{t\to\infty} \int_{\bx^\prime\in\bchi} \left[ \int_{\bx\in A} \mathcal{N}\left( \bx | \bx^\prime, \gamma_t^* \right) d\bx \right] \mu_{gt}(d\bx^\prime) \nonumber \\
	&= \int_{\bx^\prime\in\bchi} \lim_{t\to\infty} \left[ \int_{\bx\in A} \mathcal{N}\left( \bx | \bx^\prime, \gamma_t^* \right) d\bx \right] \mu_{gt}(d\bx^\prime). \label{eqn:p_theta_x}
\end{align}
The second equation that interchanges the order of the integrations admitted by Fubini's theorem. The third equation that interchanges the order of the integration and the limit is justified by the bounded convergence theorem.  We now note that the term inside the first integration, $\mathcal{N}(\bx| \bx^\prime, \gamma_t^*\bI)$,  converges to a Dirac-delta function as $\gamma_t^*\to0$. So the integration over $A$ depends on whether $\bx^\prime$ is inside $A$ or not, i.e.,
\begin{equation}
	\lim_{t\to\infty} \left[ \int_{\bx\in\bA} \mathcal{N}\left( \bx | \bx^\prime, \gamma_t^*\bI \right) d\bx \right] =
	\begin{cases}
		1 & \text{if $\bx^\prime\in\bA-\partial\bA$}, \\
		0 & \text{if $\bx^\prime\in\bA^c-\partial\bA$}.
	\end{cases}
\end{equation}

We separate the manifold $\bchi$ into three parts: $\bchi\cap\left(\bA-\partial\bA\right)$, $\bchi\cap\left(\bA^c-\partial\bA\right)$ and $\bchi\cap\partial\bA$. Then (\ref{eqn:p_theta_x}) can be separated into three parts accordingly. The first two parts can be derived as
\begin{equation}
    \int_{\bchi\cap\left(\bA-\partial\bA\right)} \lim_{t\to\infty} \left[\int_{\bA} \mathcal{N}(\bx|\bx^\prime,\gamma_t^*\bI)d\bx \right] \mu_{gt}(d\bx^\prime) = \int_{\bchi\cap\left(\bA-\partial\bA\right)} 1 ~\mu_{gt}(d\bx^\prime) = \mu_{gt}\left(\bchi\cap\left(\bA-\partial\bA\right)\right),
\end{equation}
\begin{equation}
	\int_{\bchi\cap\left(\bA^c-\partial\bA\right)} \lim_{t\to\infty} \left[\int_{\bA} \mathcal{N}(\bx|\bx^\prime,\gamma_t^*\bI)d\bx \right] \mu_{gt}(d\bx^\prime) = \int_{\bchi\cap\left(\bA-\partial\bA\right)} 0 ~\mu_{gt}(d\bx^\prime) = 0.
\end{equation}
For the third part, given the assumption that $\mu_{gt}(\partial\bA)=0$, we have
\begin{equation}
    0 \le \int_{\bchi\cap\partial\bA} \lim_{t\to\infty} \left[\int_{\bA} \mathcal{N}(\bx|\bx^\prime,\gamma_t^*\bI)d\bx \right] \mu_{gt}(d\bx^\prime) \le \int_{\bchi\cap\partial\bA} 1 ~\mu_{gt}(d\bx^\prime) = \mu_{gt}\left(\bchi\cap\partial\bA\right) = 0.
\end{equation}
Therefore we have
\begin{equation}
	\int_{\bchi\cap\partial\bA} \lim_{t\to\infty} \left[\int_{\bA} \mathcal{N}(\bx|\bx^\prime,\gamma_t^*\bI)d\bx \right] \mu_{gt}(d\bx^\prime) = 0
\end{equation}
and thus
\begin{eqnarray}
    \lim_{t\to\infty} \int_{\bA} p_{gt}(\bx;\gamma_t^*\bI) d\bx &=& \int_{\bx^\prime\in\bchi} \lim_{t\to\infty} \left[ \int_{\bx\in A} \mathcal{N}\left( \bx | \bx^\prime, \gamma_t^* \right) d\bx \right] \mu_{gt}(d\bx^\prime) \nonumber \\
    &=& \mu_{gt}\left(\bchi\cap\left(\bA-\partial\bA\right)\right) + 0 + 0 \nonumber \\
    &=& \mu_{gt}(\bchi\cap\bA),
\end{eqnarray}
leading to (\ref{eqn:match_distributiin}).

Note that this result involves a subtle requirement involving the boundary $\partial A$. This condition is only included to handle a minor, practically-inconsequential technicality.  In brief, as a density $p_{\theta_t^*}(\bx)$ will apply zero mass exactly on any low-dimensional manifold, although it can apply all of its mass to any region in the neighborhood of $\bchi$.  But suppose we choose some $A$ is a subset of $\bchi$, i.e, it is exclusively confined to the ground-truth manifold.  Then the probability mass within $A$ assigned by $\mu_{gt}$ will be nonzero while that given by $p_{\theta_t^*}(\bx)$ can still be zero.  Of course this does not mean that $p_{\theta_t^*}(\bx)$ and $\mu_{gt}$ do not match each other in any practical sense.  This is because if we expand this specialized $A$ by an arbitrary small $d$-dimensional volume, then $p_{\theta_t^*}(\bx)$ and $\mu_{gt}$ will now supply essentially the same probability mass on this infinitesimally expanded set (which is arbitrary close to $A$).


\section{Proof of Theorem~\ref{thm:decoder_variance}}\label{sec:proof_decoder_variance}
From the main text, $\{\theta^*_\gamma,\phi^*_\gamma\}$ is the optimal solution with a fixed $\gamma$. The true posterior and the approximate posterior are
\begin{eqnarray}
    p_{\theta^*_\gamma}(\bz|\bx) &=& \frac{p(\bz)p_{\theta^*_\gamma}(\bx|\bz)}{p_{\theta^*_\gamma}(\bx)}, \\
    q_{\phi^*_\gamma}(\bz|\bx) &=& \mathcal{N}\left(\bz | \bmu_z(\bx;\phi^*_\gamma), \bSigma_z\right(\bx;\phi^*_\gamma)) \label{eqn:encoder}.
\end{eqnarray}

\subsection{Case 1: $r=d$}
We first argue that the KL divergence between $p_{\theta^*_\gamma}(\bz|\bx)$ and $q_{\phi^*_\gamma}(\bz|\bx)$ is always strictly greater than zero. This can be proved by contradiction. Suppose the KL divergence exactly equals zero.  Then $p_{\theta^*_\gamma}(\bz|\bx)$ must also be a Gaussian distribution, meaning that the logarithm of $p_{\theta^*_\gamma}(\bz|\bx)$ is a quadratic form in $\bz$.  In particular, we have
\begin{eqnarray}
    \log p_{\theta^*_\gamma}(\bz|\bx) &=& \log \mathcal{N}(\bz|0,\bI) + \log \mathcal{N}(\bx|f_{\mu_x}(\bz),\gamma\bI) - \log p(\bx) \nonumber \\
    &=& -\frac{1}{2}||\bz||_2^2 - \frac{1}{2\gamma}||\bx-f_{\mu_x}(\bz)||_2^2 + \text{constant},
\end{eqnarray}
where we have absorbed all the terms not related to $\bz$ into a constant, and it must be that
\begin{equation}
    f_{\mu_x}(\bz) = \bW\bz + \bb,
\end{equation}
for some matrix $\bW$ and vector $\bb$.  Then we have
\begin{eqnarray}
    p_{\theta^*_\gamma}(\bx) &=& \int_{\mathbb{R}^\kappa} p_\theta(\bx|\bz) p(\bz) d\bz \nonumber \\
    &=& \int_{\mathbb{R}^\kappa} \mathcal{N}(\bx|\bW\bz+\bb,\gamma\bI)\mathcal{N}(\bz|0,\bI) d\bz.
\end{eqnarray}
This is a Gaussian distribution in $\mathbb{R}^d$ which contradicts our assumption that $p_{gt}(\bx)$ is not Gaussian. So the KL divergence between $p_{\theta^*_\gamma}(\bz|\bx)$ and $q_{\phi^*_\gamma}(\bz|\bx)$ is always greater than $0$. As a result, $\mathcal{L}(\theta^*_\gamma,\phi^*_\gamma)$ cannot reach the theoretical optimal solution, i.e., $\int_{\bchi} -p_{gt}(\bx) \log p_{gt}(\bx) d\bx$. Denote the gap between $\mathcal{L}(\theta^*_\gamma,\phi^*_\gamma)$ and $\int_{\bchi} -p_{gt}(\bx) \log p_{gt}(\bx) d\bx$ as $\epsilon$. According to the proof in Section~\ref{sec:proof_optima_r_eq_d}, there exists a $t_0$ such that for any $t>t_0$, the gap between the proposed solution in Section~\ref{sec:proof_optima_r_eq_d} and the theoretical optimal solution is smaller than $\epsilon$. Pick some $t>t_0$ such that $1/t<\gamma$ and let $\gamma^\prime=1/t$. Then
\begin{equation}
    \mathcal{L}(\theta_{\gamma^\prime}^*,\phi_{\gamma^\prime}^*) \le \mathcal{L}(\theta_t^*,\phi_t^*) < \mathcal{L}(\theta_\gamma^*,\phi_\gamma^*).
\end{equation}
The first inequality comes from the fact that $\{ \theta_{\gamma^\prime}^*, \phi_{\gamma^\prime}^* \}$ is the optimal solution when $\gamma$ is fixed at $\gamma^\prime$ while $\{ \theta_t^*, \phi_t^* \}$ is just one solution with $\gamma=1/t=\gamma^\prime$. The second inequality holds because we chose $\{ \theta_t^*, \phi_t^* \}$ to be a better solution than $\{ \theta_\gamma^*,\phi_\gamma^* \}$.

\subsection{Case 2: $r<d$}
In this case, $\mathbb{KL}\left[ q_{\phi_\gamma^*}(\bz|\bx) || p_{\theta_\gamma^*}(\bz|\bx) \right]$ does not need to be zero because it is possible that $- \log p_{\theta_\gamma^*}(\bx)$ diverges to negative infinity and absorbs the positive cost caused by the KL divergence. Consider the objective function expression from (\ref{eq:objective_general2}).  In can be bounded by
\begin{align}
    \mathcal{L}(\theta_\gamma^*,\phi_\gamma^*) &= \int_{\bchi} \left\{ - \mathbb{E}_{q_{\tiny \phi_\gamma^*}\left(\bz|\bx \right)} \left[\log p_{\tiny \theta_\gamma^*} \left(\bx | \bz  \right)  \right] + \mathbb{KL}\left[ q_{\phi_\gamma^*}(\bz|\bx) || p(\bz) \right]\right\} \mu_{gt}(d\bx) \nonumber \\
    & \ge \int_{\bchi} \left\{ - \mathbb{E}_{q_{\tiny \phi_\gamma^*}\left(\bz|\bx \right)} \left[ \frac{||\bx-f_{\mu_x}(\bz)||_2^2}{2\gamma} + \frac{d}{2}\log(2\pi\gamma)  \right] \right\} \mu_{gt}(d\bx) \nonumber \\
    & \ge \frac{d}{2}\log\gamma > -\infty.
\end{align}
The first inequality holds discarding the KL term, which is non-negative. The second inequality holds because a quadratic term is removed. Furthermore, according to the proof in Section~\ref{sec:proof_optima_r_less_d}, there exists a $t_0$ such that for any $t>t_0$,
\begin{equation}
    \mathcal{L}(\theta_t^*, \phi_t^*) < \frac{d}{2}\log\gamma.
\end{equation}
Again, we select a $t>t_0$ such that $1/t<\gamma$ and let $\gamma^\prime=1/t$. Then
\begin{equation}
    \mathcal{L}(\theta_{\gamma^\prime}^*,\phi_{\gamma^\prime}^*) \le \mathcal{L}(\theta_t^*,\phi_t^*) < \mathcal{L}(\theta_\gamma^*,\phi_\gamma^*).
\end{equation}

\section{Proof of Theorem~\ref{thm:decoder_mean}}\label{sec:proof_decoder_mean}

Recall that
\begin{eqnarray}
    q_{\phi_\gamma^*}(\bz|\bx) &=& \mathcal{N} \left( \bz | f_{\mu_z}(\bx;\phi_\gamma^*), f_{S_z}(\bx;\phi_\gamma^*) f_{S_z}(\bx;\phi_\gamma^*)^\top \right), \label{eqn:approx_posterior}\\
    p_{\theta_\gamma^*}(\bx|\bz) &=& \mathcal{N} \left( \bx | f_{\mu_x}(\bz;\theta_\gamma^*), \gamma\bI \right). \label{eqn:true_posterior}
\end{eqnarray}
Plugging these expressions into (\ref{eq:objective_general2}) we obtain
\begin{align}
    \mathcal{L}(\theta_\gamma^*, \phi_\gamma^*) &= \mathbb{E}_{\bz\sim q_{\phi_\gamma^*}(\bz|\bx)} \left[ \frac{1}{2\gamma} ||f_{\mu_x}(\bz) - \bx||_2^2 + \frac{d}{2}\log(2\pi\gamma) \right] + \mathbb{KL} \left[ q_{\phi_\gamma^*}(\bz|\bx) || p(\bz) \right] \label{eqn:objective_expand} \\
    &\ge \frac{1}{2\gamma} \mathbb{E}_{\epsilon\sim\mathcal{N}(0,\bI)} \left[ \left|\left| f_{\mu_x}\left[ f_{\mu_z}(\bx) + f_{S_z}(\bx)\epsilon \right] - \bx \right|\right|_2^2 \right] + \frac{d}{2}\log(2\pi\gamma),
\end{align}
where we have omitted explicit inclusion of the parameters $\phi_\gamma^*$ and $\theta_\gamma^*$ in the functions $f_{\mu_z}(\cdot), f_{S_z}(\cdot)$ and $f_{\mu_x}(\cdot)$ to avoid undue clutter. Now suppose
\begin{equation}
    \lim_{\gamma\to0} \mathbb{E}_{\epsilon\sim\mathcal{N}(0,\bI)} \left[ \left|\left| f_{\mu_x}\left[ f_{\mu_z}(\bx) + f_{S_z}(\bx)\epsilon \right] - \bx \right|\right|_2^2 \right] = \Delta \neq 0.
\end{equation}
It then follows that
\begin{equation}
    \lim_{\gamma\to0} \mathcal{L}(\theta_\gamma^*, \phi_\gamma^*) \ge \lim_{\gamma\to0} \frac{\Delta}{2\gamma} + \frac{d}{2}\log(2\pi\gamma)=+\infty,
\end{equation}
which contradicts the fact that $\mathcal{L}(\theta_\gamma^*, \phi_\gamma^*)$ converges to $-\infty$. So we must have that
\begin{equation}
    \lim_{\gamma\to0} \mathbb{E}_{\epsilon\sim\mathcal{N}(0,\bI)} \left[ \left|\left| f_{\mu_x}\left[ f_{\mu_z}(\bx) + f_{S_z}(\bx)\epsilon \right] - \bx \right|\right|_2^2 \right] = 0.
\end{equation}
Because the term inside the expectation, i.e., $\left|\left| f_{\mu_x}\left[ f_{\mu_z}(\bx) + f_{S_z}(\bx)\epsilon \right] - \bx \right|\right|_2^2$, is always non-negative, we can conclude that
\begin{equation}
    \lim_{\gamma\to0} f_{\mu_x}\left[ f_{\mu_z}(\bx) + f_{S_z}(\bx)\epsilon \right] = \bx.
\end{equation}
And if we let $\epsilon=0$, this equation then becomes
\begin{equation}
    \lim_{\gamma\to0} f_{\mu_x}\left[ f_{\mu_z}(\bx) \right] = \bx.
\end{equation}


\section{Further Analysis of the VAE Cost as $\gamma$ becomes small}\label{sec:analysis_decoder_mean}
In the main paper, we mentioned that the squared eigenvalues of $f_{S_z}(\bx;\phi_\gamma^*)$ will become arbitrary small at a rate proportional to $\gamma$. To justify this, we borrow the simplified notation from the proof of Theorem~\ref{thm:decoder_mean} and expand $f_{\mu_x}(\bz)$ at $\bz=f_{\mu_z}(\bx)$ using a Taylor series.  Omitting the high order terms (in the present narrow context around the neighborhood of VAE global optima these will be small), this gives
\begin{equation}
    f_{\mu_x}(\bz) \approx f_{\mu_x}\left[f_{\mu_z}(\bx)\right] + f_{\mu_x}^\prime\left[f_{\mu_z}(\bx)\right](\bz-f_{\mu_z}) \approx \bx + f_{\mu_x}^\prime\left[f_{\mu_z}(\bx)\right](\bz-f_{\mu_z}).
\end{equation}
Plug this expression and (\ref{eqn:approx_posterior}) into (\ref{eqn:objective_expand}), we obtain
\begin{align}
    \mathcal{L}(\theta_\gamma^*, \phi_\gamma^*) & \approx \mathbb{E}_{\bz\sim q_{\phi_\gamma^*}(\bz|\bx)} \left[ \frac{1}{2\gamma} || f_{\mu_x}^\prime\left[f_{\mu_z}(\bx)\right](\bz-f_{\mu_z}(\bx)) ||_2^2 + \frac{d}{2}\log(2\pi\gamma) \right] \nonumber \\
    &~~~~ + \frac{1}{2} \left\{ ||f_{\mu_z}(\bx)||_2^2 + \mbox{tr}\left( f_{S_z}(\bx) f_{S_z}(\bx)^\top \right) - \log |f_{S_z}(\bx) f_{S_z}(\bx)^\top| - \kappa \right\} \nonumber \\
    &= \frac{1}{2\gamma} \mbox{tr}\left( \mathbb{E}_{\bz\sim q_{\phi_\gamma^*}(\bz|\bx)} \left[ (\bz-f_{\mu_z}(\bx))^\top (\bz-f_{\mu_z}(\bx)) \right] f_{\mu_x}^\prime\left[f_{\mu_z}(\bx)\right]^\top f_{\mu_x}^\prime\left[f_{\mu_z}(\bx)\right] \right) \nonumber \\
    &~~~~ + \frac{d}{2}\log(2\pi\gamma) + \frac{1}{2} \left\{ ||f_{\mu_z}(\bx)||_2^2 + tr\left( f_{S_z}(\bx) f_{S_z}(\bx)^\top \right) - \log |f_{S_z}(\bx) f_{S_z}(\bx)^\top| - \kappa \right\} \nonumber \\
    &= tr\left( f_{S_z}(\bx) f_{S_z}(\bx)^\top \left[ \frac{1}{2}\bI + \frac{1}{2\gamma} f_{\mu_x}^\prime\left[f_{\mu_z}(\bx)\right]^\top f_{\mu_x}^\prime\left[f_{\mu_z}(\bx)\right] \right] \right) \nonumber \\
    &~~~~ + \frac{d}{2}\log(2\pi\gamma) + \frac{1}{2} \left\{ ||f_{\mu_z}(\bx)||_2^2 - \log |f_{S_z}(\bx) f_{S_z}(\bx)^\top| - \kappa \right\}.
\end{align}
From these manipulations we may conclude that the optimal value of $f_{S_z}(\bx) f_{S_z}(\bx)^\top$ must satisfy
\begin{equation}
    \left[ \frac{1}{2}\bI + \frac{1}{2\gamma} f_{\mu_x}^\prime\left[f_{\mu_z}(\bx)\right]^\top f_{\mu_x}^\prime\left[f_{\mu_z}(\bx)\right] \right] - \frac{1}{2} \left( f_{S_z}(\bx) f_{S_z}(\bx)^\top \right)^{-1} = 0.
\end{equation}
So
\begin{equation}
    f_{S_z}(\bx) f_{S_z}(\bx)^\top = \left[ \bI + \frac{1}{\gamma} f_{\mu_x}^\prime\left[f_{\mu_z}(\bx)\right]^\top f_{\mu_x}^\prime\left[f_{\mu_z}(\bx)\right] \right]^{-1}.
\end{equation}
Note that $f_{\mu_x}^\prime\left[f_{\mu_z}(\bx)\right]$ is the tangent space of the manifold $\bchi$ at $f_{\mu_x}\left[f_{\mu_z}(\bx)\right]$, so the rank must be $r$. $f_{\mu_x}^\prime\left[f_{\mu_z}(\bx)\right]^\top f_{\mu_x}^\prime\left[f_{\mu_z}(\bx)\right]$ can be decomposed as $\bU^\top\bS\bU$, where $\bU$ is a $\kappa$-dimensional orthogonal matrix and $\bS$ is a $\kappa$-dimensional diagonal matrix with $r$ nonzero elements. Denote $\text{diag}[\bS]=[S_1, S_2, ..., S_r, 0, ..., 0]$. Then
\begin{equation}
    f_{S_z}(\bx) f_{S_z}(\bx)^\top = \left[ \bU^\top \text{diag}\left[ 1+\frac{S_1}{\gamma}, ..., 1+\frac{S_r}{\gamma}, 1, ..., 1 \right] \bU \right]^{-1}.
\end{equation}

\textbf{Case 1: $r=\kappa$. } In this case, $\bS$ has no nonzero diagonal elements, and therefore
\begin{equation}
    f_{S_z}(\bx) f_{S_z}(\bx)^\top = \left[ \bU^\top \text{diag}\left[ \frac{1}{1+\frac{S_1}{\gamma}}, ..., \frac{1}{1+\frac{S_r}{\gamma}} \right] \bU \right].
\end{equation}
As $\gamma\to0$, the eigenvalues of $f_{S_z}(\bx) f_{S_z}(\bx)^\top$, which are given by $\frac{1}{1+\frac{S_i}{\gamma}}$, converge to $0$ at a rate of $O(\gamma)$.

\textbf{Case 2: $r<\kappa$. } In this case, the first $r$ eigenvalues also converge to $0$ at a rate of $O(\gamma)$, but the remaining $\kappa-r$ eigenvalues will be $1$, meaning the redundant dimensions are simply filled with noise matching the prior $p(\bz)$ as desired.

\vskip 0.2in
\bibliography{refs}

\begin{thebibliography}{44}
\providecommand{\natexlab}[1]{#1}
\providecommand{\url}[1]{\texttt{#1}}
\expandafter\ifx\csname urlstyle\endcsname\relax
  \providecommand{\doi}[1]{doi: #1}\else
  \providecommand{\doi}{doi: \begingroup \urlstyle{rm}\Url}\fi

\bibitem[Arjovsky et~al.(2017)Arjovsky, Chintala, and
  Bottou]{arjovsky2017wasserstein}
Martin Arjovsky, Soumith Chintala, and L{\'e}on Bottou.
\newblock Wasserstein generative adversarial networks.
\newblock In \emph{International Conference on Machine Learning}, pages
  214--223, 2017.

\bibitem[Bengio et~al.(2013)Bengio, Courville, and
  Vincent]{bengio2013representation}
Yoshua Bengio, Aaron Courville, and Pascal Vincent.
\newblock Representation learning: A review and new perspectives.
\newblock \emph{IEEE Transactions on Pattern Analysis and Machine
  Intelligence}, 35\penalty0 (8):\penalty0 1798--1828, 2013.

\bibitem[Berthelot et~al.(2017)Berthelot, Schumm, and Metz]{berthelot2017began}
David Berthelot, Thomas Schumm, and Luke Metz.
\newblock {BEGAN}: {B}oundary equilibrium generative adversarial networks.
\newblock \emph{arXiv:1703.10717}, 2017.

\bibitem[Bi{\'n}kowski et~al.(2018)Bi{\'n}kowski, Sutherland, Arbel, and
  Gretton]{binkowski2018demystifying}
Miko{\l}aj Bi{\'n}kowski, Dougal~J Sutherland, Michael Arbel, and Arthur
  Gretton.
\newblock Demystifying mmd gans.
\newblock \emph{arXiv:1801.01401}, 2018.

\bibitem[Bousquet et~al.(2017)Bousquet, Gelly, Tolstikhin, Simon-Gabriel, and
  Sch{\"o}lkopf]{bousquetetal2017}
Olivier Bousquet, Sylvain Gelly, Ilya Tolstikhin, Carl~Johann Simon-Gabriel,
  and Bernhard Sch{\"o}lkopf.
\newblock From optimal transport to generative modeling: the {VEGAN} cookbook.
\newblock \emph{arXiv:1611.02731}, 2017.

\bibitem[Brock et~al.(2016)Brock, Lim, Ritchie, and Weston]{brock2016neural}
Andrew Brock, Theodore Lim, James~M Ritchie, and Nick Weston.
\newblock Neural photo editing with introspective adversarial networks.
\newblock \emph{arXiv:1609.07093}, 2016.

\bibitem[Burda et~al.(2015)Burda, Grosse, and
  Salakhutdinov]{burda2015importance}
Yuri Burda, Roger Grosse, and Ruslan Salakhutdinov.
\newblock Importance weighted autoencoders.
\newblock \emph{arXiv:1509.00519}, 2015.

\bibitem[Chen et~al.(2018)Chen, Li, Grosse, and Duvenaud]{chen2018isolating}
Tian~Qi Chen, Xuechen Li, Roger Grosse, and David Duvenaud.
\newblock Isolating sources of disentanglement in variational autoencoders.
\newblock In \emph{Advances in Neural Information Processing Systems}, 2018.

\bibitem[Chen et~al.(2016{\natexlab{a}})Chen, Duan, Houthooft, Schulman,
  Sutskever, and Abbeel]{chen2016infogan}
Xi~Chen, Yan Duan, Rein Houthooft, John Schulman, Ilya Sutskever, and Pieter
  Abbeel.
\newblock Info{GAN}: {I}nterpretable representation learning by information
  maximizing generative adversarial nets.
\newblock In \emph{Advances in Neural Information Processing Systems}, pages
  2172--2180, 2016{\natexlab{a}}.

\bibitem[Chen et~al.(2016{\natexlab{b}})Chen, Kingma, Salimans, Duan, Dhariwal,
  Schulman, Sutskever, and Abbeel]{chen2016variational}
Xi~Chen, Diederik Kingma, Tim Salimans, Yan Duan, Prafulla Dhariwal, John
  Schulman, Ilya Sutskever, and Pieter Abbeel.
\newblock Variational lossy autoencoder.
\newblock \emph{arXiv:1611.02731}, 2016{\natexlab{b}}.

\bibitem[Dai and Wipf(2019)]{Dai2019iclr}
Bin Dai and David Wipf.
\newblock Diagnosing and enhancing vae models.
\newblock In \emph{International Conference on Learning Representations}, 2019.

\bibitem[Dai et~al.(2018)Dai, Wang, Aston, Hua, and Wipf]{dai2018jmlr}
Bin Dai, Yu~Wang, John Aston, Gang Hua, and David Wipf.
\newblock Connections with robust {PCA} and the role of emergent sparsity in
  variational autoencoder models.
\newblock \emph{Journal of Machine Learning Research}, 2018.

\bibitem[Davidson et~al.(2018)Davidson, Falorsi, De~Cao, Kipf, and
  Tomczak]{davidson2018hyperspherical}
Tim Davidson, Luca Falorsi, Nicola De~Cao, Thomas Kipf, and Jakub Tomczak.
\newblock Hyperspherical variational auto-encoders.
\newblock \emph{arXiv:1804.00891}, 2018.

\bibitem[Doersch(2016)]{doersch2016tutorial}
Carl Doersch.
\newblock Tutorial on variational autoencoders.
\newblock \emph{arXiv:1606.05908}, 2016.

\bibitem[Dosovitskiy and Brox(2016)]{dosovitskiy2016generating}
Alexey Dosovitskiy and Thomas Brox.
\newblock Generating images with perceptual similarity metrics based on deep
  networks.
\newblock In \emph{Advances in Neural Information Processing Systems}, pages
  658--666, 2016.

\bibitem[Fedus et~al.(2017)Fedus, Rosca, Lakshminarayanan, Dai, Mohamed, and
  Goodfellow]{fedus2017many}
William Fedus, Mihaela Rosca, Balaji Lakshminarayanan, Andrew~M Dai, Shakir
  Mohamed, and Ian Goodfellow.
\newblock Many paths to equilibrium: {GAN}s do not need to decrease a
  divergence at every step.
\newblock \emph{arXiv:1710.08446}, 2017.

\bibitem[Goodfellow et~al.(2014)Goodfellow, Pouget-Abadie, Mirza, Xu,
  Warde-Farley, Ozair, Courville, and Bengio]{goodfellow2014generative}
Ian Goodfellow, Jean Pouget-Abadie, Mehdi Mirza, Bing Xu, David Warde-Farley,
  Sherjil Ozair, Aaron Courville, and Yoshua Bengio.
\newblock Generative adversarial nets.
\newblock In \emph{Advances in Neural Information Processing Systems}, pages
  2672--2680, 2014.

\bibitem[Gretton et~al.(2007)Gretton, Borgwardt, Rasch, Sch{\"o}lkopf, and
  Smola]{gretton2007kernel}
Arthur Gretton, Karsten Borgwardt, Malte Rasch, Bernhard Sch{\"o}lkopf, and
  Alex Smola.
\newblock A kernel method for the two-sample-problem.
\newblock In \emph{Advances in Neural Information Processing Systems}, pages
  513--520, 2007.

\bibitem[Gulrajani et~al.(2017)Gulrajani, Ahmed, Arjovsky, Dumoulin, and
  Courville]{gulrajani2017improved}
Ishaan Gulrajani, Faruk Ahmed, Martin Arjovsky, Vincent Dumoulin, and Aaron
  Courville.
\newblock Improved training of {W}asserstein {GAN}s.
\newblock In \emph{Advances in Neural Information Processing Systems}, pages
  5767--5777, 2017.

\bibitem[Heusel et~al.(2017)Heusel, Ramsauer, Unterthiner, Nessler, and
  Hochreiter]{heusel2017gans}
Martin Heusel, Hubert Ramsauer, Thomas Unterthiner, Bernhard Nessler, and Sepp
  Hochreiter.
\newblock {GAN}s trained by a two time-scale update rule converge to a local
  {N}ash equilibrium.
\newblock In \emph{Advances in Neural Information Processing Systems}, pages
  6626--6637, 2017.

\bibitem[Higgins et~al.(2017)Higgins, Matthey, Pal, Burgess, Glorot, Botvinick,
  Mohamed, , and Lerchner]{higgins2017}
Irina Higgins, Loic Matthey, Arka Pal, Christopher Burgess, Xavier Glorot,
  Matthew Botvinick, Shakir Mohamed, , and Alexander Lerchner.
\newblock $\beta$-vae: Learning basic visual concepts with a constrained
  variational framework.
\newblock In \emph{International Conference on Learning Representations}, 2017.

\bibitem[Hyv{\"a}rinen and Oja(2000)]{hyvarinen2000independent}
Aapo Hyv{\"a}rinen and Erkki Oja.
\newblock Independent component analysis: algorithms and applications.
\newblock \emph{Neural networks}, 13\penalty0 (4-5):\penalty0 411--430, 2000.

\bibitem[Kingma and Welling(2014)]{Kingma2014}
Diederik Kingma and Max Welling.
\newblock Auto-encoding variational {B}ayes.
\newblock In \emph{International Conference on Learning Representations}, 2014.

\bibitem[Kingma et~al.(2016)Kingma, Salimans, Jozefowicz, Chen, Sutskever, and
  Welling]{kingma2016improved}
Diederik Kingma, Tim Salimans, Rafal Jozefowicz, Xi~Chen, Ilya Sutskever, and
  Max Welling.
\newblock Improved variational inference with inverse autoregressive flow.
\newblock In \emph{Advances in Neural Information Processing Systems}, pages
  4743--4751, 2016.

\bibitem[Kodali et~al.(2017)Kodali, Abernethy, Hays, and
  Kira]{kodali2017convergence}
Naveen Kodali, Jacob Abernethy, James Hays, and Zsolt Kira.
\newblock On convergence and stability of {GAN}s.
\newblock \emph{arXiv:1705.07215}, 2017.

\bibitem[Krizhevsky and Hinton(2009)]{krizhevsky2009learning}
Alex Krizhevsky and Geoffrey Hinton.
\newblock Learning multiple layers of features from tiny images.
\newblock Technical report, Citeseer, 2009.

\bibitem[Larsen et~al.(2015)Larsen, S{\o}nderby, Larochelle, and
  Winther]{larsen2015autoencoding}
Anders Boesen~Lindbo Larsen, S{\o}ren~Kaae S{\o}nderby, Hugo Larochelle, and
  Ole Winther.
\newblock Autoencoding beyond pixels using a learned similarity metric.
\newblock \emph{arXiv:1512.09300}, 2015.

\bibitem[LeCun et~al.(1998)LeCun, Bottou, Bengio, and
  Haffner]{lecun1998gradient}
Yann LeCun, L{\'e}on Bottou, Yoshua Bengio, and Patrick Haffner.
\newblock Gradient-based learning applied to document recognition.
\newblock \emph{Proceedings of the IEEE}, 86\penalty0 (11):\penalty0
  2278--2324, 1998.

\bibitem[Liu et~al.(2015)Liu, Luo, Wang, and Tang]{liu2015deep}
Ziwei Liu, Ping Luo, Xiaogang Wang, and Xiaoou Tang.
\newblock Deep learning face attributes in the wild.
\newblock In \emph{IEEE International Conference on Computer Vision}, pages
  3730--3738, 2015.

\bibitem[Locatello et~al.(2018)Locatello, Bauer, Lucic, Gelly, Sch{\"o}lkopf,
  and Bachem]{locatello2018challenging}
Francesco Locatello, Stefan Bauer, Mario Lucic, Sylvain Gelly, Bernhard
  Sch{\"o}lkopf, and Olivier Bachem.
\newblock Challenging common assumptions in the unsupervised learning of
  disentangled representations.
\newblock \emph{arXiv:1811.12359}, 2018.

\bibitem[Lucic et~al.(2018)Lucic, Kurach, Michalski, Gelly, and
  Bousquet]{lucic2018gans}
Mario Lucic, Karol Kurach, Marcin Michalski, Sylvain Gelly, and Olivier
  Bousquet.
\newblock Are {GAN}s created equal? {A} large-scale study.
\newblock \emph{Advances in Neural Information Processing Systems}, 2018.

\bibitem[Makhzani et~al.(2016)Makhzani, Shlens, Jaitly, Goodfellow, and
  Frey]{makhzani2016}
Alireza Makhzani, Jonathon Shlens, Navdeep Jaitly, Ian Goodfellow, and Brendan
  Frey.
\newblock Adversarial autoencoders.
\newblock \emph{arXiv:1511.05644}, 2016.

\bibitem[Mao et~al.(2017)Mao, Li, Xie, Lau, Wang, and Smolley]{mao2017least}
Xudong Mao, Qing Li, Haoran Xie, Raymond~YK Lau, Zhen Wang, and Stephen~Paul
  Smolley.
\newblock Least squares generative adversarial networks.
\newblock In \emph{IEEE International Conference on Computer Vision}, pages
  2813--2821, 2017.

\bibitem[Rezende and Mohamed(2015)]{rezende2015variational}
Danilo~Jimenez Rezende and Shakir Mohamed.
\newblock Variational inference with normalizing flows.
\newblock \emph{arXiv:1505.05770}, 2015.

\bibitem[Rezende et~al.(2014)Rezende, Mohamed, and Wierstra]{Rezende2014}
Danilo~Jimenez Rezende, Shakir Mohamed, and Daan Wierstra.
\newblock Stochastic backpropagation and approximate inference in deep
  generative models.
\newblock In \emph{International Conference on Machine Learning}, 2014.

\bibitem[Salimans et~al.(2016)Salimans, Goodfellow, Zaremba, Cheung, Radford,
  and Chen]{salimans2016improved}
Tim Salimans, Ian Goodfellow, Wojciech Zaremba, Vicki Cheung, Alec Radford, and
  Xi~Chen.
\newblock Improved techniques for training gans.
\newblock In \emph{Advances in Neural Information Processing Systems}, pages
  2234--2242, 2016.

\bibitem[Theis et~al.(2016)Theis, Oord, and Bethge]{Theis_ICLR2016}
Lucas Theis, A{\"a}ron van~den Oord, and Matthias Bethge.
\newblock A note on the evaluation of generative models.
\newblock In \emph{International Conference on Learning Representations}, pages
  1--10, 2016.

\bibitem[Tolstikhin et~al.(2018)Tolstikhin, Bousquet, Gelly, and
  Schoelkopf]{tolstikhin2018wasserstein}
Ilya Tolstikhin, Olivier Bousquet, Sylvain Gelly, and Bernhard Schoelkopf.
\newblock Wasserstein auto-encoders.
\newblock \emph{International Conference on Learning Representations}, 2018.

\bibitem[Tomczak and Welling(2018)]{tomczak2018vae}
Jakub Tomczak and Max Welling.
\newblock {VAE} with a {V}amp{P}rior.
\newblock In \emph{International Conference on Artificial Intelligence and
  Statistics}, pages 1214--1223, 2018.

\bibitem[van~den Berg et~al.(2018)van~den Berg, Hasenclever, Tomczak, and
  Welling]{van2018sylvester}
Rianne van~den Berg, Leonard Hasenclever, Jakub~M Tomczak, and Max Welling.
\newblock Sylvester normalizing flows for variational inference.
\newblock In \emph{Uncertainty in Artificial Intelligence}, 2018.

\bibitem[van~den Oord et~al.(2016)van~den Oord, Kalchbrenner, Espeholt,
  Vinyals, Graves, and Kavukcuoglu]{van2016conditional}
Aaron van~den Oord, Nal Kalchbrenner, Lasse Espeholt, Oriol Vinyals, Alex
  Graves, and Koray Kavukcuoglu.
\newblock Conditional image generation with {P}ixel{CNN} decoders.
\newblock In \emph{Advances in Neural Information Processing Systems}, pages
  4790--4798, 2016.

\bibitem[van~den Oord et~al.(2017)van~den Oord, Vinyals, and
  Kavukcuoglu]{van2017neural}
Aaron van~den Oord, Oriol Vinyals, and Koray Kavukcuoglu.
\newblock Neural discrete representation learning.
\newblock In \emph{Advances in Neural Information Processing Systems}, pages
  6306--6315, 2017.

\bibitem[Xiao et~al.(2017)Xiao, Rasul, and Vollgraf]{xiao2017/online}
Han Xiao, Kashif Rasul, and Roland Vollgraf.
\newblock Fashion-{MNIST}: {A} novel image dataset for benchmarking machine
  learning algorithms.
\newblock \emph{arXiv:1708.07747}, 2017.

\bibitem[Zhao et~al.(2018)Zhao, Kim, Zhang, Rush, and LeCun]{zhao18}
Junbo Zhao, Yoon Kim, Kelly Zhang, Alexander Rush, and Yann LeCun.
\newblock Adversarially regularized autoencoders.
\newblock In \emph{International Conference on Machine Learning}, pages
  5902--5911, 2018.

\end{thebibliography}

\end{document}